\documentclass[10pt,twocolumn,letterpaper]{article}

\usepackage[pagenumbers]{cvpr} 

\usepackage{microtype}
\renewcommand{\paragraph}[1]{\vspace{.5em}\noindent\textbf{#1.}}

\usepackage{xspace}

\usepackage{soul}
\setuldepth{foobar}

\definecolor{cVQA}{HTML}{B7E0AC}
\definecolor{cQA}{HTML}{D9CCEE}
\definecolor{cAVG}{HTML}{F7C5C0}
\definecolor{cOurs}{HTML}{D3E6F8}  
\definecolor{cSub}{HTML}{EEEEEE}
\def\confName{CVPR}
\def\confYear{2027}
\usepackage{multirow}                      
\usepackage{colortbl}                      
\definecolor{medframe}{RGB}{214,232,250}   
\graphicspath{{fig/}}

\definecolor{cvprblue}{rgb}{0.21,0.49,0.74}
\usepackage[pagebackref,breaklinks,colorlinks,allcolors=cvprblue,hypertexnames=false]{hyperref}

\newsavebox{\supptabbox}
\newsavebox{\suppfigbox}

\title{
MedFrame-R1: Reliable Reinforcement Learning Framework\\ for a Baseline Medical MLLM with Multimodal Reasoning}

\author{Shayan Nasiriboukani$^{1}$, Sara Atito$^{1,2,\star}$, Umar Marikkar$^{2}$, Iqra Farooq$^{2}$, Jiantao Wu$^{2}$,\\
Adinath Madhavrao Dukre$^{3}$, Imran Razzak$^{3}$, Muhammad Awais$^{1,2,\star}$\\[3pt]
\small $^{1}$\,Centre for Vision, Speech and Signal Processing (CVSSP), University of Surrey, UK\\
\small $^{2}$\,Surrey Institute for People-Centred AI, University of Surrey, UK\quad $^{3}$\,Mohamed bin Zayed University of Artificial Intelligence\\[2pt]
{\tt\small \{s.nasiriboukani,sara.atito,muhammad.awais\}@surrey.ac.uk}\\[-12pt]
}

\begin{document}
\maketitle
\begingroup\renewcommand\thefootnote{}\footnotetext{$^\star$Corresponding authors.}\endgroup
\begin{abstract}
Reinforcement learning (RL) with rewards from LLM judges is increasingly used to improve medical multimodal large language models (MLLMs), yet the reliability and reproducibility of these pipelines remain insufficiently examined. We identify four fundamental limitations: uninformative reward groups when sampled answers receive identical scores, reward exploitation through listing multiple possible answers or repetitive reasoning, insufficient domain knowledge in the starting model, and evaluation bias from reliance on a single judge. We introduce \textbf{MedFrame-R1}, an \textbf{open-source framework} in an effort to make medical MLLM reinforcement learning more \textbf{reliable} and \textbf{reproducible}. MedFrame-R1 measures and addresses the aforementioned limitations of existing medical MLLMs through four complementary components: dynamic sampling based on judge decisions, an embedding gate and length reward to discourage reward exploitation, a supervised cold start with chain-of-thought (CoT) reasoning, and an evaluation protocol. In a fair comparison with state-of-the-art (SoTA) medical MLLMs, averaged over the 4B and 8B/9B variants and 17 medical benchmarks, MedFrame-R1 achieves 1.5 macro and 3.3 micro average points over the existing SoTA on Qwen3-VL, and 3.0 macro and 5.2 micro average points over Qwen3.5, where only its RL stage is used. Code and models are publicly available \href{https://github.com/shayan137/MedFrame-R1}{here}.
\end{abstract}

\begin{figure}[t]
    \centering
    \includegraphics[width=\columnwidth]{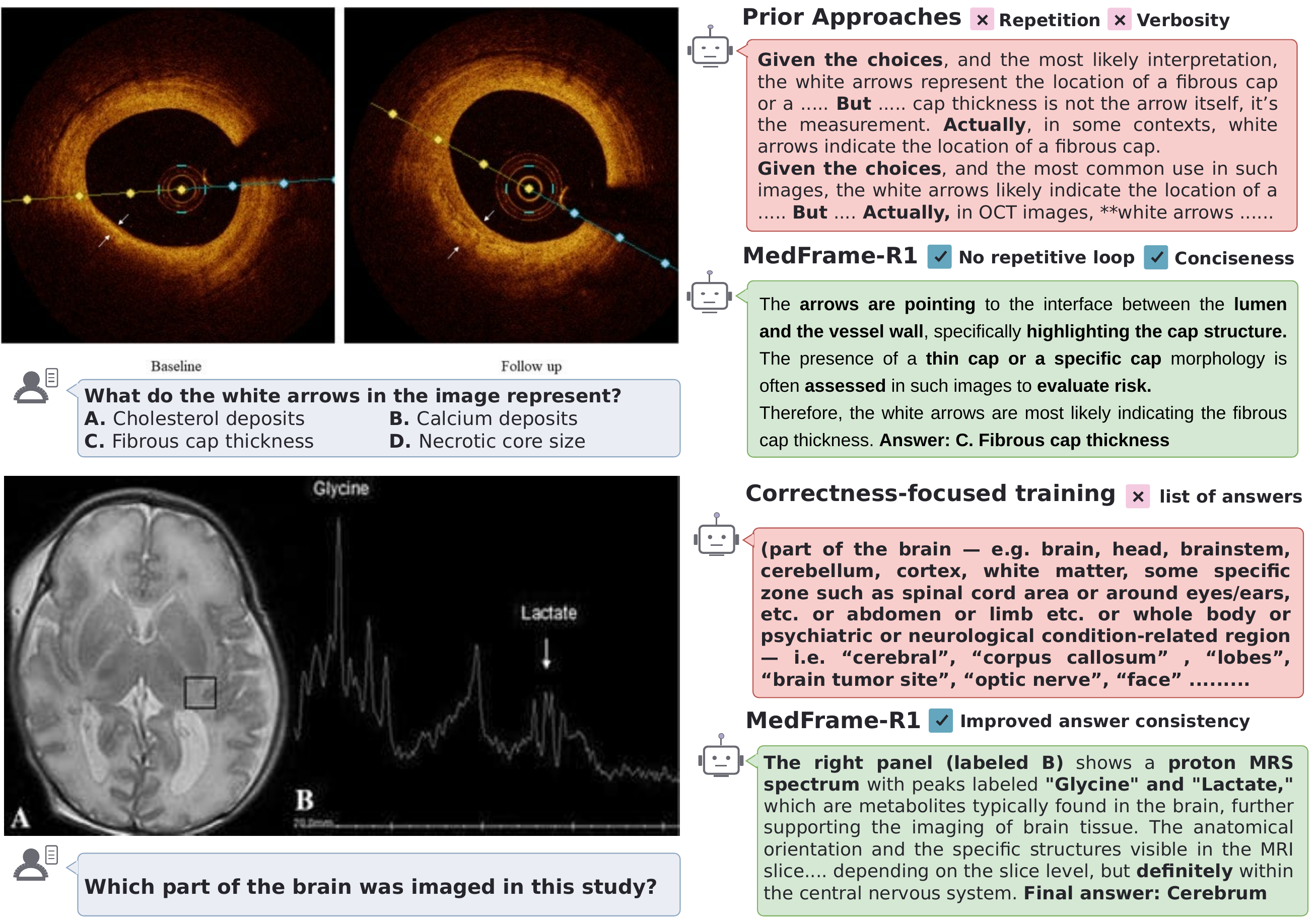}
    \caption{MedFrame-R1 avoids the repetitive reasoning of prior approaches and the answer listing that appears when training rewards correctness alone, and produces concise and focused responses.}
    \label{fig:teaser}
\end{figure}
\section{Introduction}
Interpreting a medical image means connecting what is visible with the question at hand and with prior clinical knowledge~\citep{tu2023generalistbiomedicalai}. A model must find the relevant findings, use what one imaging domain reveals to read another, and give an answer that the image supports. Multimodal large language models (MLLMs) do this in a single system, which makes them promising for medical question answering and image understanding~\citep{li2023llava}. Their accuracy still varies across imaging modalities, because the relevant visual features and clinical knowledge differ between modalities and may not have been learned during pretraining.

Supervised fine-tuning (SFT) adapts MLLMs to medical tasks by learning from reference answers, while reinforcement learning (RL) further improves their responses through rewards assigned to model-generated outputs~\citep{pmlr-v306-zhao26bz, zhang2026reinforcementfinetuningenablesmllms, ouyang2022traininglanguagemodelsfollow, wei2026reinforcementlearningimprovesllm}. Recent medical RL approaches commonly use group-relative advantages and LLM judges to assess answer correctness~\citep{shao2024deepseekmathpushinglimitsmathematical, NEURIPS2025_a4277440, lai2026med, pmlr-v315-gundersen26a}. However, improving benchmark accuracy alone does not establish that these pipelines are reliable. Their effectiveness depends on the availability of informative rewards, the robustness of judge decisions, sufficient medical knowledge in the starting model, and consistent evaluation. We identify four fundamental limitations that arise when these requirements are not adequately addressed.

\paragraph{First, limited learning signals} Group Relative Policy Optimization (GRPO) learns by comparing the rewards of several rollouts to the same question~\citep{shao2024deepseekmathpushinglimitsmathematical}. When all of them receive the same correctness reward, their relative advantages are zero and the question gives no correctness signal~\citep{NEURIPS2025_a4277440, le2026no, wang2026reinforcement, he2026advantage, zhou2026reasoning}. When zero-shot Qwen3-VL answers our 51K training samples, only about 6.5K produce mixed outcomes. The number grows as the policy learns, but plain GRPO still reaches only 11{,}439 trainable samples, less than half of the 26K reached by MedFrame-R1. The problem is sharper for specialised clinical cases with little training data~\citep{nam2025multimodal, han2025diagnostic}. The useful samples also change as the policy improves, so a fixed data selection loses its value~\citep{girgis2026constrained}.

\paragraph{Second, unreliable rewards}
LLM-based judges can be exploited by policies that learn to maximise reward without improving answer correctness~\citep{ICLR2025_fdca08d3, yang2026rubricdropoutsimpleway}. In our setting, a model may generate a list of plausible answers, including incorrect or unsupported ones, to increase the chance that the judge identifies a correct answer and assigns a positive reward~\citep{yao2026multimodalrewardhackingreinforcement}. Such behaviour can lead to reward hacking, where high reward no longer reflects reliable predictions. Moreover, without explicit length constraints, rollouts may become unnecessarily long and repetitive. These limitations motivate additional verification of judge decisions and explicit control over rollout length~\citep{ichihara2026mogrpomitigatingrewardhacking, kwok2026llmasaverifiergeneralpurposeverificationframework}.

\paragraph{Third, insufficient domain knowledge}
Effective RL requires a starting model capable of producing correct answers for at least some sampled rollouts. When the model lacks sufficient medical knowledge, particularly for rare imaging modalities and difficult clinical cases, all sampled answers may be incorrect, leaving no correctness-based learning signal. In our initial diagnostic run, approximately 21K training samples never produced an accepted answer. A supervised cold start can provide the domain knowledge and structured reasoning needed to make these cases learnable. However, the initialisation must be carefully controlled, as stronger supervised performance does not necessarily translate into better RL outcomes when rollout diversity is reduced~\citep{ICLR2026_bebed4d2}.

\paragraph{Fourth, evaluation bias}
Reliable comparison of medical MLLMs requires consistent and independent assessment of their predictions. However, evaluation based on a single LLM judge can depend on the judge's model family and scoring behaviour, potentially favouring related models~\citep{panickssery2024llm}. Repeated assessments can also be affected by anchoring when previous scores are visible~\citep{kapetanovic2026anchoringbiasllmasajudgesystems}. Such dependencies make reported results harder to reproduce and compare across methods. Deterministic scoring where possible, combined with independent judgements from multiple model families, can reduce reliance on any individual evaluator. \Cref{tab:intro_failures} compares the training design and observed failure modes of existing medical RL approaches.

\begin{table}[t]
\centering
\caption{\textbf{RL design and failure modes of medical MLLMs.}
Columns indicate informative-sample selection, explicit
reasoning, answer-listing avoidance, and length control.
A cross denotes an unmet criterion.
Failure examples are shown in
\cref{fig:app-fail-1,fig:app-fail-2}.}
\footnotesize
\newcommand{\yes}{\textcolor{ForestGreen}{$\checkmark$}}
\newcommand{\no}{\textcolor{BrickRed}{$\boldsymbol{\times}$}}
\newcommand{\hack}{\textcolor{BrickRed}{$\boldsymbol{\times}$}}
\newcommand{\clean}{\textcolor{ForestGreen}{$\checkmark$}}
\setlength{\tabcolsep}{4.5pt}
\renewcommand{\arraystretch}{1.15}
\setlength{\aboverulesep}{0pt}
\setlength{\belowrulesep}{0pt}
\begin{tabular}{lcccc}
\toprule
Method & \begin{tabular}[c]{@{}c@{}}Informative\\samples\end{tabular} & \begin{tabular}[c]{@{}c@{}}Explicit\\reasoning\end{tabular} & \begin{tabular}[c]{@{}c@{}}Answer\\listing\end{tabular} & \begin{tabular}[c]{@{}c@{}}Length\\control\end{tabular} \\
\midrule
MedVLM-R1~\citep{pan2025medvlm} & \no & \yes & \clean & \no \\
MedMO~\citep{deria2026medmo} & \no & \no & \clean & \yes \\
QoQ-Med~\citep{dai2026qoq} & \no & \yes & \hack & \no \\
MediX-R1~\citep{mullappilly2026medix} & \no & \yes & \hack & \no \\
\midrule
\rowcolor{cOurs}
\textbf{MedFrame-R1 (ours)} & \yes & \yes & \clean & \yes \\
\bottomrule
\end{tabular}

\label{tab:intro_failures}
\end{table}

These limitations affect not only the performance of medical MLLMs but also the reliability of their training and the reproducibility of their evaluation. As shown in \cref{tab:intro_failures}, existing state-of-the-art medical RL methods exhibit different weaknesses, including uninformative training samples, answer listing, and uncontrolled response length. Rather than introducing a more complex optimisation algorithm, we propose \textbf{MedFrame-R1}, an \textbf{open-source framework for reliable and reproducible medical MLLM reinforcement learning} that systematically addresses these limitations. MedFrame-R1 integrates dynamic sampling, reward verification, a supervised cold start, and independent multi-judge evaluation within a unified framework. Our experiments demonstrate that addressing these fundamental problems improves learning efficiency, reward robustness, and downstream performance across different model families.

Our \textbf{contributions} address these four limitations:
\begin{enumerate}

\item \textbf{Informative sampling:}
We introduce dynamic sampling based on raw judge decisions to select training groups with non-uniform correctness outcomes. This expands the pool of samples providing a correctness-based learning signal from approximately 6.5K to 26K.

\item \textbf{Reward robustness:}
We introduce an embedding gate that verifies semantic agreement between predicted and reference answers before accepting positive judge rewards, reducing exploitation through multiple answers. Also, the length reward discourages unnecessarily long and repetitive responses, promoting concise reasoning.

\item \textbf{Medical domain knowledge:}
We apply a supervised cold start using structured chain-of-thought reasoning for all 51K training examples. Its Adaptive Early-Stop Loss (AESL) balances the acquisition of medical knowledge with the preservation of rollout diversity required for effective RL.

\item \textbf{Reproducible evaluation:}
We establish an evaluation protocol that scores approximately 70\% of questions using deterministic rules and the remaining 30\% using three independent LLM judges from different model families. Majority voting and independent assessments reduce dependence on individual judges and support consistent comparisons across models.

\end{enumerate}

\section{Related Work}
\label{sec:related_work}

\paragraph{Medical Multimodal Large Language Models
(MLLMs)}
MLLMs connect visual encoders with pretrained language models to interpret medical images and answer questions using biomedical knowledge. Early approaches established these capabilities by adapting general vision and language models through biomedical pretraining and instruction tuning~\citep{li2023llava, zhang2023pmc}, while others explored few-shot in-context learning~\citep{moor2023med}. Building on these foundations, subsequent work has pursued two complementary directions: deeper expertise within individual specialties and broader coverage across medical domains. Specialised models adapt their visual representations and training data to chest radiography and radiology report generation~\citep{thawakar2024xraygpt, pellegrini2025radialog, hyland2023maira, chen2024vision, bannur2024maira}, dermatology~\citep{zhou2023skingpt}, and pathology~\citep{sun2024pathasst, lu2023foundational}. Depending on their design, these systems support tasks ranging from image interpretation and report generation to finding localisation and region-level grounding. Broader models instead bring together heterogeneous biomedical data to support multiple tasks~\citep{wu2025towards,zhang2023biomedgpt, chen2024towards} and imaging modalities, including volumetric 3D scans, within a shared framework~\citep{bai2024m3d}. Generalist biomedical systems further expand this scope by integrating diverse medical images with clinical information and, in some cases, genomic data~\citep{yang2024advancing}. Despite differences in scope, these approaches primarily develop their capabilities through pretraining, supervised fine-tuning, and in-context learning. Their progress raises a further question: how can models use feedback on their own generated answers to improve medical reasoning beyond these forms of adaptation? This motivates investigating reinforcement learning as a complementary training approach.

\paragraph{Reinforcement Learning for Multimodal Reasoning}
Medical multimodal models use reinforcement learning to improve different aspects of reasoning according to the target task. For medical question answering, RL encourages correct answers and aims to improve generalisation, commonly using GRPO~\citep{pan2025medvlm, lai2026med} and verifiable correctness rewards~\citep{huang2025medvlthinker}. Beyond rewarding final answers, radiological reasoning methods use process supervision to assess intermediate clinical reasoning steps~\citep{fan2025chestx}. Grounding and localisation methods extend reward design to spatial accuracy, using signals such as region overlap and agreement between predicted locations and target abnormalities~\citep{yan2026medreasoner, deria2026medmo}. At a broader level, generalist medical models incorporate RL into post-training across multiple tasks and modalities~\citep{sellergren2025medgemma}, while domain-aware approaches adjust optimisation to account for differences in domain rarity and difficulty~\citep{dai2026qoq}. For open-ended medical rollouts, composite rewards combine judge-based correctness, medical embedding similarity, rollout format, and modality recognition to provide feedback beyond exact answer matching~\citep{mullappilly2026medix}. Together, these approaches show how RL can support medical reasoning through task-specific feedback. However, their effectiveness also depends on whether sampled rollouts provide informative reward differences and whether those differences reflect genuine improvements in answer quality.

\paragraph{Zero-Advantage Challenges in Medical Reinforcement Learning}
A central limitation of group-based RL is that some samples produce no relative learning signal. In GRPO, when all sampled rollouts receive the same reward, their relative advantages become zero, removing their contribution to the advantage-based policy update~\citep{shao2024deepseekmathpushinglimitsmathematical}. This can occur when every rollout is correct or when all rollouts are incorrect. The latter is a particular concern when applying general MLLMs to medicine, where gaps in specialised knowledge and limited coverage of rare modalities may make difficult questions consistently unsolvable under the current policy~\citep{nam2025multimodal, han2025diagnostic}. Existing approaches address this limitation in several ways. Hint-based methods provide additional guidance during training to help the model solve difficult examples and recover reward variation~\citep{zhang2026bread, zhang2026scaf, amani2506rl, qu2601pope}. However, conditioning training rollouts on hints can create a mismatch with inference, where such guidance is usually unavailable, so improvements on hinted inputs may not fully transfer to answering without hints~\citep{agrawal2026off}. Other methods modify the optimisation objective to learn directly from groups with identical rewards. RL-ZVP, for example, uses entropy-guided advantage shaping to reward correct rollouts and penalise incorrect ones without requiring contrasting outcomes within a group~\citep{le2026no}. A complementary approach is supervised cold-start initialisation, which provides reasoning examples before RL to strengthen the starting policy~\citep{ICLR2026_bebed4d2}. These strategies act on different parts of the learning process: the information supplied during training, the construction of advantages, and the model's initial capabilities.

\section{Methodology}
\label{sec:methodology}

\begin{figure*}[t]
    \centering
    \includegraphics[width=0.9\textwidth]{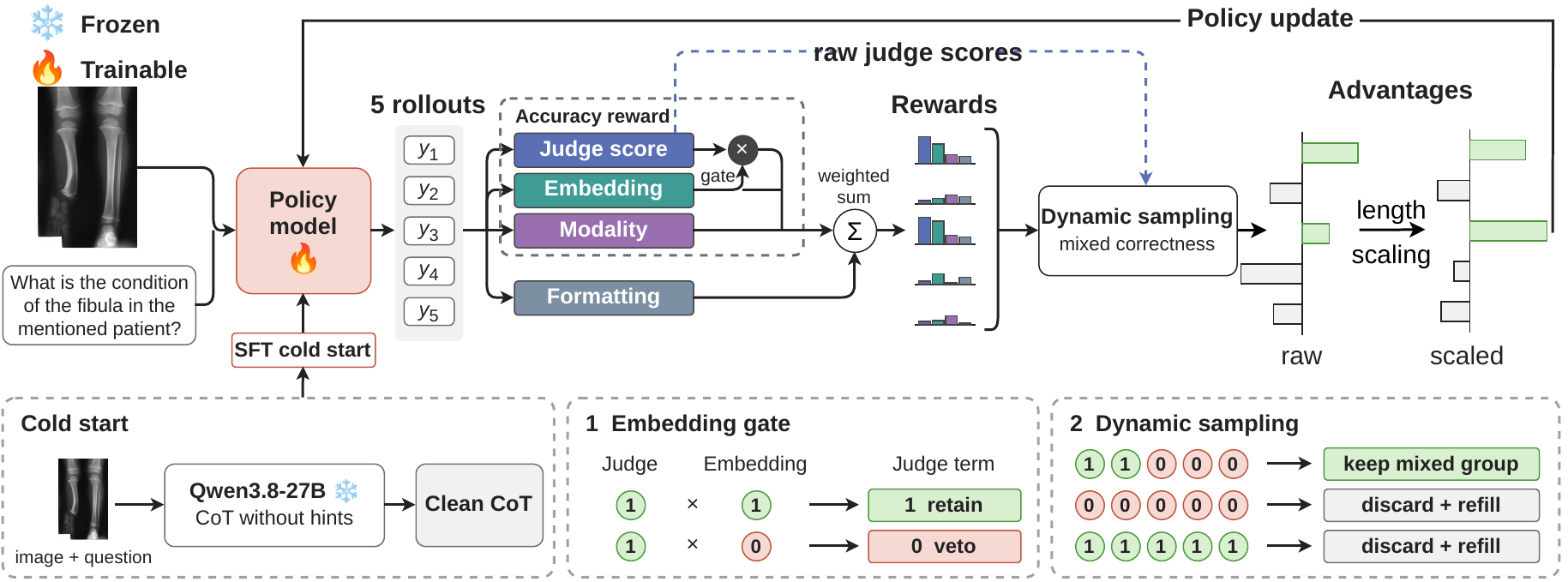}
 
    \caption{\textbf{MedFrame-R1 training pipeline.}
    A frozen reasoning model generates chain-of-thought supervision for the SFT cold start. During RL, the policy generates five rollouts per sample, evaluated using judge, embedding, modality, and formatting rewards. (1) An embedding gate verifies positive judge scores against reference answers. (2) Dynamic sampling retains groups with mixed raw judge scores. Finally, the length reward favours short correct rollouts and penalises long incorrect ones before the policy update.}

    \label{fig:medframe_main_pipeline}
\end{figure*}

The overall methodology and training pipeline of MedFrame-R1 are illustrated in \cref{fig:medframe_main_pipeline}. Our framework consists of two main stages. (1) Supervised cold start, which first generates chain-of-thought reasoning traces and corresponding answers, followed by supervised fine-tuning using Adaptive Early-Stop Loss (AESL) to improve medical reasoning while preserving model diversity for reinforcement learning. (2) Reinforcement learning, which employs the dynamic sampling of DAPO (Decoupled Clip and Dynamic sAmpling Policy Optimization)~\citep{NEURIPS2025_a4277440} to prioritise samples with informative reward variation and introduces an embedding gate to validate the LLM judge's correctness rewards and prevent reward hacking. The following subsections describe each stage and its key components, while the evaluation protocol is presented in \cref{sec:evaluation_protocol}.

\subsection{Structured Reasoning Cold Start}
\label{sec:reasoning_cold_start}

Medical MLLMs can struggle with rare modalities and difficult clinical questions~\citep{hu2024omnimedvqa}. These examples are particularly challenging for group-based RL: when every sampled rollout is incorrect, the correctness reward provides no within-group variation and therefore no relative learning signal~\citep{NEURIPS2025_a4277440}. We therefore begin with a reasoning-based supervised cold start, aiming to expose the model to the full training distribution and provide a stronger initialisation for difficult examples before RL~\citep{feng2025videor1reinforcingvideoreasoning,ICLR2026_bebed4d2}. For each training sample, we use a stronger model~\citep{qwen38} to generate a structured chain of thought (CoT) together with the final answer. Generation proceeds in two passes. First, the model solves the problem using only the image and question without seeing the reference answer. This encourages the reasoning to follow from the input rather than being constructed afterward to justify a known answer, reducing the risk of post-hoc rationalisation~\citep{turpin2023language}. If the answer is correct, we keep the reasoning. Otherwise, we run a second pass with the reference answer provided as guidance so that difficult examples are not discarded~\citep{zelikman2022star}. This gives difficult examples another opportunity to enter the supervised dataset while preserving answer-blind reasoning whenever possible.

Having constructed these demonstrations, we next consider how to learn from their reasoning and answers. The standard cross-entropy (CE) objective is
\begin{equation}
    \mathcal{L}_{\mathrm{CE}}(\theta)
    =
    -\mathbb{E}_{q,s^*\sim\mathcal{D}_{\mathrm{SFT}}}
    \left[
        \log\pi_\theta(s_t^*\mid q,s_{<t})
    \right],
    \label{eq:cold-start-ce}
\end{equation}
where $\mathcal{D}_{\mathrm{SFT}}$ denotes the dataset, $q$ is the input, $s_t^*$ is the target token, and $\pi_\theta(s_t^* \mid q, s_{<t}^*)$ represents the probability of the target token given the input and preceding tokens. Since cross-entropy (CE) loss supervises every target token, longer CoT rationales contribute substantially more to the training objective than the shorter final answers. To emphasise final answer prediction while retaining supervision for reasoning and formatting, we adopt answer-weighted cross-entropy (wCE) loss. However, this weighting alone does not distinguish between confidently predicted tokens and those requiring further learning, potentially introducing unnecessary updates during the supervised cold start. We therefore incorporate Adaptive Early-Stop Loss (AESL)~\citep{ICLR2026_bebed4d2}, which adaptively reduces supervision on confident tokens to preserve the base model's distribution while learning new patterns. The adaptive weighting is defined as

\begin{multline}
    p(q,s_t^*,\pi_\theta)
    = \\
    1-\operatorname{softmax}\left[
        \frac{
            y(s_t^*\mid q,s_{<t})
        }{
            -t_{\mathrm{scaling}}\cdot
            \frac{1}{|t|}
            \sum_{i=1}^{t}
            \log\pi_\theta(s_i^*\mid q,s_{<i})
        }
    \right].
    \label{eq:cold-start-aesl-weight}
\end{multline}

Here, $y$ denotes the pre-softmax logits, and $t_{\mathrm{scaling}}$ controls temperature scaling. By combining answer weighting with AESL, we prioritise final answer prediction while reducing unnecessary supervision on confident tokens, providing a more suitable initialisation for reinforcement learning. The complete loss formulations, implementation details, and training hyperparameters are provided in \cref{sec:sft} of the supplementary material.

\subsection{Embedding Gating of Judge Rewards} \label{sec:reward_gating}

In the reinforcement learning stage, we employ DAPO~\citep{NEURIPS2025_a4277440} dynamic sampling based on LLM judge correctness rewards to retain only samples whose rollouts exhibit different correctness outcomes, ensuring that training focuses on informative examples. We evaluate each rollout using four reward signals: rollout format, modality recognition, answer correctness, and semantic similarity. The format and modality rewards encourage the model to maintain the structured rollouts and modality recognition learned during supervised fine-tuning. Answer correctness is assessed by an LLM judge, which compares the predicted answer with the reference based on semantic equivalence. However, the judge may incorrectly reward overly broad rollouts that include the correct answer among several alternatives, potentially encouraging reward hacking. To address this, we introduce an embedding gate that validates positive judge decisions against the reference answer, preventing semantically inconsistent predictions from receiving correctness rewards. The final reward combines gated judge correctness, embedding similarity, modality recognition, and format compliance. Further analysis of reward hacking, reward formulation, coefficients, and implementation details is provided in \cref{sec:app-reward-hacking} of the supplementary material.

\begin{table*}[t]
\caption{\textbf{Performance on medical VQA and QA benchmarks.}
Models are grouped by parameter scale (up to 4B and 7B--9B).
\textbf{Bold} and \underline{underlined} values indicate the best
and second-best scores within each group, including ties.
Column headers are colour-coded:
\colorbox{cVQA}{\strut VQA benchmarks},
\colorbox{cQA}{\strut QA benchmarks}, and
\colorbox{cAVG}{\strut averages}.
\emph{VQA} and \emph{QA} are the means over 6 image-based and
11 text-based benchmarks, respectively.
\emph{Macro} is the unweighted mean across 17 benchmarks,
while \emph{Micro} is the pooled accuracy over 19{,}200 samples.
The backbone is indicated for MedFrame-R1 and MediX-R1,
and $^{\dagger}$ denotes our reproduction of MediX-R1.
See Supplementary \cref{app:full_results} (\cref{tab:main_full}).}
\label{tab:main}
\centering
\small
\setlength{\tabcolsep}{5.5pt}
\renewcommand{\arraystretch}{1.1}
\setlength{\aboverulesep}{0pt}\setlength{\belowrulesep}{0pt}
\resizebox{\textwidth}{!}{%
\begin{tabular}{l|cccccc|ccccccccccc|cccc}
\toprule
\multicolumn{1}{c|}{\rule{0pt}{2.2cm}} & \multicolumn{6}{c|}{\cellcolor{cVQA}} & \multicolumn{11}{c|}{\cellcolor{cQA}} & \multicolumn{4}{c}{\cellcolor{cAVG}} \\[-2.37cm]
\multicolumn{1}{c|}{\rule{0pt}{2.2cm}\raisebox{0.93cm}[0pt][0pt]{\large\textbf{Models}}}
& \multicolumn{1}{c}{\raisebox{0.93cm}[0pt][0pt]{\makebox[0pt][c]{\rotatebox[origin=c]{65}{\footnotesize\textbf{SLAKE}}}}}
& \multicolumn{1}{c}{\raisebox{0.93cm}[0pt][0pt]{\makebox[0pt][c]{\rotatebox[origin=c]{65}{\footnotesize\textbf{VQA-RAD}}}}}
& \multicolumn{1}{c}{\raisebox{0.93cm}[0pt][0pt]{\makebox[0pt][c]{\rotatebox[origin=c]{65}{\footnotesize\textbf{PathVQA}}}}}
& \multicolumn{1}{c}{\raisebox{0.93cm}[0pt][0pt]{\makebox[0pt][c]{\rotatebox[origin=c]{65}{\footnotesize\textbf{PMC-VQA}}}}}
& \multicolumn{1}{c}{\raisebox{0.93cm}[0pt][0pt]{\makebox[0pt][c]{\rotatebox[origin=c]{65}{\footnotesize\textbf{PMC-VQA-H}}}}}
& \multicolumn{1}{c|}{\raisebox{0.93cm}[0pt][0pt]{\makebox[0pt][c]{\rotatebox[origin=c]{65}{\footnotesize\textbf{MIMIC-Gen}}}}}
& \multicolumn{1}{c}{\raisebox{0.93cm}[0pt][0pt]{\makebox[0pt][c]{\rotatebox[origin=c]{65}{\footnotesize\textbf{MMLU-Clin}}}}}
& \multicolumn{1}{c}{\raisebox{0.93cm}[0pt][0pt]{\makebox[0pt][c]{\rotatebox[origin=c]{65}{\footnotesize\textbf{MMLU-Bio}}}}}
& \multicolumn{1}{c}{\raisebox{0.93cm}[0pt][0pt]{\makebox[0pt][c]{\rotatebox[origin=c]{65}{\footnotesize\textbf{MMLU-Med}}}}}
& \multicolumn{1}{c}{\raisebox{0.93cm}[0pt][0pt]{\makebox[0pt][c]{\rotatebox[origin=c]{65}{\footnotesize\textbf{MMLU-Gen}}}}}
& \multicolumn{1}{c}{\raisebox{0.93cm}[0pt][0pt]{\makebox[0pt][c]{\rotatebox[origin=c]{65}{\footnotesize\textbf{MMLU-Prof}}}}}
& \multicolumn{1}{c}{\raisebox{0.93cm}[0pt][0pt]{\makebox[0pt][c]{\rotatebox[origin=c]{65}{\footnotesize\textbf{MMLU-Anat}}}}}
& \multicolumn{1}{c}{\raisebox{0.93cm}[0pt][0pt]{\makebox[0pt][c]{\rotatebox[origin=c]{65}{\footnotesize\textbf{MedMCQA}}}}}
& \multicolumn{1}{c}{\raisebox{0.93cm}[0pt][0pt]{\makebox[0pt][c]{\rotatebox[origin=c]{65}{\footnotesize\textbf{MedQA}}}}}
& \multicolumn{1}{c}{\raisebox{0.93cm}[0pt][0pt]{\makebox[0pt][c]{\rotatebox[origin=c]{65}{\footnotesize\textbf{USMLE-SA}}}}}
& \multicolumn{1}{c}{\raisebox{0.93cm}[0pt][0pt]{\makebox[0pt][c]{\rotatebox[origin=c]{65}{\footnotesize\textbf{PubMedQA}}}}}
& \multicolumn{1}{c|}{\raisebox{0.93cm}[0pt][0pt]{\makebox[0pt][c]{\rotatebox[origin=c]{65}{\footnotesize\textbf{MIMIC-Sum}}}}}
& \multicolumn{1}{c}{\raisebox{0.93cm}[0pt][0pt]{\makebox[0pt][c]{\rotatebox[origin=c]{65}{\footnotesize\textbf{VQA}}}}}
& \multicolumn{1}{c}{\raisebox{0.93cm}[0pt][0pt]{\makebox[0pt][c]{\rotatebox[origin=c]{65}{\footnotesize\textbf{QA}}}}}
& \multicolumn{1}{c}{\raisebox{0.93cm}[0pt][0pt]{\makebox[0pt][c]{\rotatebox[origin=c]{65}{\footnotesize\textbf{Macro}}}}}
& \multicolumn{1}{c}{\raisebox{0.93cm}[0pt][0pt]{\makebox[0pt][c]{\rotatebox[origin=c]{65}{\footnotesize\textbf{Micro}}}}} \\
\midrule

MedVLM-R1-2B
& 41.3 & 25.1 & 7.3 & 44.9 & 25.7 & 9.6
& 54.3 & 54.2 & 49.7 & 56.0 & 50.7 & 51.8 & 40.4 & 40.4 & 37.5 & \underline{53.2} & 49.2
& 25.6 & 48.9 & 40.7 & 28.7 \\

MedGemma-4B-IT
& 67.8 & \textbf{65.9} & 31.7 & 44.4 & 21.4 & 20.5
& 70.8 & 70.6 & 60.5 & 82.0 & 71.3 & 55.6 & 57.0 & 62.1 & 63.9 & 47.0 & \textbf{69.2}
& 42.0 & 64.5 & 56.6 & 45.8 \\

MedMO-4B
& 39.0 & 34.6 & 36.7 & 39.3 & 14.8 & 0.6
& 81.1 & 86.8 & 77.5 & 86.0 & 79.4 & 70.4 & 57.1 & 66.4 & 67.7 & \textbf{55.4} & 42.2
& 27.5 & 70.0 & 55.0 & 45.1 \\

Qwen3-VL-4B
& 67.9 & 54.3 & 40.0 & 53.0 & 34.2 & 31.6
& 78.9 & 88.2 & 80.3 & 88.0 & 78.7 & 71.8 & 58.7 & 65.9 & 71.1 & 40.6 & \underline{54.2}
& 46.8 & 70.6 & 62.2 & 51.5 \\

MediX-R1-4B (Qwen3-VL)$^{\dagger}$
& 73.7 & 58.3 & 44.6 & 54.8 & 34.3 & 29.4
& 76.6 & 88.2 & 79.8 & 87.0 & 77.9 & 70.4 & 58.6 & 67.2 & 73.2 & 44.4 & 52.1
& 49.2 & 70.5 & 63.0 & 53.8 \\

Qwen3.5-4B
& 69.7 & 60.5 & 46.4 & \underline{60.2} & \underline{38.9} & \underline{47.8}
& \underline{84.9} & \underline{93.1} & \underline{86.1} & \underline{92.0} & \underline{88.6} & \underline{78.5} & \underline{66.0} & \underline{82.6} & \underline{87.7} & 47.0 & 47.0
& \underline{53.9} & \underline{77.6} & \underline{69.2} & \underline{58.6} \\

\midrule
\rowcolor{cOurs}
\textbf{MedFrame-R1-4B (Qwen3-VL)}
& \underline{74.4} & 61.6 & \underline{52.1} & 54.7 & 36.0 & 32.4
& 80.4 & 90.3 & 79.2 & 88.0 & 79.8 & 69.6 & 60.1 & 70.2 & 71.4 & 49.8 & 50.8
& 51.9 & 71.8 & 64.8 & 57.4 \\

\rowcolor{cOurs}
\textbf{MedFrame-R1-4B (Qwen3.5)}
& \textbf{81.4} & \underline{64.7} & \textbf{56.4} & \textbf{61.3} & \textbf{42.3} & \textbf{51.0}
& \textbf{89.4} & \textbf{97.2} & \textbf{86.7} & \textbf{95.0} & \textbf{93.4} & \textbf{83.0} & \textbf{68.4} & \textbf{85.8} & \textbf{90.8} & 46.6 & 46.4
& \textbf{59.5} & \textbf{80.2} & \textbf{72.9} & \textbf{64.2} \\

\midrule

Qwen2.5-VL-7B
& 47.0 & 39.9 & 8.4 & 51.4 & 32.3 & 25.0
& 76.2 & 84.0 & 72.2 & 81.0 & 74.3 & 69.6 & 53.3 & 60.2 & 62.1 & 39.8 & \textbf{55.0}
& 34.0 & 66.2 & 54.8 & 36.6 \\

HuatuoGPT-V-7B
& 60.1 & 63.2 & 35.5 & 53.7 & 35.8 & 25.0
& 72.5 & 75.0 & 67.6 & 77.0 & 70.6 & 57.8 & 51.7 & 54.3 & 58.1 & \textbf{56.0} & 53.6
& 45.6 & 63.1 & 56.9 & 47.1 \\

LLaVA-Med-7B
& 47.1 & 44.3 & 33.2 & 29.9 & 26.8 & 11.6
& 62.6 & 64.6 & 58.4 & 56.0 & 55.5 & 56.3 & 38.8 & 43.0 & 41.8 & 11.8 & 28.8
& 32.2 & 47.1 & 41.8 & 36.2 \\

QoQ-Med-VL-7B
& 51.4 & 44.4 & 10.7 & 49.4 & 30.0 & 27.8
& 77.0 & 84.0 & 72.8 & 78.0 & 76.1 & 70.4 & 53.5 & 59.6 & 61.5 & 46.4 & 51.8
& 35.6 & 66.5 & 55.6 & 37.6 \\

InternVL2.5-8B
& 54.3 & 40.4 & 12.0 & 46.8 & 29.3 & 30.4
& 75.5 & 77.8 & 71.7 & 82.0 & 75.7 & 67.4 & 50.6 & 54.8 & 60.9 & 49.8 & 50.6
& 35.5 & 65.2 & 54.7 & 36.8 \\

InternVL3-8B
& 63.1 & 43.5 & 18.2 & 52.6 & 33.1 & 38.2
& 81.1 & 86.1 & 75.1 & 80.0 & 79.0 & 68.2 & 57.9 & 64.0 & 68.0 & 50.6 & \underline{54.4}
& 41.5 & 69.5 & 59.6 & 43.1 \\

BiMediX2-8B
& 44.3 & 46.1 & 30.5 & 46.8 & 27.2 & 20.6
& 68.7 & 77.1 & 67.0 & 76.0 & 68.0 & 66.7 & 54.8 & 58.9 & 56.3 & \underline{52.8} & 51.0
& 35.9 & 63.4 & 53.7 & 43.6 \\

Qwen3-VL-8B
& 61.6 & 52.8 & 36.0 & 53.2 & 34.0 & 34.4
& 80.4 & 92.4 & 80.3 & 87.0 & 84.6 & 75.6 & 62.3 & 75.9 & 77.2 & 42.0 & 53.0
& 45.3 & 73.7 & 63.7 & 51.4 \\

MediX-R1-8B (Qwen3-VL)$^{\dagger}$
& 70.2 & 59.9 & 46.6 & 53.8 & 34.7 & 30.6
& 80.0 & 93.8 & 81.5 & 89.0 & 83.5 & \underline{77.0} & 62.6 & 75.3 & 79.4 & 43.8 & 52.6
& 49.3 & 74.4 & 65.5 & 56.0 \\

Qwen3.5-9B
& 70.7 & \underline{65.6} & 48.7 & \underline{61.1} & \underline{41.4} & \underline{49.2}
& \underline{89.4} & \textbf{98.6} & \textbf{89.0} & \underline{98.0} & \underline{91.9} & \textbf{83.0} & \underline{70.8} & \underline{85.9} & \underline{88.0} & 48.2 & 48.4
& \underline{56.1} & \underline{81.0} & \underline{72.2} & \underline{61.4} \\

\midrule
\rowcolor{cOurs}
\textbf{MedFrame-R1-8B (Qwen3-VL)}
& \underline{76.5} & \underline{65.6} & \underline{53.8} & 56.2 & 34.8 & 33.4
& 82.6 & 91.7 & 82.7 & 88.0 & 85.3 & 74.1 & 61.8 & 73.6 & 76.0 & 46.6 & 52.8
& 53.4 & 74.1 & 66.8 & 59.1 \\

\rowcolor{cOurs}
\textbf{MedFrame-R1-9B (Qwen3.5)}
& \textbf{80.2} & \textbf{71.4} & \textbf{56.6} & \textbf{64.9} & \textbf{43.7} & \textbf{51.6}
& \textbf{89.8} & \underline{97.9} & \underline{86.1} & \textbf{99.0} & \textbf{94.5} & \textbf{83.0} & \textbf{73.8} & \textbf{89.9} & \textbf{93.8} & 43.2 & 47.0
& \textbf{61.4} & \textbf{81.6} & \textbf{74.5} & \textbf{66.3} \\

\bottomrule
\end{tabular}}
\end{table*}

To further address reward hacking through repetitive and excessively long rollouts, we use a length reward~\citep{li2026drpo}, which treats correct and incorrect rollouts differently. For correct rollouts with positive advantages, shorter rollouts receive greater weight, while longer incorrect rollouts with negative advantages receive stronger penalties. This discourages unnecessary generation while preserving the original reward function and allowing sufficient reasoning. Further details on dynamic sampling, advantage computation, length scaling, and training hyperparameters are provided in Supplementary \cref{sec:app-dapo,app:reward-length}.

\begin{figure*}[t]
    \centering

    \begin{subfigure}[t]{0.24\textwidth}
        \centering
        \includegraphics[width=\linewidth]{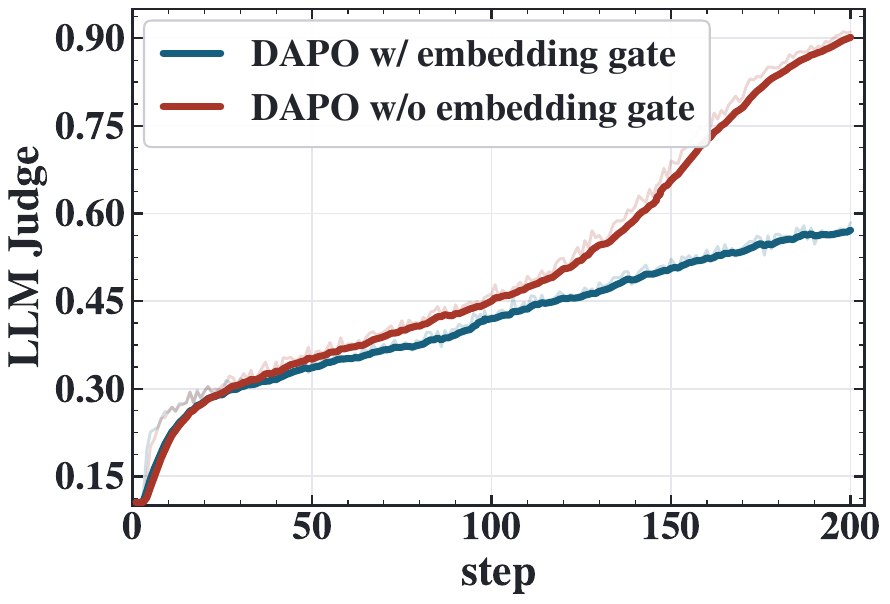}
        \caption{\textbf{LLM judge accuracy.}
        Without the embedding gate, judge reward continues to increase.}
        \label{fig:judge_accuracy}
    \end{subfigure}
    \hfill
    \begin{subfigure}[t]{0.24\textwidth}
        \centering
        \includegraphics[width=\linewidth]{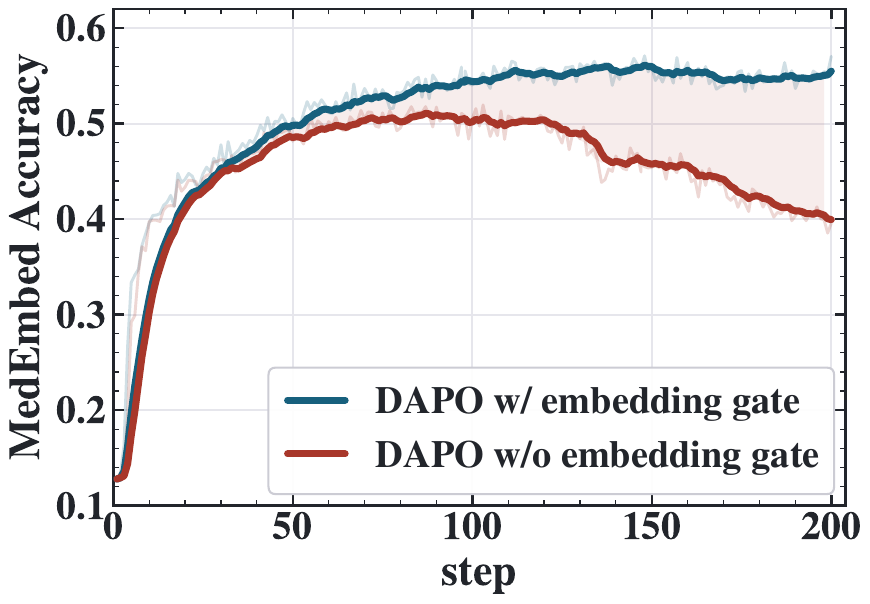}
        \caption{\textbf{Embedding accuracy.}
        Semantic agreement declines without the embedding gate.}
        \label{fig:embedding_accuracy}
    \end{subfigure}
    \hfill
    \begin{subfigure}[t]{0.24\textwidth}
        \centering
        \includegraphics[width=\linewidth]{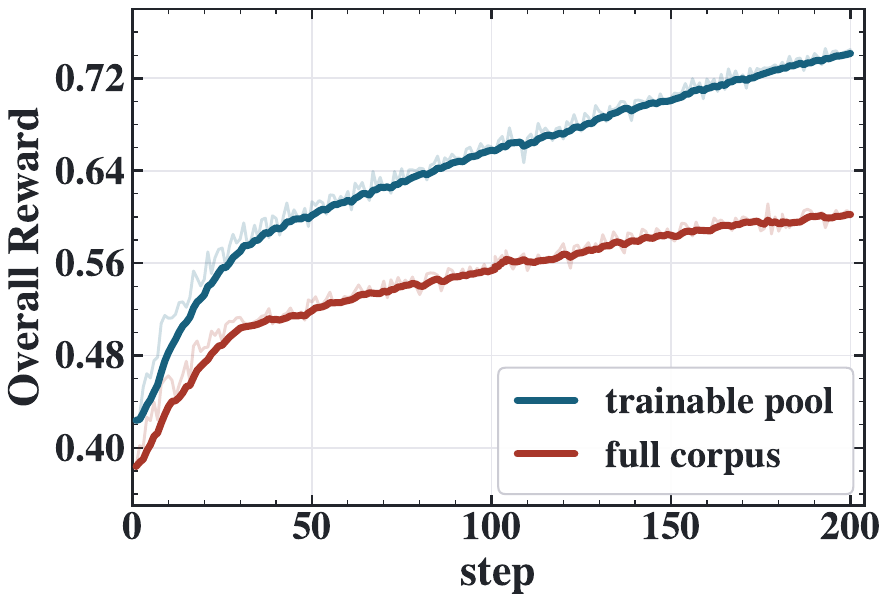}
        \caption{\textbf{Informative training signals.}
        The trainable pool maintains higher rewards.}
        \label{fig:trainable_reward}
    \end{subfigure}
    \hfill
    \begin{subfigure}[t]{0.24\textwidth}
        \centering
        \includegraphics[width=\linewidth]{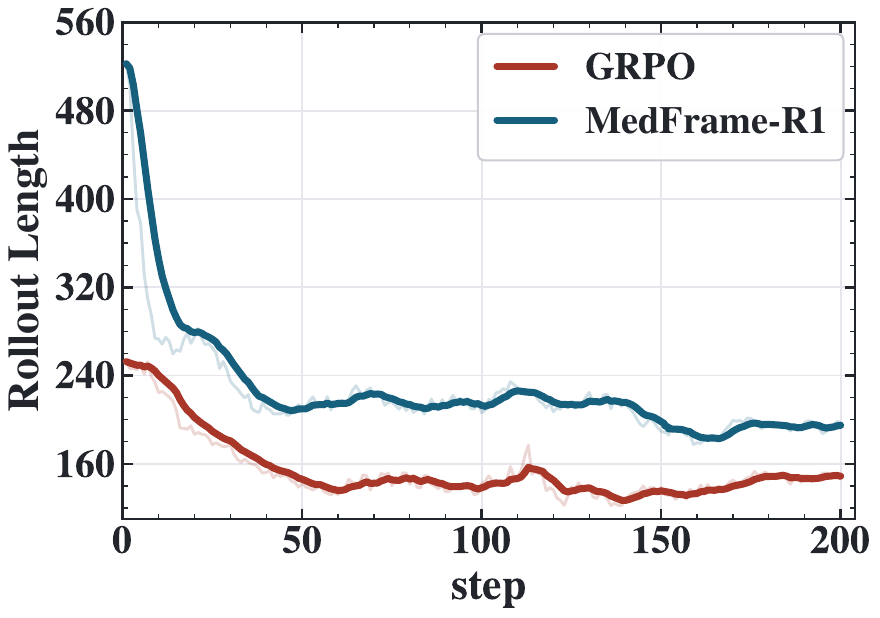}
        \caption{\textbf{Rollout length.}
        Rollouts become shorter during training.}
        \label{fig:response_length}
    \end{subfigure}

    \caption{
    \textbf{Reward dynamics and training behaviour during RL.}
    (a) Without embedding gating, judge rewards increase despite
    (b) declining semantic agreement with reference answers.
    (c) Informative sample selection maintains higher training rewards
    than the full data pool.
    (d) Response lengths decrease as training progresses.
    }
    \label{fig:training_signal_reasoning}
\end{figure*}

\section{Experiments}
\label{sec:experiments}

We evaluate MedFrame-R1 on 6 medical visual question answering (VQA) and 11 question answering (QA) benchmarks. We first describe the evaluation protocol and then compare MedFrame-R1 with existing models.

\subsection{Evaluation Protocol}
\label{sec:evaluation_protocol}

We evaluate all models on 19,200 items across 17 medical benchmarks covering language and vision tasks. All models use the same inputs and greedy decoding with their respective chat templates, without additional system prompts or formatting instructions. For responses containing thinking blocks, only the text after the final closing thinking tag is scored.
To improve evaluation reliability, we first apply deterministic scoring wherever correctness can be established without model-based judgement. For multiple-choice questions, explicit option letters take priority over restricted option-text matching. For binary questions, the scorer handles negative statements, prioritises final answers, and rejects conflicting answers, while eligible short-answer questions are assessed through restricted answer matching. Together, these rules resolve approximately 70\% of evaluation items, while unresolved responses and all reports are evaluated by LLM judges.

Unlike previous protocols that rely on a single judge~\citep{mullappilly2026medix}, we use three independent judges from the InternVL~\citep{wang2025internvl3}, Qwen~\citep{qwen38}, and MedGemma~\citep{sellergren2025medgemma} families to reduce dependence on individual evaluators and potential self-preference bias~\citep{panickssery2024llm}.
Each judge independently evaluates the question, reference, and processed answer without access to other judges' decisions~\citep{kapetanovic2026anchoringbiasllmasajudgesystems}. For Qwen3-VL models, we retain Qwen3.8 as a judge because its hybrid architecture differs from the Qwen3-VL backbone. For Qwen3.5, we replace Qwen3.8 with gpt-oss-20b~\citep{agarwal2025gpt} due to their architectural similarities, reducing potential self-preference bias. The judges' decisions are combined through majority voting to determine correctness, while report generation and summarisation are assessed using a clinical rubric from 0 to 5, with the median score normalised for reporting.
We report performance across individual tasks, macro and micro averages, deterministic scoring coverage, and agreement among judges. Further details on the evaluation procedure, scoring rules, and judge configurations are provided in Supplementary \cref{sec:protocol-final}.

\subsection{Main Results}
\label{sec:main_results}

\Cref{tab:main} compares MedFrame-R1 with existing medical multimodal models across 17 benchmarks. With Qwen3-VL, MedFrame-R1 achieves macro averages of 64.8 and 66.8 at 4B and 8B, respectively, outperforming MediX-R1 under the same backbone and training data by up to 1.8 macro and 3.6 micro points. The strongest improvements arise in VQA, with gains of 5.1 and 8.1 points over the base models at 4B and 8B, respectively. Notably, these improvements extend across multiple imaging benchmarks, including PathVQA, SLAKE, and VQA-RAD, rather than being concentrated on a single dataset. Despite training exclusively on VQA tasks without KL regularisation, MedFrame-R1 also preserves and slightly improves performance on unseen medical QA benchmarks, suggesting that the additional exploration does not compromise general medical reasoning.

\paragraph{Scaling to Qwen3.5}
To assess whether these improvements extend to stronger backbones, we apply MedFrame-R1 directly to Qwen3.5 4B and 9B without supervised cold start (Supplementary \cref{sec:app-qwen35}). Despite the substantially stronger initial performance of these models, RL alone delivers improvements of up to 3.7 macro and 5.6 micro points over the released checkpoints. The largest gains again occur in VQA, particularly on PathVQA and SLAKE, while performance on medical QA tasks also improves. Averaged over the two sizes, the gains over zero-shot Qwen3.5 are 3.0 macro and 5.2 micro points, and MedFrame-R1 achieves the highest overall results in both model-size groups in \cref{tab:main}. These findings demonstrate that the benefits of our framework persist across model generations and do not depend on supervised cold-start training.

\section{Ablation Studies}
\label{sec:ablation_studies}

We ablate the training components of MedFrame-R1, the threshold of the embedding gate, the threshold of the length reward, the trainable pool, and KL regularisation.

\begin{table}[t]

\caption{\textbf{Component ablation on VQA benchmarks.}
Comparison of the supervised cold start, RL optimisation, the embedding gate, and the length reward for the 4B and 8B models. MediX-R1$^{\dagger}$ denotes our reproduction. Bold and underlined values indicate the best and second-best results within each model size. Macro and Micro are the averages over benchmarks and over samples. Full results are provided in Supplementary \cref{app:ablation_full}.}

\label{tab:ablation}

\centering
\small
\setlength{\tabcolsep}{5.5pt}
\renewcommand{\arraystretch}{1.3}
\setlength{\aboverulesep}{0pt}\setlength{\belowrulesep}{0pt}
\resizebox{\columnwidth}{!}{%
\begin{tabular}{l|cccccc|cc}
\toprule
\multicolumn{1}{c|}{\rule{0pt}{1.9cm}} & \multicolumn{6}{c|}{} & \multicolumn{2}{c}{} \\[-2.07cm]
\multicolumn{1}{c|}{\rule{0pt}{1.9cm}\raisebox{0.78cm}[0pt][0pt]{\large\textbf{Method}}}
& \multicolumn{1}{c}{\raisebox{0.78cm}[0pt][0pt]{\makebox[0pt][c]{\rotatebox[origin=c]{65}{\footnotesize\textbf{SLAKE}}}}}
& \multicolumn{1}{c}{\raisebox{0.78cm}[0pt][0pt]{\makebox[0pt][c]{\rotatebox[origin=c]{65}{\footnotesize\textbf{VQA-RAD}}}}}
& \multicolumn{1}{c}{\raisebox{0.78cm}[0pt][0pt]{\makebox[0pt][c]{\rotatebox[origin=c]{65}{\footnotesize\textbf{PathVQA}}}}}
& \multicolumn{1}{c}{\raisebox{0.78cm}[0pt][0pt]{\makebox[0pt][c]{\rotatebox[origin=c]{65}{\footnotesize\textbf{PMC-VQA}}}}}
& \multicolumn{1}{c}{\raisebox{0.78cm}[0pt][0pt]{\makebox[0pt][c]{\rotatebox[origin=c]{65}{\footnotesize\textbf{PMC-VQA-H}}}}}
& \multicolumn{1}{c|}{\raisebox{0.78cm}[0pt][0pt]{\makebox[0pt][c]{\rotatebox[origin=c]{65}{\footnotesize\textbf{MIMIC-Gen}}}}}
& \multicolumn{1}{c}{\raisebox{0.78cm}[0pt][0pt]{\makebox[0pt][c]{\rotatebox[origin=c]{65}{\footnotesize\textbf{Macro}}}}}
& \multicolumn{1}{c}{\raisebox{0.78cm}[0pt][0pt]{\makebox[0pt][c]{\rotatebox[origin=c]{65}{\footnotesize\textbf{Micro}}}}} \\
\midrule
\rowcolor{cSub}
\textbf{Qwen3-VL-4B} (zero-shot)
& 67.9 & 54.3 & 40.0 & 53.0 & 34.2 & 31.6 & 46.8 & 44.5 \\
\hspace{1.2em}MediX-R1$^{\dagger}$
& 73.7 & 58.3 & 44.6 & \underline{54.8} & 34.3 & 29.4 & 49.2 & 48.1 \\
\hspace{1.2em}DAPO (judge only)
& 74.2 & \underline{58.5} & 48.3 & 54.2 & \underline{35.7} & \underline{33.0} & 50.6 & 50.4 \\
\hspace{1.2em}DAPO + gate
& \underline{74.3} & 57.4 & \underline{50.6} & \textbf{55.9} & 35.4 & \textbf{33.4} & \underline{51.2} & \underline{52.0} \\
\hspace{1.2em}wCE + AESL SFT cold start
& 73.9 & 54.8 & 49.5 & 50.4 & 32.0 & \textbf{33.4} & 49.0 & 49.8 \\
\midrule
\rowcolor{cOurs}
\hspace{2.4em}\textbf{MedFrame-R1-4B}
& \textbf{74.4} & \textbf{61.6} & \textbf{52.1} & 54.7 & \textbf{36.0} & 32.4 & \textbf{51.9} & \textbf{52.8} \\
\midrule
\rowcolor{cSub}
\textbf{Qwen3-VL-8B} (zero-shot)
& 61.6 & 52.8 & 36.0 & 53.2 & 34.0 & \textbf{34.4} & 45.3 & 41.6 \\
\hspace{1.2em}MediX-R1$^{\dagger}$
& 70.2 & 59.9 & 46.6 & \underline{53.8} & \underline{34.7} & 30.6 & 49.3 & 48.9 \\
\hspace{1.2em}GRPO
& 69.3 & 56.1 & 48.6 & \underline{53.8} & 33.8 & 20.4 & 47.0 & 49.6 \\
\hspace{1.2em}wCE + AESL SFT cold start
& 72.3 & 60.5 & \underline{52.1} & 50.6 & \underline{34.7} & 28.0 & 49.7 & \underline{51.7} \\
\hspace{2.4em}MedFrame-R1 (no length reward)
& \underline{74.3} & \underline{61.2} & 51.7 & 48.2 & 33.4 & \underline{33.6} & \underline{50.4} & 51.2 \\
\midrule
\rowcolor{cOurs}
\hspace{2.4em}\textbf{MedFrame-R1-8B} (+ length reward)
& \textbf{76.5} & \textbf{65.6} & \textbf{53.8} & \textbf{56.2} & \textbf{34.8} & 33.4 & \textbf{53.4} & \textbf{54.3} \\
\bottomrule
\end{tabular}}
\end{table}

\subsection{Component Ablation}
\label{sec:component_ablation}

\begin{figure}[h]
    \centering

    \begin{subfigure}[t]{0.48\columnwidth}
        \centering
        \includegraphics[width=\linewidth]{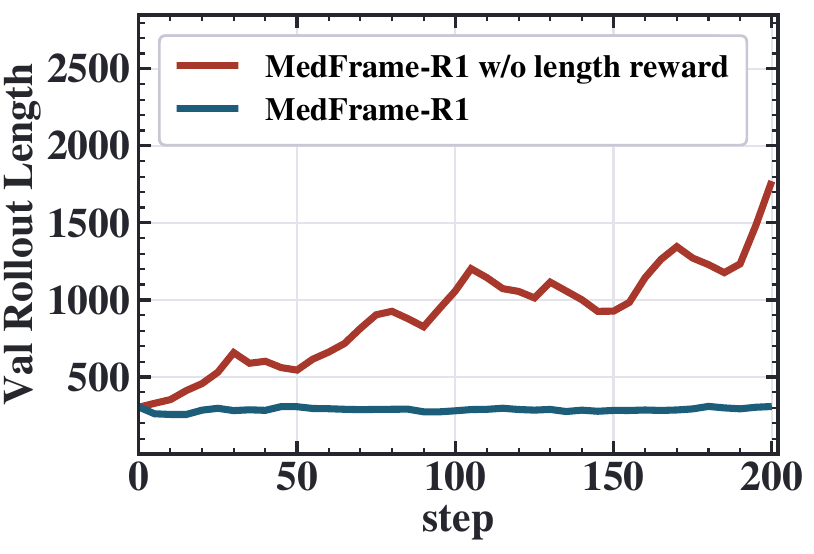}
        \caption{\textbf{The length reward stabilises rollout length during RL training.}
        }
        \label{fig:response_length_lw_val}
    \end{subfigure}
    \hfill
    \begin{subfigure}[t]{0.48\columnwidth}
        \centering
        \includegraphics[width=\linewidth]{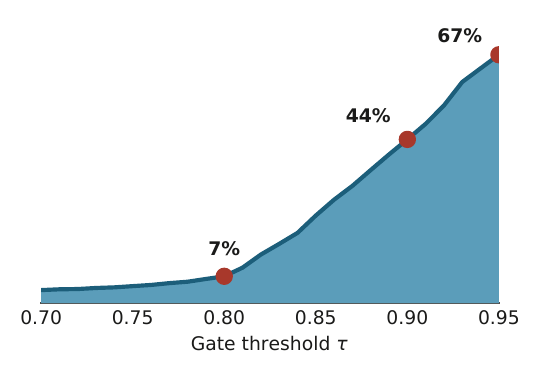}
        \caption{\textbf{Effect of gate threshold.}
        Higher thresholds filter more correct rollouts.}
        \label{fig:gate_threshold_area}
    \end{subfigure}

    \captionsetup{font=footnotesize}
    \caption{\textbf{Reward and Length Control.}
    (a) The length reward helps control excessive rollout length during RL training. 
    (b) Increasing the embedding gate threshold filters a larger fraction of correct rollouts.}
    \label{fig:lw_and_gate}
\end{figure}

\Cref{tab:ablation} examines the contribution of different training components to VQA performance. For the 4B model, applying DAPO with LLM judge rewards improves the macro average from 46.8 to 50.6, demonstrating the benefit of selecting informative rollout groups with different correctness outcomes. However, relying on the judge alone can encourage reward hacking. As shown in \cref{fig:judge_accuracy,fig:embedding_accuracy}, the judge reward continues to increase while embedding-based semantic accuracy declines, suggesting that the policy learns to satisfy the judge without improving answer correctness. Adding the embedding gate reduces this issue by requiring semantic agreement before accepting positive judge rewards, improving the macro and micro averages to 51.2 and 52.0, respectively. Combined with supervised initialisation, the final MedFrame-R1-4B achieves the highest macro average of 51.9. Further qualitative examples of reward hacking are provided in Supplementary \cref{sec:app-reward-hacking}.
For the 8B model, the supervised cold start improves the macro average from 45.3 to 49.7, providing a stronger initialisation for RL. While the 4B model maintains shorter rollouts during training (\cref{fig:response_length}), the 8B model exhibits excessive length and repetition. We therefore apply a length reward~\citep{li2026drpo}, which favours shorter correct rollouts and penalises longer incorrect ones. As shown in \cref{fig:response_length_lw_val}, this stabilises rollout length and improves the macro and micro averages from 50.4 to 53.4 and from 51.2 to 54.3, respectively. These results demonstrate the importance of both reliable correctness rewards and controlled rollout generation during RL.

\begin{table}[h]

\caption{\textbf{Effect of embedding similarity thresholds on VQA performance.} We compare rule-based verification with embedding gates at different thresholds $\tau$, where $\tau{=}0.80$ is the final MedFrame-R1-4B. Bold and underlined values indicate the best and second-best results. Macro and Micro denote benchmark and sample averages. Full results are in Supplementary \cref{app:gate_full}.}

\label{tab:gate}

\centering
\small
\setlength{\tabcolsep}{5.5pt}
\renewcommand{\arraystretch}{1.3}
\setlength{\aboverulesep}{0pt}\setlength{\belowrulesep}{0pt}
\resizebox{\columnwidth}{!}{%
\begin{tabular}{l|cccccc|cc}
\toprule
\multicolumn{1}{c|}{\rule{0pt}{1.9cm}} & \multicolumn{6}{c|}{} & \multicolumn{2}{c}{} \\[-2.07cm]
\multicolumn{1}{c|}{\rule{0pt}{1.9cm}\raisebox{0.78cm}[0pt][0pt]{\large\textbf{Method}}}
& \multicolumn{1}{c}{\raisebox{0.78cm}[0pt][0pt]{\makebox[0pt][c]{\rotatebox[origin=c]{65}{\footnotesize\textbf{SLAKE}}}}}
& \multicolumn{1}{c}{\raisebox{0.78cm}[0pt][0pt]{\makebox[0pt][c]{\rotatebox[origin=c]{65}{\footnotesize\textbf{VQA-RAD}}}}}
& \multicolumn{1}{c}{\raisebox{0.78cm}[0pt][0pt]{\makebox[0pt][c]{\rotatebox[origin=c]{65}{\footnotesize\textbf{PathVQA}}}}}
& \multicolumn{1}{c}{\raisebox{0.78cm}[0pt][0pt]{\makebox[0pt][c]{\rotatebox[origin=c]{65}{\footnotesize\textbf{PMC-VQA}}}}}
& \multicolumn{1}{c}{\raisebox{0.78cm}[0pt][0pt]{\makebox[0pt][c]{\rotatebox[origin=c]{65}{\footnotesize\textbf{PMC-VQA-H}}}}}
& \multicolumn{1}{c|}{\raisebox{0.78cm}[0pt][0pt]{\makebox[0pt][c]{\rotatebox[origin=c]{65}{\footnotesize\textbf{MIMIC-Gen}}}}}
& \multicolumn{1}{c}{\raisebox{0.78cm}[0pt][0pt]{\makebox[0pt][c]{\rotatebox[origin=c]{65}{\footnotesize\textbf{Macro}}}}}
& \multicolumn{1}{c}{\raisebox{0.78cm}[0pt][0pt]{\makebox[0pt][c]{\rotatebox[origin=c]{65}{\footnotesize\textbf{Micro}}}}} \\
\midrule

Rule-based verification
& 74.5 & 58.0 & 48.1 & 54.7 & 35.9 & 32.7 & 50.7 & 50.4 \\
$\tau{=}0.70$
& 73.9 & 55.4 & 46.5 & \textbf{55.1} & 35.4 & \textbf{34.2} & 50.1 & 49.4 \\
$\tau{=}0.75$
& \underline{74.6} & \underline{58.1} & 49.4 & 54.7 & 35.5 & \underline{32.8} & 50.8 & 51.1 \\
$\tau{=}0.78$
& \textbf{75.1} & 56.8 & \underline{49.8} & \underline{55.0} & \textbf{37.0} & 32.6 & \underline{51.0} & \underline{51.6} \\
\midrule
\rowcolor{cOurs}
$\tau{=}0.80$ \textbf{(MedFrame-R1-4B)}
& 74.4 & \textbf{61.6} & \textbf{52.1} & 54.7 & \underline{36.0} & 32.4 & \textbf{51.9} & \textbf{52.8} \\
\bottomrule
\end{tabular}}

\end{table}

\subsection{Effect of Embedding Gate Threshold}
\label{sec:gate_threshold}

We further investigate how the embedding similarity threshold affects reward reliability and training performance. As shown in \cref{tab:gate}, increasing the threshold generally improves VQA performance, with the selected threshold of $\tau=0.80$ achieving the highest macro and micro averages. We also compare against rule-based verification, which penalises rollouts listing multiple candidate answers. Although this approach reduces reward hacking, it performs worse than the embedding gate at $\tau=0.80$, suggesting that semantic agreement provides a more effective learning signal than fixed rules.

However, a stricter threshold does not necessarily provide a better learning signal, as it may reject correct rollouts that could contribute to RL training. \Cref{fig:gate_threshold_area} illustrates this trade-off. At $\tau=0.80$, only 7\% of correct rollouts are filtered, while increasing the threshold to 0.90 and 0.95 raises this percentage to 44\% and 67\%, respectively. Such aggressive filtering reduces useful positive rewards and may limit the informative rollouts available for optimisation. We therefore use $\tau=0.80$ to balance reward reliability and the preservation of correct rollouts, reducing reward hacking without unnecessarily restricting the training signal.

\begin{table}[h]

\caption{\textbf{Effect of length threshold $T$ on VQA.}
Qwen3.5 zero-shot and trained with $T\in\{500,1500\}$. Tokens denotes mean response length. Rep. is the percentage of responses repeating an eight-word phrase at least three times. Bold and underline indicate best and second-best. Macro/Micro are benchmark/sample averages. Full results in Supplementary \cref{sec:app-qwen35-length}.}
\label{tab:length_thresholds}

\centering
\small
\setlength{\tabcolsep}{5.5pt}
\renewcommand{\arraystretch}{1.3}
\setlength{\aboverulesep}{0pt}\setlength{\belowrulesep}{0pt}
\resizebox{\columnwidth}{!}{%
\begin{tabular}{l|cc|cccccc|cc}
\toprule
\multicolumn{1}{c|}{\rule{0pt}{1.9cm}} & \multicolumn{2}{c|}{} & \multicolumn{6}{c|}{} & \multicolumn{2}{c}{} \\[-2.07cm]
\multicolumn{1}{c|}{\rule{0pt}{1.9cm}\raisebox{0.78cm}[0pt][0pt]{\large\textbf{Method}}}
& \multicolumn{1}{c}{\raisebox{0.78cm}[0pt][0pt]{\makebox[0pt][c]{\rotatebox[origin=c]{65}{\footnotesize\textbf{Tokens}}}}}
& \multicolumn{1}{c|}{\raisebox{0.78cm}[0pt][0pt]{\makebox[0pt][c]{\rotatebox[origin=c]{65}{\footnotesize\textbf{Rep.}}}}}
& \multicolumn{1}{c}{\raisebox{0.78cm}[0pt][0pt]{\makebox[0pt][c]{\rotatebox[origin=c]{65}{\footnotesize\textbf{SLAKE}}}}}
& \multicolumn{1}{c}{\raisebox{0.78cm}[0pt][0pt]{\makebox[0pt][c]{\rotatebox[origin=c]{65}{\footnotesize\textbf{VQA-RAD}}}}}
& \multicolumn{1}{c}{\raisebox{0.78cm}[0pt][0pt]{\makebox[0pt][c]{\rotatebox[origin=c]{65}{\footnotesize\textbf{PathVQA}}}}}
& \multicolumn{1}{c}{\raisebox{0.78cm}[0pt][0pt]{\makebox[0pt][c]{\rotatebox[origin=c]{65}{\footnotesize\textbf{PMC-VQA}}}}}
& \multicolumn{1}{c}{\raisebox{0.78cm}[0pt][0pt]{\makebox[0pt][c]{\rotatebox[origin=c]{65}{\footnotesize\textbf{PMC-VQA-H}}}}}
& \multicolumn{1}{c|}{\raisebox{0.78cm}[0pt][0pt]{\makebox[0pt][c]{\rotatebox[origin=c]{65}{\footnotesize\textbf{MIMIC-Gen}}}}}
& \multicolumn{1}{c}{\raisebox{0.78cm}[0pt][0pt]{\makebox[0pt][c]{\rotatebox[origin=c]{65}{\footnotesize\textbf{Macro}}}}}
& \multicolumn{1}{c}{\raisebox{0.78cm}[0pt][0pt]{\makebox[0pt][c]{\rotatebox[origin=c]{65}{\footnotesize\textbf{Micro}}}}} \\
\midrule
\rowcolor{cSub}
\textbf{Qwen3.5-4B} (zero-shot)
& 2{,}296 & 53.6 & 69.7 & 60.5 & 46.4 & 60.2 & 38.9 & 47.8 & 53.9 & 50.5 \\
\hspace{1.2em}$T{=}1500$
& 773 & 3.4 & \underline{79.1} & \underline{64.5} & \underline{56.3} & \textbf{62.1} & \textbf{42.9} & \underline{48.6} & \underline{58.9} & \underline{58.0} \\
\rowcolor{cOurs}
\hspace{1.2em}\textbf{$T{=}500$ (MedFrame-R1)}
& 441 & 1.7 & \textbf{81.4} & \textbf{64.7} & \textbf{56.4} & \underline{61.3} & \underline{42.3} & \textbf{51.0} & \textbf{59.5} & \textbf{58.1} \\
\midrule
\rowcolor{cSub}
\textbf{Qwen3.5-9B} (zero-shot)
& 2{,}247 & 46.6 & 70.7 & 65.6 & 48.7 & 61.1 & 41.4 & 49.2 & 56.1 & 52.6 \\
\hspace{1.2em}$T{=}1500$
& 749 & 2.9 & \underline{79.3} & \underline{66.3} & \underline{55.1} & \underline{63.9} & \textbf{44.2} & \underline{50.8} & \underline{59.9} & \underline{57.9} \\
\rowcolor{cOurs}
\hspace{1.2em}\textbf{$T{=}500$ (MedFrame-R1)}
& 359 & 0.7 & \textbf{80.2} & \textbf{71.4} & \textbf{56.6} & \textbf{64.9} & \underline{43.7} & \textbf{51.6} & \textbf{61.4} & \textbf{59.1} \\
\bottomrule
\end{tabular}}
\end{table}

\subsection{Effect of the Length Threshold}
\label{sec:length_threshold}

\begin{figure}[t]
    \centering
    \includegraphics[width=\columnwidth]{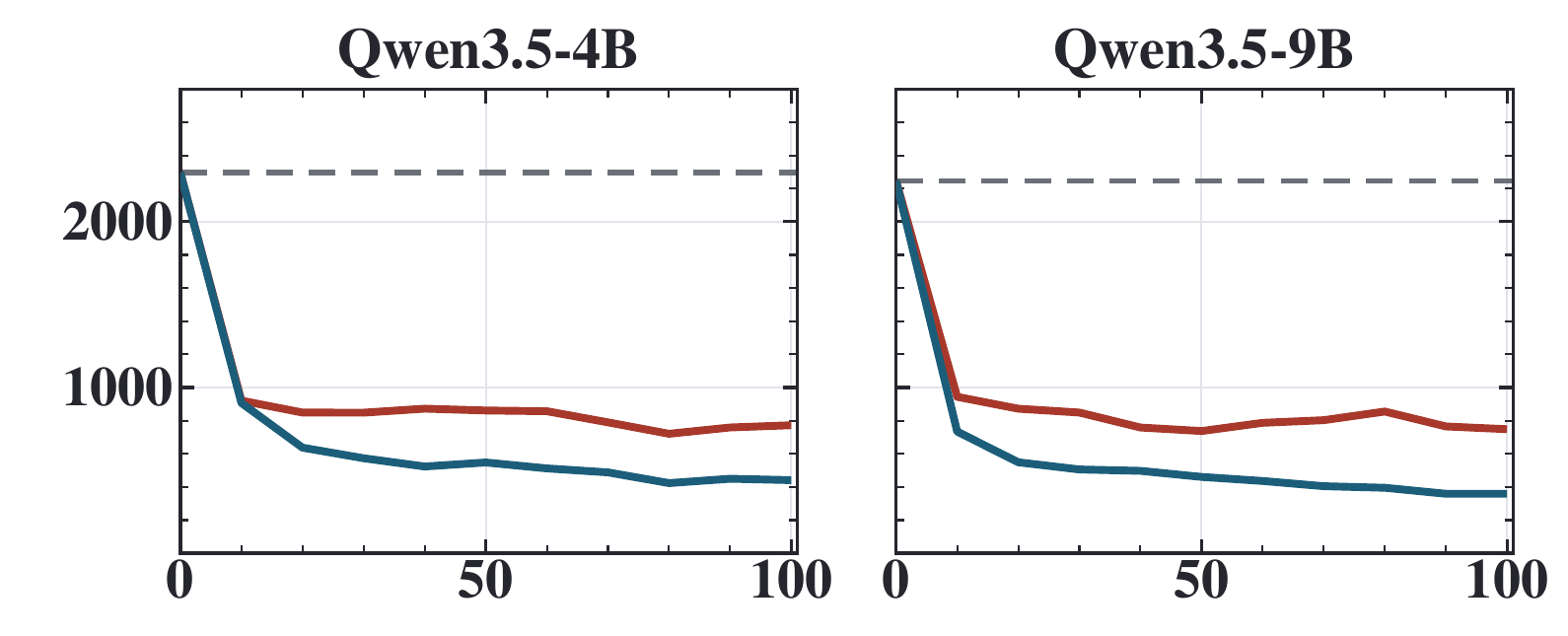}
    \captionsetup{font=footnotesize}
    \setlength{\fboxsep}{1.5pt}
    \caption{\textbf{Rollout length during RL.} Validation rollout length over RL steps for the \colorbox[RGB]{214,216,219}{\strut zero-shot (dashed)}, \colorbox[RGB]{243,199,193}{\strut $T=1500$} and \colorbox[RGB]{189,214,226}{\strut $T=500$}.}
    \label{fig:lw_threshold}
\end{figure}

The length threshold $T$ determines how many reasoning tokens are allowed without penalty. As shown in \cref{tab:length_thresholds}, zero-shot Qwen3.5 produces lengthy, repetitive responses, averaging over 2,200 tokens with repetition rates of 53.6\% (4B) and 46.6\% (9B). In contrast, applying a length threshold ($T=500,1500$) substantially reduces both length and repetition within the first ten steps (\cref{fig:lw_threshold}). At $T=500$, responses are $>\!80\%$ shorter, with repetition rates below 2\%, while maintaining the same accuracy. Increasing the length threshold to $T=1500$ yields minor improvements over $T=500$ on some benchmarks but is negligible overall (See Tab. 5), indicating that concise responses are more effective. We therefore adopt $T=500$, which also accelerates training by approximately 17\%.

\subsection{Effect of the Trainable Pool}
\label{sec:selection_analysis}

Beyond reward reliability, we examine how the trainable pool affects training performance and efficiency. As shown in \cref{tab:trainable_4b}, training MedFrame-R1 on the 26K trainable pool improves the VQA macro average from 49.7 to 51.9 and the micro average from 51.7 to 52.8 compared with training on the full 51K dataset. \Cref{fig:trainable_reward} further shows that the trainable pool maintains higher training rewards, reaching approximately 0.74 compared with 0.60 for the full corpus.
These improvements arise from focusing training on samples that previously exhibited different correctness outcomes, reducing repeated sampling of questions that provide limited learning signals. Consequently, training time for 200 steps decreases from approximately 56 to 30 hours, improving efficiency while achieving better VQA performance. However, this selection may exclude difficult or rare questions that the model cannot yet answer correctly. We therefore use a supervised cold start on the full dataset before RL to provide learning opportunities for these examples. Further analysis of the trainable pool, training efficiency, and complete benchmark results is provided in Supplementary \cref{sec:app-trainable-data}.

\begin{table}[t]

\caption{\textbf{Effect of the trainable pool on VQA.}
Comparison of the full data and the trainable pool. Macro and Micro denote benchmark and sample averages. Full results are provided in Supplementary \cref{sec:app-trainable-data}.}
\label{tab:trainable_4b}

\centering
\small
\setlength{\tabcolsep}{5.5pt}
\renewcommand{\arraystretch}{1.3}
\setlength{\aboverulesep}{0pt}
\setlength{\belowrulesep}{0pt}
\resizebox{\columnwidth}{!}{%
\begin{tabular}{l|cccccc|cc}
\toprule
\multicolumn{1}{c|}{\rule{0pt}{1.9cm}} & \multicolumn{6}{c|}{} & \multicolumn{2}{c}{} \\[-2.07cm]
\multicolumn{1}{c|}{\rule{0pt}{1.9cm}\raisebox{0.78cm}[0pt][0pt]{\large\textbf{Method}}}
& \multicolumn{1}{c}{\raisebox{0.78cm}[0pt][0pt]{\makebox[0pt][c]{\rotatebox[origin=c]{65}{\footnotesize\textbf{SLAKE}}}}}
& \multicolumn{1}{c}{\raisebox{0.78cm}[0pt][0pt]{\makebox[0pt][c]{\rotatebox[origin=c]{65}{\footnotesize\textbf{VQA-RAD}}}}}
& \multicolumn{1}{c}{\raisebox{0.78cm}[0pt][0pt]{\makebox[0pt][c]{\rotatebox[origin=c]{65}{\footnotesize\textbf{PathVQA}}}}}
& \multicolumn{1}{c}{\raisebox{0.78cm}[0pt][0pt]{\makebox[0pt][c]{\rotatebox[origin=c]{65}{\footnotesize\textbf{PMC-VQA}}}}}
& \multicolumn{1}{c}{\raisebox{0.78cm}[0pt][0pt]{\makebox[0pt][c]{\rotatebox[origin=c]{65}{\footnotesize\textbf{PMC-VQA-H}}}}}
& \multicolumn{1}{c|}{\raisebox{0.78cm}[0pt][0pt]{\makebox[0pt][c]{\rotatebox[origin=c]{65}{\footnotesize\textbf{MIMIC-Gen}}}}}
& \multicolumn{1}{c}{\raisebox{0.78cm}[0pt][0pt]{\makebox[0pt][c]{\rotatebox[origin=c]{65}{\footnotesize\textbf{Macro}}}}}
& \multicolumn{1}{c}{\raisebox{0.78cm}[0pt][0pt]{\makebox[0pt][c]{\rotatebox[origin=c]{65}{\footnotesize\textbf{Micro}}}}} \\
\midrule
\textbf{MedFrame-R1-4B} (Full 51K)
& 73.3 & 56.3 & 51.1 & 54.6 & 34.2 & 28.8 & 49.7 & 51.7 \\

\rowcolor{medframe}
\textbf{MedFrame-R1-4B} (Trainable 26K)
& \textbf{74.4} & \textbf{61.6} & \textbf{52.1} & \textbf{54.7} & \textbf{36.0} & \textbf{32.4} & \textbf{51.9} & \textbf{52.8} \\
\bottomrule
\end{tabular}}
\end{table}

\begin{table}[h]

\caption{
\textbf{Comparison of DAPO, MediX-R1, and MedFrame-R1 with and without KL regularisation on Qwen3-VL-8B.} Bold and underlined values indicate the best and second-best results. Full results are provided in Supplementary \cref{sec:app-kl-ablation}.}
\label{tab:kl_ablation}

\centering
\small
\setlength{\tabcolsep}{5.5pt}
\renewcommand{\arraystretch}{1.3}
\setlength{\aboverulesep}{0pt}
\setlength{\belowrulesep}{0pt}
\resizebox{\columnwidth}{!}{%
\begin{tabular}{l|cccccc|cc}
\toprule
\multicolumn{1}{c|}{\rule{0pt}{1.9cm}} & \multicolumn{6}{c|}{} & \multicolumn{2}{c}{} \\[-2.07cm]
\multicolumn{1}{c|}{\rule{0pt}{1.9cm}\raisebox{0.78cm}[0pt][0pt]{\large\textbf{Method}}}
& \multicolumn{1}{c}{\raisebox{0.78cm}[0pt][0pt]{\makebox[0pt][c]{\rotatebox[origin=c]{65}{\footnotesize\textbf{SLAKE}}}}}
& \multicolumn{1}{c}{\raisebox{0.78cm}[0pt][0pt]{\makebox[0pt][c]{\rotatebox[origin=c]{65}{\footnotesize\textbf{VQA-RAD}}}}}
& \multicolumn{1}{c}{\raisebox{0.78cm}[0pt][0pt]{\makebox[0pt][c]{\rotatebox[origin=c]{65}{\footnotesize\textbf{PathVQA}}}}}
& \multicolumn{1}{c}{\raisebox{0.78cm}[0pt][0pt]{\makebox[0pt][c]{\rotatebox[origin=c]{65}{\footnotesize\textbf{PMC-VQA}}}}}
& \multicolumn{1}{c}{\raisebox{0.78cm}[0pt][0pt]{\makebox[0pt][c]{\rotatebox[origin=c]{65}{\footnotesize\textbf{PMC-VQA-H}}}}}
& \multicolumn{1}{c|}{\raisebox{0.78cm}[0pt][0pt]{\makebox[0pt][c]{\rotatebox[origin=c]{65}{\footnotesize\textbf{MIMIC-Gen}}}}}
& \multicolumn{1}{c}{\raisebox{0.78cm}[0pt][0pt]{\makebox[0pt][c]{\rotatebox[origin=c]{65}{\footnotesize\textbf{Macro}}}}}
& \multicolumn{1}{c}{\raisebox{0.78cm}[0pt][0pt]{\makebox[0pt][c]{\rotatebox[origin=c]{65}{\footnotesize\textbf{Micro}}}}} \\
\midrule
DAPO (No KL)
& 71.9 & \underline{63.6} & 47.8 & \underline{54.0} & \textbf{35.2} & \textbf{34.4} & 51.2 & 50.0 \\
MediX-R1 (DAPO + KL)
& 70.2 & 59.9 & 46.6 & 53.8 & 34.7 & 30.6 & 49.3 & 48.9 \\

MedFrame-R1 (+ KL)
& \textbf{78.1} & 61.4 & \underline{53.6} & 53.6 & \underline{35.0} & 30.0 & \underline{52.0} & \underline{53.7} \\
\rowcolor{medframe}
\textbf{MedFrame-R1 (No KL)}
& \underline{76.5} & \textbf{65.6} & \textbf{53.8} & \textbf{56.2} & 34.8 & \underline{33.4} & \textbf{53.4} & \textbf{54.3} \\
\bottomrule
\end{tabular}}
\end{table}

\subsection{Effect of KL Regularisation}
\label{sec:kl_ablation}

KL regularisation introduces a trade-off between preserving previously learned knowledge and allowing the policy to explore new solutions during RL. While this constraint may help maintain performance on QA tasks excluded from training, it can also limit improvements in visual reasoning. As shown in \cref{tab:kl_ablation}, removing KL regularisation raises the VQA macro average from 49.3 to 51.2 for DAPO and from 52.0 to 53.4 for MedFrame-R1, although MedFrame-R1 drops slightly on SLAKE and PMC-VQA-H.
Despite the greater freedom during optimisation, both MedFrame-R1 variants maintain QA performance above the initial checkpoint, suggesting that removing KL does not cause substantial forgetting on the evaluated benchmarks. These findings support training without KL regularisation to encourage exploration while preserving broader medical knowledge. Further comparisons and complete benchmark results are provided in Supplementary \cref{sec:app-kl-ablation}.



\section{Conclusion}
\label{sec:conclusion}

We introduced MedFrame-R1, an open-source framework designed to improve the reliability and reproducibility of medical MLLMs. We identified four fundamental limitations in existing approaches: uninformative reward groups, reward exploitation, insufficient domain knowledge in the starting model, and evaluation bias. MedFrame-R1 systematically addresses these limitations through dynamic sampling, embedding-gated rewards with length control, supervised cold-start reasoning, and an evaluation protocol combining deterministic scoring with three independent judges. Our analyses demonstrate that higher judge rewards do not necessarily reflect better answers, highlighting the importance of reward verification. Across different model families, MedFrame-R1 improves medical benchmark performance while supporting more consistent training and evaluation. By integrating these components into an open-source framework, we provide a transparent and reproducible foundation for future research on medical multimodal RL.

\paragraph{Limitations and Future Work}
Despite these improvements, MedFrame-R1 has several limitations. Dynamic sampling requires generating and evaluating rollouts that may subsequently be discarded, increasing computational overhead when few groups contain informative reward variation. Moreover, although supervised cold-start training covers the full dataset, only approximately 26K of the 51K examples provide informative groups during RL. Future work could explore hint-based approaches~\citep{zhang2026scaf, amani2506rl, qu2601pope} or methods that learn from groups with identical rewards~\citep{le2026no} to exploit otherwise uninformative examples. Constructing datasets with a higher proportion of informative samples may also improve training efficiency, although trainability depends on the starting model and evolves during optimisation.

\section*{LLM Usage Statement}
Claude Opus 5.5 (Anthropic) was used in the preparation of this manuscript solely to assist with language editing,
including improving wording, grammar, clarity and readability, as well as providing support with manuscript
formatting. It was not used to generate or validate the scientific contributions, design experiments, analyse
experimental results or draw scientific conclusions. All technical content, methodological decisions, experimental
design, implementation, analysis and conclusions were produced and verified by the authors, who take full
responsibility for the content of this work.

{
    \small
    \addtolength{\bibsep}{-0.5pt}
    \bibliographystyle{ieeenat_fullname}
    \bibliography{main}
}

\clearpage
\setcounter{section}{0}\setcounter{table}{0}\setcounter{figure}{0}\setcounter{equation}{0}
\renewcommand{\theHsection}{S\arabic{section}}
\renewcommand{\theHsubsection}{S\arabic{section}.\arabic{subsection}}
\renewcommand{\theHsubsubsection}{S\arabic{section}.\arabic{subsection}.\arabic{subsubsection}}
\renewcommand{\theHtable}{S\arabic{table}}
\renewcommand{\theHfigure}{S\arabic{figure}}
\renewcommand{\theHequation}{S\arabic{equation}}
\renewcommand{\thesection}{S\arabic{section}}
\renewcommand{\thetable}{S\arabic{table}}
\renewcommand{\thefigure}{S\arabic{figure}}
\renewcommand{\theequation}{S\arabic{equation}}

\maketitlesupplementary

\renewcommand{\dbltopfraction}{0.85}
\renewcommand{\dblfloatpagefraction}{0.75}
\renewcommand{\topfraction}{0.85}
\renewcommand{\floatpagefraction}{0.75}
\renewcommand{\textfraction}{0.12}
\setcounter{dbltopnumber}{2}
\setcounter{topnumber}{3}
\setcounter{totalnumber}{5}

This supplementary material provides the implementation details, evaluation protocol, complete results, and qualitative examples that support the main paper.

\section{Implementation Details and Configuration}
\label{sec:app-configs}

This section provides the implementation details and hyperparameter configurations of MedFrame-R1, complementing the methodological description in the main paper. \Cref{tab:app-configs,tab:app-configs-rl} summarise the settings used for supervised cold-start training, reward computation, reinforcement learning, length regularisation, and evaluation, as recorded in the final run configurations.

\begin{table}[h]
\centering
\sbox{\supptabbox}{%
\footnotesize
\renewcommand{\arraystretch}{1.05}
\setlength{\tabcolsep}{4pt}
\begin{tabular}{p{0.45\columnwidth}p{0.43\columnwidth}}
\toprule
\textbf{Setting} & \textbf{Value} \\
\midrule
\rowcolor{cSub}\multicolumn{2}{l}{\textbf{Supervised cold start (\cref{sec:sft})}} \\
Training samples & 51{,}335 \\
Target generator & Qwen3.8-27B \\
Loss & Weighted CE with AESL \\
Answer weight $\alpha$ & 0.6 \\
AESL temperature $t_{\mathrm{scaling}}$ & 5 \\
Instruction placement & System turn \\
Optimiser & AdamW \\
Learning rate & $10^{-5}$ \\
Schedule & Cosine, 3\% warm-up \\
Weight decay & 0 \\
Gradient clip & 1.0 \\
Epochs & 1 \\
Effective batch size & 32 \\
Seed & 42 \\
\midrule
\rowcolor{cSub}\multicolumn{2}{l}{\textbf{Reward (\cref{sec:app-reward-hacking})}} \\
Format weight $F$ & 0.1 \\
Modality weight $M$ & 0.045 \\
Embedding weight $E$ & 0.3375 \\
Gated judge weight $JE$ & 0.5175 \\
Judge $J$, Qwen3-VL-4B policy & Qwen3-8B \\
Judge $J$, Qwen3-VL-8B policy & Qwen3-14B \\
Embedding model & MedEmbed-large \\
Gate threshold $\tau$ & 0.8 \\
Format violation reward & $-1$ \\
\bottomrule
\end{tabular}}
\captionsetup{width=\wd\supptabbox}
\caption{\textbf{Training configuration.} The reward weights are the effective coefficients of \cref{eq:app-reward}, where $F$, $M$, $E$, and $J$ are the format, modality, embedding, and judge terms.}
\usebox{\supptabbox}
\label{tab:app-configs}
\end{table}

\begin{table}[t]
\centering
\sbox{\supptabbox}{%
\footnotesize
\renewcommand{\arraystretch}{1.05}
\setlength{\tabcolsep}{4pt}
\begin{tabular}{p{0.45\columnwidth}p{0.43\columnwidth}}
\toprule
\textbf{Setting} & \textbf{Value} \\
\midrule
\rowcolor{cSub}\multicolumn{2}{l}{\textbf{Reinforcement learning (\cref{sec:app-dapo,sec:app-trainable-data})}} \\
Algorithm & GRPO with DAPO \\
Rollouts per sample $G$ & 5 \\
Temperature & 1.0 \\
Top-p & 1.0 \\
Samples per step & 512 \\
Generation round & 128 samples \\
Update mini batch & 128 samples \\
Retained groups & 1 to 4 correct of 5 \\
Maximum rounds per step & 300 \\
Clip $\epsilon_{\mathrm{low}}$ & 0.2 \\
Clip $\epsilon_{\mathrm{high}}$ & 0.3 \\
Dual clip & 3.0 \\
KL regularisation & Off \\
Optimiser & AdamW \\
Learning rate & $10^{-6}$ \\
Schedule & Constant \\
Weight decay & 0.01 \\
Gradient clip & 1.0 \\
Steps & 200 \\
Maximum prompt length & 4{,}352 tokens \\
Maximum rollout length & 4{,}096 tokens \\
Maximum image pixels & 4{,}194{,}304 \\
Hardware & 1 node, 4 GPUs \\
\midrule
\rowcolor{cSub}\multicolumn{2}{l}{\textbf{Length reward (\cref{app:reward-length})}} \\
Length threshold $T$ & 500 tokens \\
Decay scale $d$ & 500 tokens \\
Rollout cap $C$ & 4{,}096 tokens \\
Maximum penalty & $2\times$ \\
Correctness threshold & 0.5 \\
\midrule
\rowcolor{cSub}\multicolumn{2}{l}{\textbf{Evaluation (\cref{sec:protocol-final})}} \\
Benchmarks & 17 \\
Items & 19{,}200 \\
Decoding & Greedy \\
Context window & 32K tokens \\
Judges & InternVL3.5-30B-A3B \\
 & Qwen3.8-27B \\
 & MedGemma-27B \\
Judge for Qwen3.5 models & gpt-oss-20b instead of Qwen3.8 \\
Aggregation & Majority vote \\
Judge budget & 4{,}096 tokens \\
Report judge budget & 16{,}384 tokens \\
Binary answer window & 12 words after the marker \\
Fallback window & First 30 words \\
Short answer limit & 25 words \\
\bottomrule
\end{tabular}}
\captionsetup{width=\wd\supptabbox}
\caption{\textbf{Training and evaluation configuration.} $T$, $d$, and $C$ are the length threshold, decay scale, and rollout cap of \cref{eq:lw-weight}. The settings that differ for the Qwen3.5 runs are listed in \cref{tab:app-configs-q35}.}
\usebox{\supptabbox}
\label{tab:app-configs-rl}
\end{table}

\section{Supervised Cold Start}
\label{sec:sft}

This section provides additional details on the supervised cold start, including the training data, loss functions, and their effects, with hyperparameters summarised in \cref{tab:app-configs}. As discussed in \cref{sec:reasoning_cold_start} of the main paper, the cold start addresses samples for which the initial policy fails to generate correct rollouts. In our initial diagnostic run, approximately 21{,}000 samples yielded no judge-accepted rollouts, providing no successful rollouts for relative comparison during RL. Supervised fine-tuning provides targets for these otherwise uninformative samples while establishing the structured output format required by the reward, including modality tags, reasoning blocks, and answer blocks.

\subsection{Supervision Targets and Prompt Formats}
\label{sec:app-cold-targets}

For the supervised cold start, we use the same 51{,}335 samples as the RL stage and train for one epoch. Each training target contains modality, reasoning, and answer tags:
\begin{quote}
\ttfamily
<MODALITY>\\
<thinking>reasoning</thinking>\\
<answer>answer</answer>
\end{quote}

As described in \cref{sec:reasoning_cold_start}, a reasoning model first answers each question without seeing the reference. Incorrect answers are regenerated with reference guidance to retain difficult examples. Rationales that reveal the reference are also regenerated to reduce answer leakage. 
We compare two prompting strategies: \emph{sys}, which uses a system prompt that explicitly instructs the model to reason step by step before producing the final answer, and \emph{no sys}, which does not include these explicit reasoning instructions. All models are evaluated under the same conditions without additional system instructions to ensure a fair comparison.

\subsection{Cross Entropy and Answer Weighting}
\label{sec:app-cold-ce}

We start from standard cross entropy (CE), which supervises every target token of the rationale and the answer. Following the notation of~\citep{ICLR2026_bebed4d2}, the objective is
\begin{equation}
    \mathcal{L}_{\mathrm{CE}}(\theta)
    =
    -\mathbb{E}_{q,s^*\sim\mathcal{D}_{\mathrm{SFT}}}
    \left[
        \log\pi_\theta(s_t^*\mid q,s_{<t})
    \right],
    \label{eq:app-ce}
\end{equation}
where $\mathcal{D}_{\mathrm{SFT}}$ is the demonstration set, $q$ is the input including the image, $s^*$ is the target sequence, and $\pi_\theta$ is the model. Because the rationale is much longer than the answer, it occupies most of the supervised positions and dominates the loss, and tokens that are already predicted with confidence keep receiving supervision. This motivates two changes, which are to control the weight of the answer span and to reduce repeated updates on confident tokens.

To emphasise the answer while retaining supervision for the reasoning, we compute separate average losses for the answer and the remaining tokens. Here, $\mathcal{A}$ denotes the supervised answer positions, while $\mathcal{C}$ denotes the remaining supervised positions, including reasoning and structural tokens. The token-level loss is defined as $\ell_t=-\log\pi_\theta(s_t^*\mid q,s_{<t})$. The weighted objective is

\begin{equation}
    \mathcal{L}_{\mathrm{wCE}}
    =
    \alpha\frac{1}{|\mathcal{A}|}
    \sum_{t\in\mathcal{A}}\ell_t
    +
    (1-\alpha)\frac{1}{|\mathcal{C}|}
    \sum_{t\in\mathcal{C}}\ell_t,
    \qquad \alpha=0.6.
    \label{eq:app-wce}
\end{equation}
We assign weights of $0.6$ and $0.4$ to the average losses of the answer and remaining tokens, respectively. Since each loss is averaged separately, longer reasoning sequences do not receive greater weight simply because they contain more tokens. These weights control the contribution of each loss term, rather than the proportion of the gradient norm.

\begin{table*}[t]
\centering
\caption{\textbf{Supervised cold starts for Qwen3-VL-4B and 8B.} Sys uses a system prompt with step-by-step reasoning instructions, while no sys omits them. CE denotes cross-entropy, wCE weights answer and reasoning losses, and AESL adaptively selects tokens. Best and second-best scores per model are bold and underlined, respectively.}
\small
\setlength{\tabcolsep}{5.5pt}
\renewcommand{\arraystretch}{1.3}
\setlength{\aboverulesep}{0pt}\setlength{\belowrulesep}{0pt}
\resizebox{\textwidth}{!}{%
\begin{tabular}{l|cccccc|ccccccccccc|cccc}
\toprule
\multicolumn{1}{c|}{\rule{0pt}{2.2cm}} & \multicolumn{6}{c|}{\cellcolor{cVQA}} & \multicolumn{11}{c|}{\cellcolor{cQA}} & \multicolumn{4}{c}{\cellcolor{cAVG}} \\[-2.37cm]
\multicolumn{1}{c|}{\rule{0pt}{2.2cm}\raisebox{0.93cm}[0pt][0pt]{\large\textbf{Method}}}
& \multicolumn{1}{c}{\raisebox{0.93cm}[0pt][0pt]{\makebox[0pt][c]{\rotatebox[origin=c]{65}{\footnotesize\textbf{SLAKE}}}}}
& \multicolumn{1}{c}{\raisebox{0.93cm}[0pt][0pt]{\makebox[0pt][c]{\rotatebox[origin=c]{65}{\footnotesize\textbf{VQA-RAD}}}}}
& \multicolumn{1}{c}{\raisebox{0.93cm}[0pt][0pt]{\makebox[0pt][c]{\rotatebox[origin=c]{65}{\footnotesize\textbf{PathVQA}}}}}
& \multicolumn{1}{c}{\raisebox{0.93cm}[0pt][0pt]{\makebox[0pt][c]{\rotatebox[origin=c]{65}{\footnotesize\textbf{PMC-VQA}}}}}
& \multicolumn{1}{c}{\raisebox{0.93cm}[0pt][0pt]{\makebox[0pt][c]{\rotatebox[origin=c]{65}{\footnotesize\textbf{PMC-VQA-H}}}}}
& \multicolumn{1}{c|}{\raisebox{0.93cm}[0pt][0pt]{\makebox[0pt][c]{\rotatebox[origin=c]{65}{\footnotesize\textbf{MIMIC-Gen}}}}}
& \multicolumn{1}{c}{\raisebox{0.93cm}[0pt][0pt]{\makebox[0pt][c]{\rotatebox[origin=c]{65}{\footnotesize\textbf{MMLU-Clin}}}}}
& \multicolumn{1}{c}{\raisebox{0.93cm}[0pt][0pt]{\makebox[0pt][c]{\rotatebox[origin=c]{65}{\footnotesize\textbf{MMLU-Bio}}}}}
& \multicolumn{1}{c}{\raisebox{0.93cm}[0pt][0pt]{\makebox[0pt][c]{\rotatebox[origin=c]{65}{\footnotesize\textbf{MMLU-Med}}}}}
& \multicolumn{1}{c}{\raisebox{0.93cm}[0pt][0pt]{\makebox[0pt][c]{\rotatebox[origin=c]{65}{\footnotesize\textbf{MMLU-Gen}}}}}
& \multicolumn{1}{c}{\raisebox{0.93cm}[0pt][0pt]{\makebox[0pt][c]{\rotatebox[origin=c]{65}{\footnotesize\textbf{MMLU-Prof}}}}}
& \multicolumn{1}{c}{\raisebox{0.93cm}[0pt][0pt]{\makebox[0pt][c]{\rotatebox[origin=c]{65}{\footnotesize\textbf{MMLU-Anat}}}}}
& \multicolumn{1}{c}{\raisebox{0.93cm}[0pt][0pt]{\makebox[0pt][c]{\rotatebox[origin=c]{65}{\footnotesize\textbf{MedMCQA}}}}}
& \multicolumn{1}{c}{\raisebox{0.93cm}[0pt][0pt]{\makebox[0pt][c]{\rotatebox[origin=c]{65}{\footnotesize\textbf{MedQA}}}}}
& \multicolumn{1}{c}{\raisebox{0.93cm}[0pt][0pt]{\makebox[0pt][c]{\rotatebox[origin=c]{65}{\footnotesize\textbf{USMLE-SA}}}}}
& \multicolumn{1}{c}{\raisebox{0.93cm}[0pt][0pt]{\makebox[0pt][c]{\rotatebox[origin=c]{65}{\footnotesize\textbf{PubMedQA}}}}}
& \multicolumn{1}{c|}{\raisebox{0.93cm}[0pt][0pt]{\makebox[0pt][c]{\rotatebox[origin=c]{65}{\footnotesize\textbf{MIMIC-Sum}}}}}
& \multicolumn{1}{c}{\raisebox{0.93cm}[0pt][0pt]{\makebox[0pt][c]{\rotatebox[origin=c]{65}{\footnotesize\textbf{VQA}}}}}
& \multicolumn{1}{c}{\raisebox{0.93cm}[0pt][0pt]{\makebox[0pt][c]{\rotatebox[origin=c]{65}{\footnotesize\textbf{QA}}}}}
& \multicolumn{1}{c}{\raisebox{0.93cm}[0pt][0pt]{\makebox[0pt][c]{\rotatebox[origin=c]{65}{\footnotesize\textbf{Macro}}}}}
& \multicolumn{1}{c}{\raisebox{0.93cm}[0pt][0pt]{\makebox[0pt][c]{\rotatebox[origin=c]{65}{\footnotesize\textbf{Micro}}}}} \\
\midrule
\rowcolor{cSub}
\textbf{Qwen3-VL-4B (zero-shot)}
& 67.9 & 54.3 & 40.0 & \textbf{53.0} & \textbf{34.2} & \underline{31.6} & 78.9 & 88.2 & \textbf{80.3} & \underline{88.0} & \textbf{78.7} & \underline{71.8} & \textbf{58.7} & \underline{65.9} & \textbf{71.1} & 40.6 & 54.2 & 46.8 & \textbf{70.6} & \textbf{62.2} & 51.5 \\
\hspace{1.2em}CE, no sys
& 70.8 & \underline{55.4} & 40.5 & \underline{51.4} & 33.1 & 30.0 & \underline{80.0} & \underline{88.9} & 78.6 & \textbf{89.0} & 76.8 & \textbf{73.3} & \underline{58.6} & \textbf{66.6} & 65.2 & 42.4 & \underline{54.8} & 46.9 & \underline{70.4} & \underline{62.1} & 51.5 \\
\hspace{1.2em}CE, sys
& 71.0 & \textbf{58.3} & 42.9 & 51.2 & \underline{33.8} & 30.0 & 76.6 & \textbf{89.6} & 74.6 & 87.0 & \underline{77.9} & 71.1 & 57.8 & 65.5 & \underline{70.2} & 44.8 & 52.8 & \underline{47.9} & 69.8 & \underline{62.1} & 52.3 \\
\hspace{1.2em}wCE, sys
& \underline{73.6} & 53.0 & \underline{49.0} & 49.5 & 30.7 & 31.2 & \textbf{80.4} & 87.5 & 78.6 & 85.0 & 72.8 & 69.6 & 56.2 & 60.2 & 61.9 & \underline{45.4} & 52.8 & 47.8 & 68.2 & 61.0 & \underline{53.2} \\
\rowcolor{cOurs}
\hspace{1.2em}\textbf{wCE + AESL, sys}
& \textbf{73.9} & 54.8 & \textbf{49.5} & 50.4 & 32.0 & \textbf{33.4} & 78.9 & 87.5 & \underline{79.2} & 85.0 & 74.3 & 69.6 & 56.4 & 61.7 & 61.9 & \textbf{48.8} & \textbf{56.0} & \textbf{49.0} & 69.0 & 62.0 & \textbf{53.9} \\
\midrule
\rowcolor{cSub}
\textbf{Qwen3-VL-8B (zero-shot)}
& 61.6 & 52.8 & 36.0 & \textbf{53.2} & \underline{34.0} & \textbf{34.4} & 80.4 & 92.4 & 80.3 & \underline{87.0} & \textbf{84.6} & \underline{75.6} & \textbf{62.3} & \textbf{75.9} & \textbf{77.2} & 42.0 & \underline{53.0} & 45.3 & \underline{73.7} & 63.7 & 51.4 \\
\hspace{1.2em}CE, no sys
& \textbf{74.1} & 54.8 & 44.0 & 51.5 & 33.7 & 24.8 & \textbf{83.4} & 91.0 & 79.8 & \underline{87.0} & 78.3 & 74.8 & 60.4 & 69.7 & 68.9 & \textbf{52.4} & 51.8 & 47.2 & 72.5 & 63.5 & 54.0 \\
\hspace{1.2em}CE, sys
& \underline{73.2} & 53.7 & 44.0 & \underline{52.5} & 31.2 & 26.6 & \underline{82.6} & \textbf{94.4} & \textbf{82.7} & 86.0 & \underline{80.9} & 72.6 & 61.3 & 71.6 & 69.5 & \underline{46.6} & 51.6 & 46.9 & 72.7 & 63.6 & 54.0 \\
\hspace{1.2em}wCE, sys
& 72.3 & \underline{60.1} & \underline{50.5} & 51.4 & 33.2 & 24.0 & 82.3 & \underline{93.1} & \underline{81.5} & 86.0 & \underline{80.9} & 74.8 & 61.1 & \underline{71.8} & 73.5 & 45.0 & 51.8 & \underline{48.6} & 72.9 & \underline{64.3} & \underline{56.4} \\
\rowcolor{cOurs}
\hspace{1.2em}\textbf{wCE + AESL, sys}
& 72.3 & \textbf{60.5} & \textbf{52.1} & 50.6 & \textbf{34.7} & \underline{28.0} & \underline{82.6} & 92.4 & \textbf{82.7} & \textbf{88.0} & \textbf{84.6} & \textbf{78.5} & \underline{61.4} & \underline{71.8} & \underline{74.5} & 44.2 & \textbf{53.6} & \textbf{49.7} & \textbf{74.0} & \textbf{65.4} & \textbf{57.2} \\
\bottomrule
\end{tabular}}

\label{tab:coldstart-ladder}
\end{table*}

\subsection{Adaptive Early-Stop Loss}
\label{sec:app-aesl}

Answer weighting balances the loss between reasoning and answer tokens but does not account for how confidently the model predicts each token. We therefore apply Adaptive Early-Stop Loss (AESL)~\citep{ICLR2026_bebed4d2}, which reduces the loss for tokens the model already predicts with high confidence. This allows the model to focus on less familiar patterns while limiting changes to its existing knowledge before RL. The original AESL formulation uses an adaptively weighted cross-entropy (CE) objective,

\begin{multline}
    \mathcal{L}_{\mathrm{Ada\text{-}stop}}(\theta)
    = \\
    -\mathbb{E}_{q,s^*\sim\mathcal{D}_{\mathrm{SFT}}}
    \left[
        p(q,s_t^*,\pi_\theta)
        \cdot\log\pi_\theta(s_t^*\mid q,s_{<t})
    \right],
    \label{eq:app-aesl-loss}
\end{multline}
with the adaptive weighting
\begin{multline}
    p(q,s_t^*,\pi_\theta)
    = \\
    1-\operatorname{softmax}\left[
        \frac{
            y(s_t^*\mid q,s_{<t})
        }{
            -t_{\mathrm{scaling}}\cdot
            \frac{1}{|t|}
            \sum_{i=1}^{t}
            \log\pi_\theta(s_i^*\mid q,s_{<i})
        }
    \right].
    \label{eq:app-aesl-weight}
\end{multline}

Here, $y$ represents the logits before softmax, and $t_{\mathrm{scaling}}$ controls the temperature. The softmax gives the target token probability, assigning lower weights to tokens predicted with high confidence. The denominator accounts for the average prediction difficulty of the sequence up to the current token.
In our implementation, AESL uses a stochastic mask instead of directly weighting the loss. Each supervised token is retained with probability $1-p$, where $p$ is its target probability under the rescaled distribution. For example, tokens with probabilities of $0.95$ and $0.20$ are retained with probabilities of $5\%$ and $80\%$, respectively. The mask is computed without gradients, and standard cross-entropy is applied only to retained tokens. For the weighted variant, answer and reasoning weights are also applied, with $t_{\mathrm{scaling}}=5$. Skipped tokens remain in the sequence and may be selected in later updates.
Our implementation also uses a slightly modified difficulty statistic, computed from preceding supervised tokens and clipped to $[0,10]$ for numerical stability. During the middle of the 8B training run, approximately $10\%$ of supervised tokens were skipped, with a higher proportion of skipped tokens in the reasoning than in the final answer.

\subsection{Effect of the Objectives}
\label{sec:app-cold-effect}

\Cref{tab:coldstart-ladder} compares the cold-start objectives for both model sizes. For Qwen3-VL-8B, combining answer weighting with AESL achieves the best results, improving macro accuracy from $63.7$ to $65.4$ and micro accuracy from $51.4$ to $57.2$ compared with the base model. The largest gains are observed on PathVQA and SLAKE. For Qwen3-VL-4B, the combined objective keeps macro accuracy ($62.0$ against $62.2$) and raises micro accuracy from $51.5$ to $53.9$. These cold-start models serve as the initial policies for RL, where the reward function described in the next section guides further training.

\section{Reward Design and Embedding Gating}
\label{sec:app-reward-hacking}

The cold start gives the policy a structured output format, and reinforcement learning then improves its answers through a reward. This section defines that reward, explains how the embedding gate protects it from an imperfect judge, and shows the reward hacking that motivated the gate.

\subsection{Reward Formulation}
\label{sec:app-reward-formula}

Each sampled rollout receives four signals. The format reward $F$ is one when the rollout contains a thinking block followed by an answer block and zero otherwise. For the 8B run and the Qwen3.5 runs the check is strict, so exactly one thinking block and one answer block must appear, no tag may occur inside either block, and nothing may follow the closing answer tag. A rollout that violates this rule is not judged and receives a reward of $-1$. The modality reward $M$ is one when the modality tag generated before the reasoning matches the modality of the reference. The judge reward $J$ is the binary decision of an LLM judge that compares the extracted answer with the reference and considers only semantic equivalence, not reasoning style or length. The embedding indicator $E$ is one when the cosine similarity between the embeddings of the predicted and the reference answer is at least $\tau=0.80$, computed with the MedEmbed-large model~\cite{balachandran2024medembed}. Answers that are identical to the reference are accepted directly, and answers that are empty, a single character, or malformed receive no content reward. The final reward is
\begin{equation}
    R = 0.1\,F + 0.9\left(0.05\,M + 0.375\,E + 0.575\,J\,E\right).
    \label{eq:app-reward}
\end{equation}
The effective weights of the format, modality, embedding, and gated judge terms are therefore $0.1$, $0.045$, $0.3375$, and $0.5175$~\cite{mullappilly2026medix}, so semantic agreement dominates the reward while the format and modality terms preserve the structure learned during the cold start. We also track rollouts that the judge accepts and the embedding rejects, because they signal disagreement between the two evaluators.

\subsection{What the Embedding Gate Enforces}
\label{sec:app-gate}

The product $J\,E$ in \cref{eq:app-reward} represents the gated judge reward. Here, $J(a,y)$ indicates whether the training judge considers the predicted answer $a$ correct relative to the reference $y$, while $E(a,y)$ indicates whether their embedding cosine similarity meets the threshold. The gated reward is therefore
\begin{equation}
    J_{\mathrm{gated}}(a,y)=J(a,y)E(a,y).
    \label{eq:app-gated-judge}
\end{equation}
A positive judge reward is given only when the predicted answer also passes the semantic similarity check. This discourages the model from listing multiple possible answers to increase the chance of matching the reference. However, the gate does not guarantee medical correctness or eliminate all such rollouts. It evaluates only the final answer, not the reasoning, and embedding similarity remains a training signal rather than an independent measure of clinical accuracy.

Because the gate multiplies the judge decision, it can only remove credit that the judge has given and never adds credit to an answer that the judge rejects. The gate is not used at evaluation, where every reported score comes from the deterministic rules and the three-judge panel of \cref{sec:protocol-final}.

\subsection{How Well the Gate Separates Correct from Wrong Answers}
\label{sec:app-gate-separation}

To examine the embedding gate, we measure the similarity between the references and correct and incorrect rollouts of the 8B cold-start model on multiple-choice questions, whose correctness can be read without a judge, sampling up to 1{,}100 correct and 1{,}100 incorrect rollouts.

\paragraph{Multiple-choice answers}
The gate separates correct and incorrect answers well, with an ROC-AUC of $0.97$. At $\tau=0.80$, about $97\%$ of correct answers pass, while about $73\%$ of incorrect answers are rejected.

\paragraph{Open-ended answers}
For open-ended answers, the effect of the gate is shown by training and benchmark results. Without the gate, the judge reward keeps rising while embedding agreement falls (\cref{fig:judge_accuracy,fig:embedding_accuracy}), which is the pattern of answer listing. Adding the gate raises accuracy under the independent evaluation from $50.6$ to $51.2$ macro and from $50.4$ to $52.0$ micro on the VQA benchmarks (\cref{tab:ablation} and \cref{tab:ablation_full}). At $\tau=0.80$, only about $7\%$ of correct rollouts are filtered, against $44\%$ and $67\%$ at $0.90$ and $0.95$ (\cref{fig:gate_threshold_area}), which is why we use $\tau=0.80$.

\paragraph{Why the gate reads only the final answer}
We evaluated whether comparing the model's reasoning with the reference answer could help identify correct predictions. However, reasoning similarity was a weak indicator of correctness (ROC-AUC $=0.63$), and no more than $5\%$ of reasoning texts met the similarity threshold. This is because reasoning often contains explanations and intermediate steps that differ from the short reference answer, even when the prediction is correct. Therefore, we apply the similarity gate only to the extracted final answer.

\subsection{Reward Hacking Examples}
\label{sec:app-hack-examples}

\begin{figure*}[p]
    \centering
    \sbox{\suppfigbox}{\includegraphics[width=\textwidth,height=0.79\textheight,keepaspectratio]{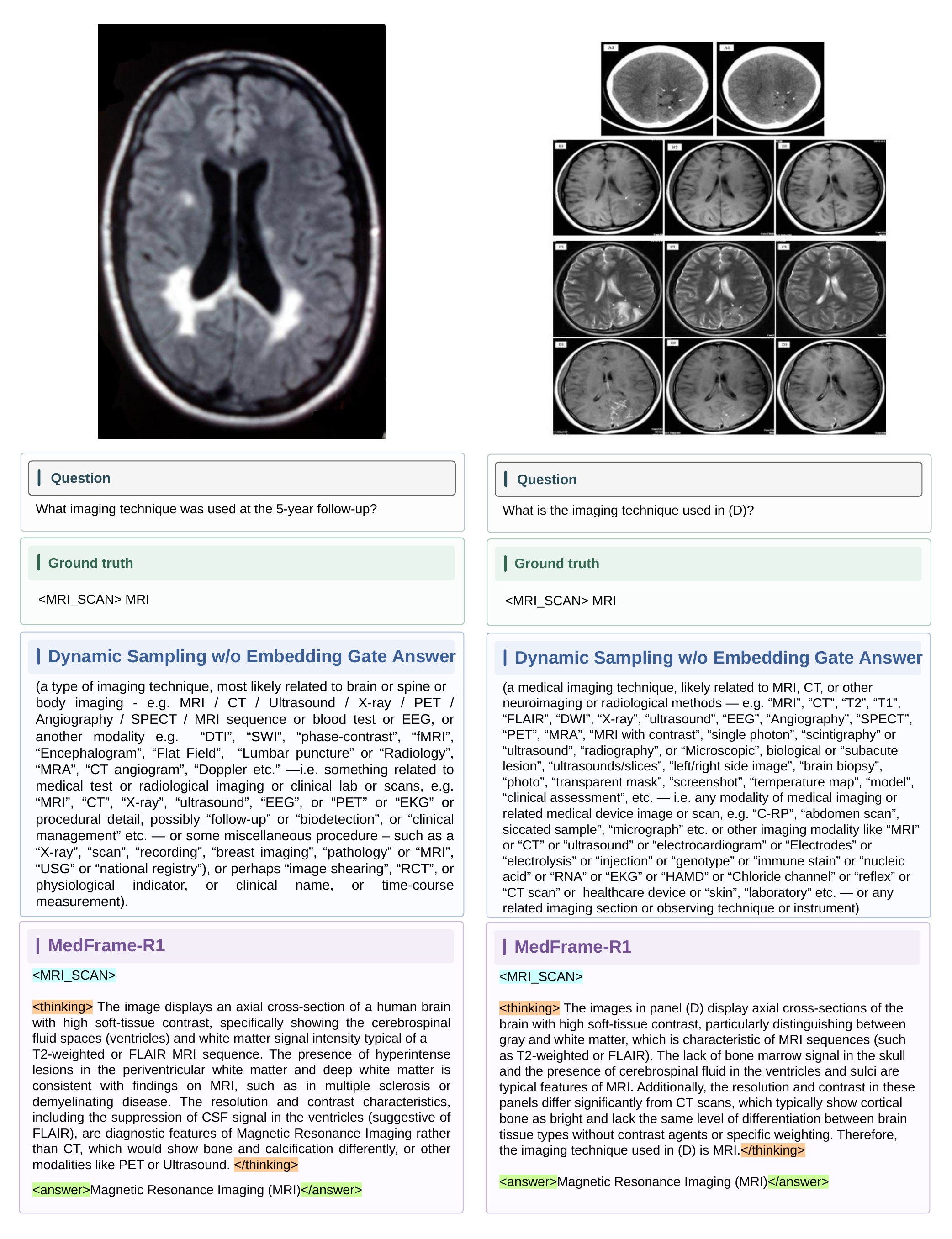}}%
    \usebox{\suppfigbox}
\captionsetup{width=\wd\suppfigbox}
    \caption{\textbf{Modality lists.} The ungated rollouts hide MRI inside long lists of techniques, while MedFrame-R1 selects MRI.}
    \label{fig:app-hack-1}
\end{figure*}

\begin{figure*}[p]
    \centering
    \sbox{\suppfigbox}{\includegraphics[width=\textwidth,height=0.79\textheight,keepaspectratio]{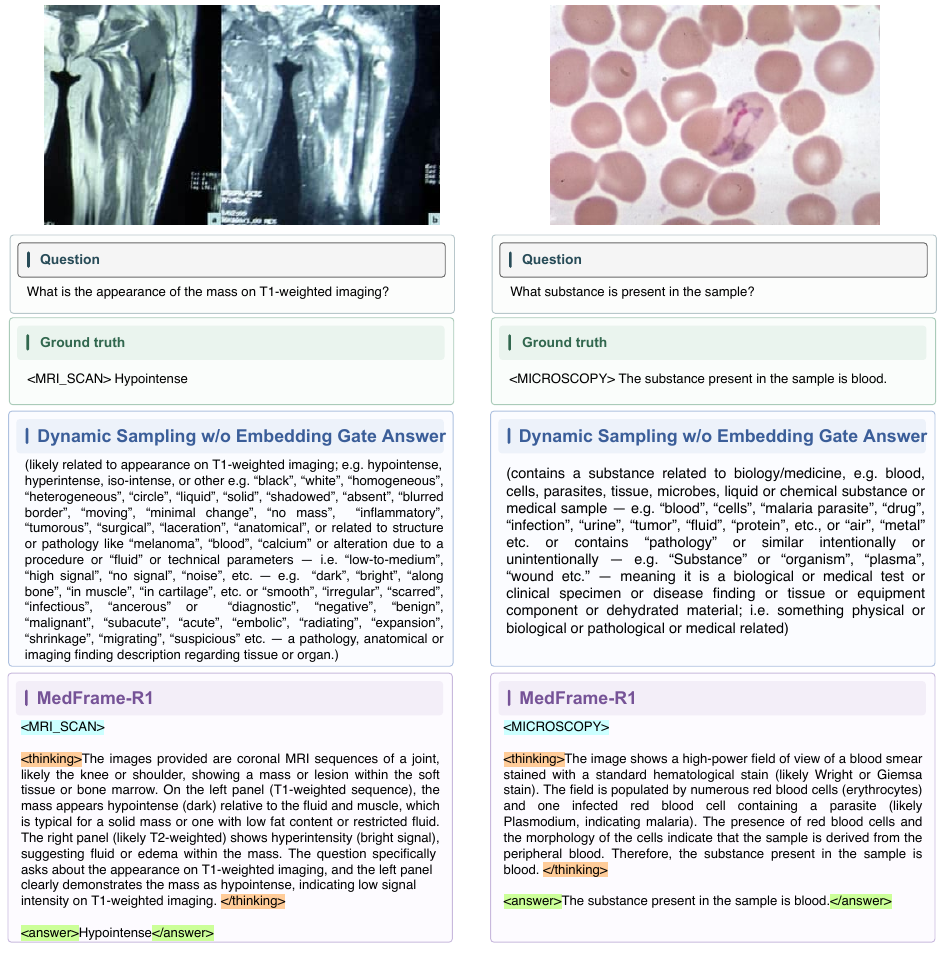}}%
    \usebox{\suppfigbox}
\captionsetup{width=\wd\suppfigbox}
    \caption{\textbf{Contradictory medical terms.} The ungated policy lists incompatible findings, while MedFrame-R1 selects hypointense and blood.}
    \label{fig:app-hack-4}
\end{figure*}

The training judge can be exploited during RL, as repeated optimisation may reward rollouts that appear correct without providing a clear answer. \Cref{fig:app-hack-1,fig:app-hack-4} compare dynamic sampling without the embedding gate against MedFrame-R1. Without the gate, the model generates long, repetitive, and sometimes contradictory lists of possible answers, increasing the chance that the reference appears somewhere in the rollout. This allows the judge reward to increase even as embedding agreement decreases.
For example, in \cref{fig:app-hack-1}, the ungated model lists several imaging techniques instead of directly predicting MRI. Similarly, \cref{fig:app-hack-4} shows lists of conflicting alternatives, while MedFrame-R1 produces focused answers such as ``hypointense'' and ``blood''. Similar behaviour, not shown here, occurs in location and counting questions.

\section{Dynamic Sampling and Policy Optimisation}
\label{sec:app-dapo}

A reliable reward helps only when the sampled rollouts for a sample differ in correctness, so the RL stage decides which samples are used and how the policy is updated. This section defines the samples that dynamic sampling keeps and then gives the objective and the optimisation settings. \Cref{sec:app-trainable-data} examines how many samples contribute over a full run.

\subsection{What Makes a Sample Trainable}
\label{sec:app-trainability-definition}

For each sample $q$, the policy generates $G=5$ rollouts. The raw judge assigns a binary correctness score $c_i\in\{0,1\}$ to each rollout. Dynamic sampling retains a group when

\begin{equation}
    0 < \sum_{i=1}^{G}c_i < G,
    \label{eq:app-trainability}
\end{equation}
so the group contains at least one accepted and one rejected rollout, and the update can raise the relative probability of rollouts that receive a higher reward. We call a sample \emph{easy} when all five rollouts are accepted, \emph{hard} when all five are rejected, and \emph{trainable} when the group contains both outcomes. These labels describe the observed group under the current policy and judge and do not make a sample permanently easy or impossible. The selection uses the raw judge decision before the gate, so the gate changes the reward but never decides which samples are kept.

\subsection{Objective and Optimisation Settings}
\label{sec:app-dapo-settings}

Each retained group updates the policy with the DAPO objective~\citep{NEURIPS2025_a4277440},
\begin{multline}
    \mathcal{J}_{\mathrm{DAPO}}(\theta) =
    \mathbb{E}\Bigg[
    \frac{1}{\sum_{i=1}^{G}|o_i|}
    \sum_{i=1}^{G}\sum_{t=1}^{|o_i|}\\
    \min\Big(
    r_{i,t}(\theta)\hat{A}_i,\,
    \operatorname{clip}\left(
    r_{i,t}(\theta),
    1-\epsilon_{\mathrm{low}},
    1+\epsilon_{\mathrm{high}}
    \right)\hat{A}_i
    \Big)
    \Bigg],
    \label{eq:app-dapo}
\end{multline}
where $o_i$ is the $i$th of the $G$ rollouts for sample $q$ and $|o_i|$ is its length in tokens. The probability ratio
\begin{equation}
    r_{i,t}(\theta)=\frac{\pi_\theta(o_{i,t}\mid q,o_{i,<t})}{\pi_{\theta_{\mathrm{old}}}(o_{i,t}\mid q,o_{i,<t})}
    \label{eq:app-ratio}
\end{equation}
compares the current and the old policy at token $t$. The expectation is taken over samples from the training set and rollouts of the old policy, the sum is normalised by the total number of generated tokens, and $\epsilon_{\mathrm{low}}$ and $\epsilon_{\mathrm{high}}$ are the asymmetric clipping bounds. The advantage of rollout $i$ is normalised within its group,
\begin{equation}
    \hat{A}_i=\frac{R_i-\operatorname{mean}(\{R_j\}_{j=1}^{G})}{\operatorname{std}(\{R_j\}_{j=1}^{G})+\epsilon},
    \label{eq:app-advantage}
\end{equation}
where $R_i$ is the final gated reward of \cref{eq:app-reward} and $\epsilon$ is a small constant for numerical stability. Each rollout is compared only with the other rollouts to the same question, so rollouts above the group mean receive a positive advantage and those below it a negative one. Before the update, the length reward of \cref{app:reward-length} rescales these advantages according to the rollout length.

The probability ratio is clipped asymmetrically with $\epsilon_{\mathrm{low}}=0.2$ and $\epsilon_{\mathrm{high}}=0.3$, a dual clip of $3.0$ bounds the update for negative advantages, and the loss is averaged over tokens. Each training step collects 512 samples with mixed outcomes. Samples are generated 128 at a time with five rollouts each, and generation repeats until the batch is full, after which the policy is updated in mini batches of 128 samples. We disable KL regularisation and keep no separate reference model, which reduces memory use and computation. The optimiser is AdamW with a constant learning rate of $10^{-6}$, weight decay $0.01$, and gradient clipping at $1.0$, and a run lasts 200 steps. The full list of settings is given in \cref{tab:app-configs,tab:app-configs-rl}.

\section{Length Reward}
\label{app:reward-length}

\begin{figure*}[tp]
    \centering
    \sbox{\suppfigbox}{\includegraphics[width=\textwidth]{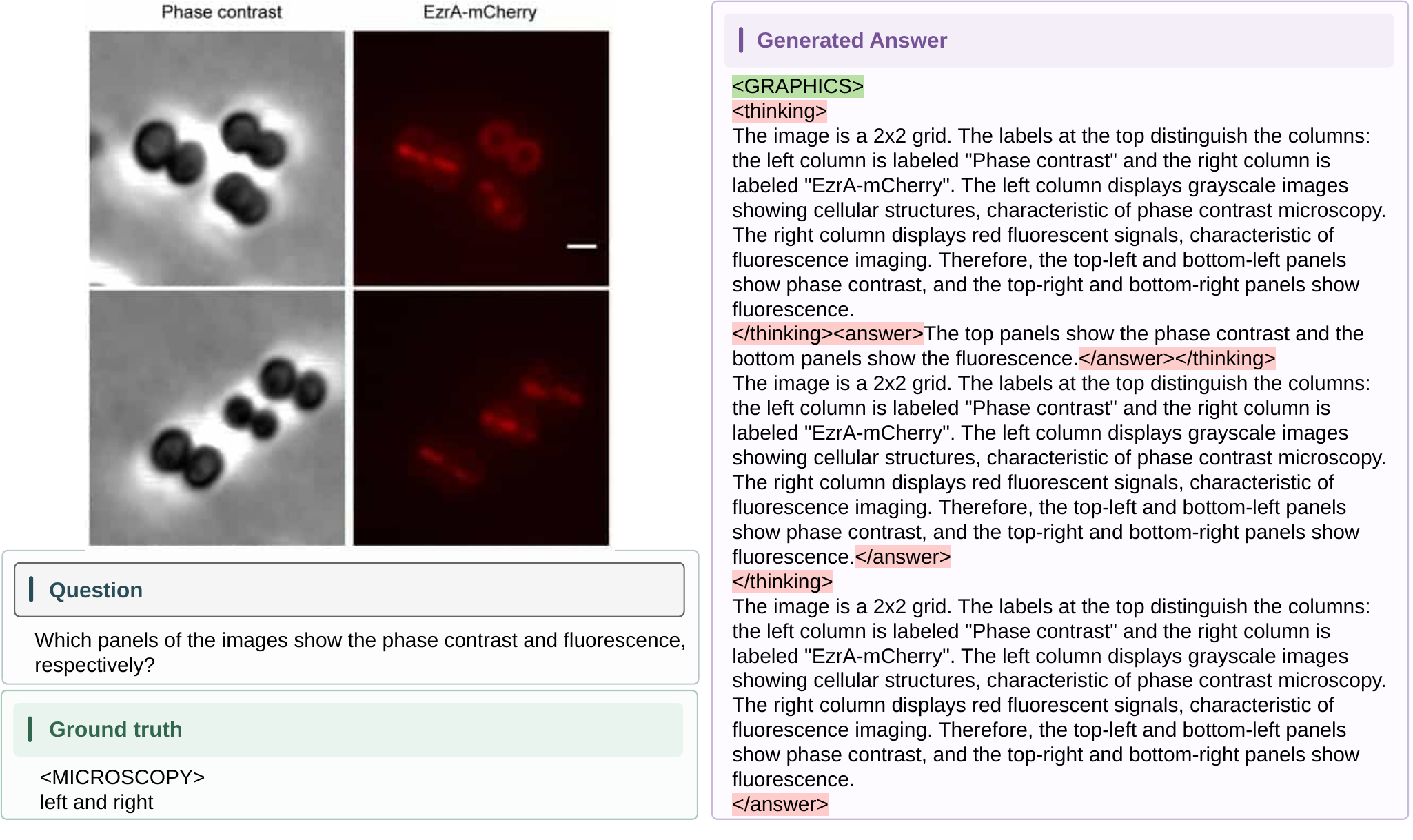}}%
    \usebox{\suppfigbox}
\captionsetup{width=\wd\suppfigbox}
    \caption{\textbf{Repeated reasoning and tags.} A training rollout of the Qwen3-VL-8B policy without the length reward duplicates its explanation and repeats the closing thinking and answer tags.}
    \label{fig:lw1}
\end{figure*}

\begin{figure*}[p]
    \centering
    \sbox{\suppfigbox}{\includegraphics[width=\textwidth,height=0.79\textheight,keepaspectratio]{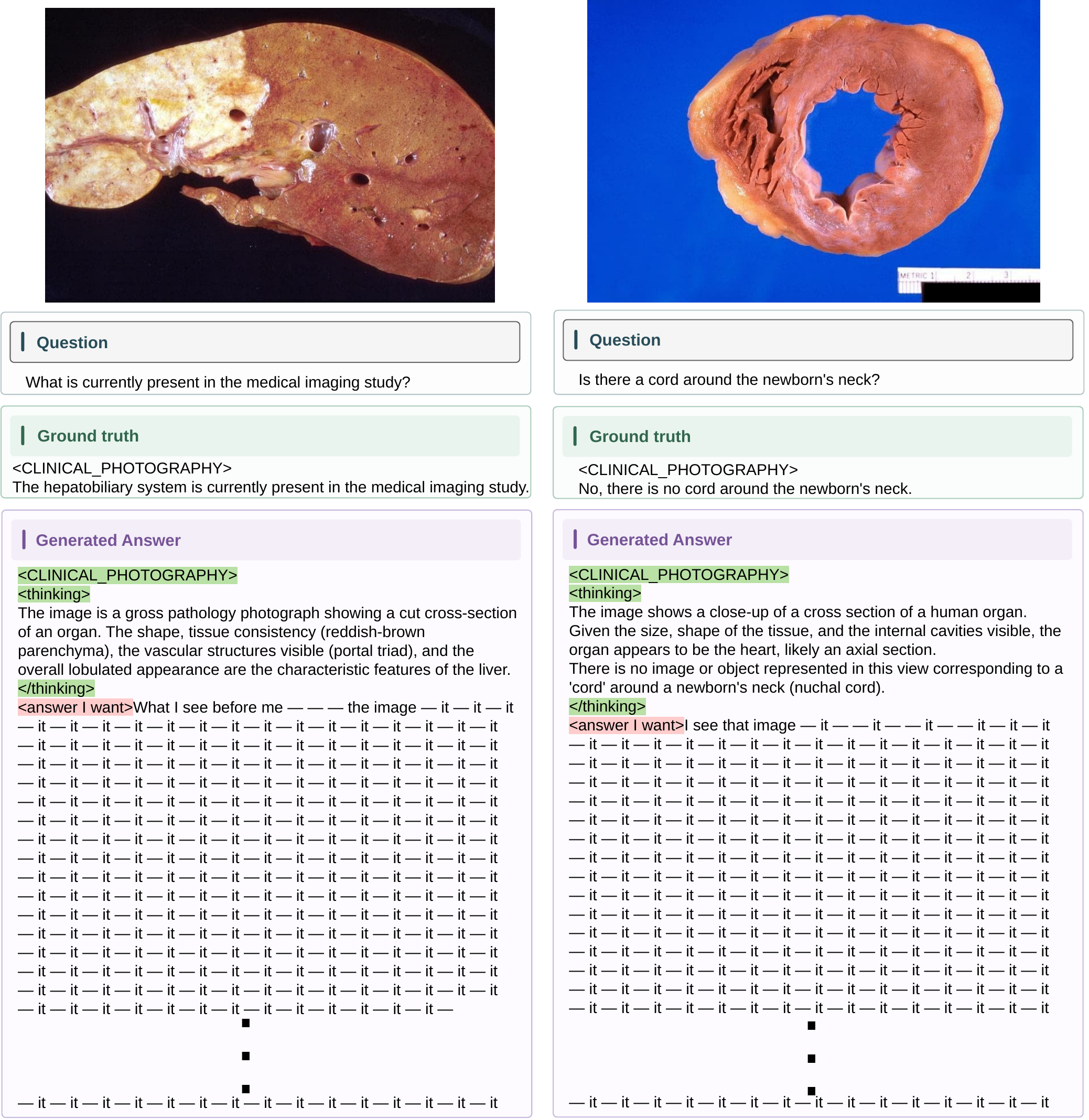}}%
    \usebox{\suppfigbox}
\captionsetup{width=\wd\suppfigbox}
    \caption{\textbf{Word and punctuation loops.} Training rollouts of the Qwen3-VL-8B policy without the length reward. Clinical photography answers loop on the word it and on dashes.}
    \label{fig:lw-tokens-4}
\end{figure*}

The reward and optimisation objective described so far do not directly control rollout length. We observed excessive generation and repetition when scaling MedFrame-R1 to Qwen3-VL-8B and when applying RL directly to Qwen3.5-4B and Qwen3.5-9B.
For Qwen3-VL-8B, the model sometimes repeats reasoning and answer tags, generates multiple reasoning blocks, or produces loops of words and punctuation (\cref{fig:lw1,fig:lw-tokens-4}). Similar problems occur in the zero-shot Qwen3.5 models. Before training, their rollouts average approximately 2{,}300 tokens, with about a quarter reaching the 4{,}096-token limit. Using a repetition criterion in which eight consecutive words appear at least three times, we find that $53.6\%$ of Qwen3.5-4B and $46.6\%$ of Qwen3.5-9B rollouts are repetitive (\cref{tab:app-q35-rep-split}). These rollouts are longer and less likely to produce a complete answer. \Cref{fig:app-q35-rep-example} shows examples where both models repeatedly switch between possible answers until reaching the generation limit.

To address these problems, we introduce a \emph{length reward} that discourages unnecessary generation while preserving useful reasoning. Despite its name, the method does not modify the reward directly. Instead, it rescales the advantages after reward normalisation. We retain the GRPO objective with PPO clipping and DAPO dynamic sampling, using $G=5$ rollouts per sample. The method combines DRPO-inspired weighting of correct rollouts with positive advantages and length-dependent scaling of incorrect rollouts with negative advantages~\cite{li2026drpo}.
For each rollout $o_i$, $L_i$ denotes its generated token length, and $R_i$ denotes its original reward. We first compute the group-relative advantage,
\begin{equation}
    A_i = \frac{R_i-\operatorname{mean}_{j\in g}R_j}
                  {\operatorname{std}_{j\in g}R_j},
    \label{eq:lw-original-advantage}
\end{equation}
with the zero variance handling and numerical stabilisation of the training implementation. Correctness for the length rules is defined by
\begin{equation}
    c_i = \mathbf{1}\!\left[
        \mathrm{acc}_{\mathrm{llm},i}\geq 0.5
        \;\lor\;
        \mathrm{acc}_{\mathrm{embed},i}\geq 0.5
    \right],
    \label{eq:lw-correctness}
\end{equation}
which differs from the criterion used by the dynamic sampling filter. We use a training rollout cap of $C=4{,}096$ tokens, a length threshold of $T=500$ tokens, and an exponential decay scale of $d=500$ tokens. The length weight is
\begin{equation}
    w_i =
    \begin{cases}
        1, & L_i\leq T,\\
        \exp\!\left(-\dfrac{L_i-T}{d}\right), & L_i>T,
    \end{cases}
    \label{eq:lw-weight}
\end{equation}
and $P_g=\{j\in g:c_j=1,\ A_j>0\}$ contains rollouts classified as correct by the length-weighting rule and having positive advantages in group $g$. Their weights are normalised within $P_g$, while the second component increases the magnitude of a negative advantage when an incorrect rollout exceeds the threshold. Together the two components give
\begin{equation}
    \begin{gathered}
    \widetilde A_i=f_i A_i,\\[4pt]
    f_i=
    \begin{cases}
        \dfrac{w_i}{|P_g|^{-1}\sum_{j\in P_g}w_j},
            & i\in P_g,\\[8pt]
        1+\min\!\left(1,\dfrac{L_i-T}{C-T}\right),
            & \substack{c_i=0,\ A_i<0,\\L_i>T,}\\[6pt]
        1, & \text{otherwise}.
    \end{cases}
    \end{gathered}
    \label{eq:lw-rescaling}
\end{equation}

The first case applies only when $P_g$ is non-empty. The adjusted advantage $\widetilde A_i$ then replaces $A_i$ in the clipped policy objective without further normalisation.
The first component follows DRPO~\cite{li2026drpo}, which favours shorter correct rollouts. While DRPO applies this idea within the DisCO objective~\cite{li2026disco}, we apply it to positive GRPO advantages. Shorter rollouts in this set receive higher weights, and groups with only one eligible rollout retain a weight of one. The second component extends this approach by increasing the penalty for long rollouts classified as incorrect by the length-weighting rule. For example, the magnitude of a negative advantage increases by approximately $1.14\times$, $1.42\times$, and $1.70\times$ at 1{,}000, 2{,}000, and 3{,}000 tokens, reaching $2\times$ at 4{,}096 tokens. All multipliers are positive, so the signs of the advantages remain unchanged. Other rollouts retain their original advantages.

We use a soft length threshold of $T=500$ tokens to allow reasoning while discouraging unnecessary generation. The reference answers are generally short, with a median of 13 tokens. We compare $T=500$ and $T=1500$ on Qwen3.5 below.
\Cref{fig:response_length_lw_val} shows that the length reward stabilises training rollouts at approximately 300--400 tokens and validation rollouts at 250--300 tokens. Without it, rollout lengths increase substantially, reaching approximately 1{,}800 tokens during validation. These results show that the length reward controls generation length, while its effect on accuracy is evaluated in \cref{tab:ablation} of the main paper.

\subsection{Qwen3.5}
\label{sec:app-qwen35-recipe}
\label{sec:app-qwen35}
\label{sec:app-qwen35-coldstart}

We also apply MedFrame-R1 to Qwen3.5-4B and Qwen3.5-9B without a supervised cold start. The released checkpoints already achieve zero-shot macro averages of $69.2$ and $72.2$, respectively (\cref{tab:main}), so we train them directly with RL. This also tests whether the proposed training method can improve a stronger starting model without supervised fine-tuning.
Both runs use the same 51{,}335 samples, dynamic sampling based on the raw judge decision, the embedding gate with $\tau=0.80$, a strict format check with a reward of $-1$ for violations, and no KL regularisation. Training lasts 100 steps, with samples generated in rounds of 1{,}024 to improve computational efficiency. Since dynamic sampling retains only informative groups, generating a larger batch reduces the overhead of repeated generation rounds. Any informative groups beyond those needed for the current update are carried over to the next step, avoiding unnecessary regeneration and improving training speed. The 4B policy uses Qwen3.5-9B as its training judge, whereas the 9B policy uses Qwen3-14B. Other settings follow \cref{tab:app-configs,tab:app-configs-rl}, and \cref{tab:app-configs-q35} summarises the differences. Evaluation follows \cref{sec:protocol-final}.
The Qwen3.5 zero-shot rows in \cref{tab:main} evaluate the released checkpoints.

\paragraph{Results on Qwen3.5}
RL alone improves both released checkpoints (\cref{tab:main}). The macro average rises from $69.2$ to $72.9$ at 4B and from $72.2$ to $74.5$ at 9B, and the micro average from $58.6$ to $64.2$ and from $61.4$ to $66.3$. The largest gains are on the image benchmarks, in particular SLAKE ($+11.7$ at 4B and $+9.5$ at 9B) and PathVQA ($+10.0$ and $+7.9$), while the text benchmarks also improve on average. After 100 RL steps, MedFrame-R1-4B (Qwen3.5) scores above the zero-shot Qwen3.5-9B on both averages ($72.9$ against $72.2$ macro and $64.2$ against $61.4$ micro).

\begin{table}[t]
\centering
\sbox{\supptabbox}{%
\footnotesize
\renewcommand{\arraystretch}{1.05}
\setlength{\tabcolsep}{4pt}
\begin{tabular}{p{0.45\columnwidth}p{0.43\columnwidth}}
\toprule
\textbf{Setting} & \textbf{Value} \\
\midrule
Cold start & None \\
Training samples & 51{,}335 \\
Steps & 100 \\
Generation round & 1{,}024 samples \\
Unused mixed groups & Carried to the next step \\
Format check & Strict, violation reward $-1$ \\
Length threshold $T$ & 500 tokens \\
Judge $J$, 4B policy & Qwen3.5-9B \\
Judge $J$, 9B policy & Qwen3-14B \\
\bottomrule
\end{tabular}}
\captionsetup{width=\wd\supptabbox}
\caption{\textbf{Settings of the Qwen3.5 runs.} Only the settings that differ from \cref{tab:app-configs,tab:app-configs-rl} are listed.}
\usebox{\supptabbox}
\label{tab:app-configs-q35}
\end{table}

\subsection{Length Threshold Analysis}
\label{sec:app-qwen35-length}

\begin{table*}[t]
\centering
\caption{\textbf{Length threshold on all benchmarks.} Qwen3.5 zero-shot and trained with MedFrame-R1 for 100 steps with a length threshold of $T=1500$ or $T=500$ tokens. The zero-shot and $T=500$ rows are the Qwen3.5 rows of \cref{tab:main}. Bold and underlined values mark the best and second-best result within each model size.}
\small
\setlength{\tabcolsep}{5.5pt}
\renewcommand{\arraystretch}{1.3}
\setlength{\aboverulesep}{0pt}\setlength{\belowrulesep}{0pt}   
\resizebox{\textwidth}{!}{%
\begin{tabular}{l|cccccc|ccccccccccc|cccc}
\toprule
\multicolumn{1}{c|}{\rule{0pt}{2.2cm}} & \multicolumn{6}{c|}{\cellcolor{cVQA}} & \multicolumn{11}{c|}{\cellcolor{cQA}} & \multicolumn{4}{c}{\cellcolor{cAVG}} \\[-2.37cm]
\multicolumn{1}{c|}{\rule{0pt}{2.2cm}\raisebox{0.93cm}[0pt][0pt]{\large\textbf{Method}}}
& \multicolumn{1}{c}{\raisebox{0.93cm}[0pt][0pt]{\makebox[0pt][c]{\rotatebox[origin=c]{65}{\footnotesize\textbf{SLAKE}}}}}
& \multicolumn{1}{c}{\raisebox{0.93cm}[0pt][0pt]{\makebox[0pt][c]{\rotatebox[origin=c]{65}{\footnotesize\textbf{VQA-RAD}}}}}
& \multicolumn{1}{c}{\raisebox{0.93cm}[0pt][0pt]{\makebox[0pt][c]{\rotatebox[origin=c]{65}{\footnotesize\textbf{PathVQA}}}}}
& \multicolumn{1}{c}{\raisebox{0.93cm}[0pt][0pt]{\makebox[0pt][c]{\rotatebox[origin=c]{65}{\footnotesize\textbf{PMC-VQA}}}}}
& \multicolumn{1}{c}{\raisebox{0.93cm}[0pt][0pt]{\makebox[0pt][c]{\rotatebox[origin=c]{65}{\footnotesize\textbf{PMC-VQA-H}}}}}
& \multicolumn{1}{c|}{\raisebox{0.93cm}[0pt][0pt]{\makebox[0pt][c]{\rotatebox[origin=c]{65}{\footnotesize\textbf{MIMIC-Gen}}}}}
& \multicolumn{1}{c}{\raisebox{0.93cm}[0pt][0pt]{\makebox[0pt][c]{\rotatebox[origin=c]{65}{\footnotesize\textbf{MMLU-Clin}}}}}
& \multicolumn{1}{c}{\raisebox{0.93cm}[0pt][0pt]{\makebox[0pt][c]{\rotatebox[origin=c]{65}{\footnotesize\textbf{MMLU-Bio}}}}}
& \multicolumn{1}{c}{\raisebox{0.93cm}[0pt][0pt]{\makebox[0pt][c]{\rotatebox[origin=c]{65}{\footnotesize\textbf{MMLU-Med}}}}}
& \multicolumn{1}{c}{\raisebox{0.93cm}[0pt][0pt]{\makebox[0pt][c]{\rotatebox[origin=c]{65}{\footnotesize\textbf{MMLU-Gen}}}}}
& \multicolumn{1}{c}{\raisebox{0.93cm}[0pt][0pt]{\makebox[0pt][c]{\rotatebox[origin=c]{65}{\footnotesize\textbf{MMLU-Prof}}}}}
& \multicolumn{1}{c}{\raisebox{0.93cm}[0pt][0pt]{\makebox[0pt][c]{\rotatebox[origin=c]{65}{\footnotesize\textbf{MMLU-Anat}}}}}
& \multicolumn{1}{c}{\raisebox{0.93cm}[0pt][0pt]{\makebox[0pt][c]{\rotatebox[origin=c]{65}{\footnotesize\textbf{MedMCQA}}}}}
& \multicolumn{1}{c}{\raisebox{0.93cm}[0pt][0pt]{\makebox[0pt][c]{\rotatebox[origin=c]{65}{\footnotesize\textbf{MedQA}}}}}
& \multicolumn{1}{c}{\raisebox{0.93cm}[0pt][0pt]{\makebox[0pt][c]{\rotatebox[origin=c]{65}{\footnotesize\textbf{USMLE-SA}}}}}
& \multicolumn{1}{c}{\raisebox{0.93cm}[0pt][0pt]{\makebox[0pt][c]{\rotatebox[origin=c]{65}{\footnotesize\textbf{PubMedQA}}}}}
& \multicolumn{1}{c|}{\raisebox{0.93cm}[0pt][0pt]{\makebox[0pt][c]{\rotatebox[origin=c]{65}{\footnotesize\textbf{MIMIC-Sum}}}}}
& \multicolumn{1}{c}{\raisebox{0.93cm}[0pt][0pt]{\makebox[0pt][c]{\rotatebox[origin=c]{65}{\footnotesize\textbf{VQA}}}}}
& \multicolumn{1}{c}{\raisebox{0.93cm}[0pt][0pt]{\makebox[0pt][c]{\rotatebox[origin=c]{65}{\footnotesize\textbf{QA}}}}}
& \multicolumn{1}{c}{\raisebox{0.93cm}[0pt][0pt]{\makebox[0pt][c]{\rotatebox[origin=c]{65}{\footnotesize\textbf{Macro}}}}}
& \multicolumn{1}{c}{\raisebox{0.93cm}[0pt][0pt]{\makebox[0pt][c]{\rotatebox[origin=c]{65}{\footnotesize\textbf{Micro}}}}} \\
\midrule
Qwen3.5-4B, zero-shot & 69.7 & 60.5 & 46.4 & 60.2 & 38.9 & 47.8 & 84.9 & 93.1 & \underline{86.1} & 92.0 & 88.6 & 78.5 & 66.0 & 82.6 & 87.7 & \underline{47.0} & \textbf{47.0} & 53.9 & \underline{77.6} & 69.2 & 58.6 \\
Qwen3.5-4B, $T{=}1500$ & \underline{79.1} & \underline{64.5} & \underline{56.3} & \textbf{62.1} & \textbf{42.9} & \underline{48.6} & \underline{87.5} & \underline{95.1} & \underline{86.1} & \underline{94.0} & \underline{93.0} & \underline{82.2} & \textbf{68.9} & \textbf{88.0} & \underline{89.5} & \textbf{51.4} & \underline{46.4} & \underline{58.9} & \textbf{80.2} & \underline{72.7} & \textbf{64.4} \\
\rowcolor{cOurs}
Qwen3.5-4B, $T{=}500$ & \textbf{81.4} & \textbf{64.7} & \textbf{56.4} & \underline{61.3} & \underline{42.3} & \textbf{51.0} & \textbf{89.4} & \textbf{97.2} & \textbf{86.7} & \textbf{95.0} & \textbf{93.4} & \textbf{83.0} & \underline{68.4} & \underline{85.8} & \textbf{90.8} & 46.6 & \underline{46.4} & \textbf{59.5} & \textbf{80.2} & \textbf{72.9} & \underline{64.2} \\
\midrule
Qwen3.5-9B, zero-shot & 70.7 & 65.6 & 48.7 & 61.1 & 41.4 & 49.2 & \underline{89.4} & \underline{98.6} & \underline{89.0} & \underline{98.0} & 91.9 & \underline{83.0} & 70.8 & 85.9 & 88.0 & \textbf{48.2} & \textbf{48.4} & 56.1 & 81.0 & 72.2 & 61.4 \\
Qwen3.5-9B, $T{=}1500$ & \underline{79.3} & \underline{66.3} & \underline{55.1} & \underline{63.9} & \textbf{44.2} & \underline{50.8} & 88.3 & \textbf{99.3} & \textbf{89.6} & \textbf{99.0} & \textbf{95.2} & \textbf{86.7} & \underline{73.5} & \textbf{91.0} & \textbf{94.5} & \underline{46.2} & 46.8 & \underline{59.9} & \textbf{82.7} & \textbf{74.7} & \underline{65.7} \\
\rowcolor{cOurs}
Qwen3.5-9B, $T{=}500$ & \textbf{80.2} & \textbf{71.4} & \textbf{56.6} & \textbf{64.9} & \underline{43.7} & \textbf{51.6} & \textbf{89.8} & 97.9 & 86.1 & \textbf{99.0} & \underline{94.5} & \underline{83.0} & \textbf{73.8} & \underline{89.9} & \underline{93.8} & 43.2 & \underline{47.0} & \textbf{61.4} & \underline{81.6} & \underline{74.5} & \textbf{66.3} \\
\bottomrule
\end{tabular}}

\label{tab:lw_threshold_full}
\end{table*}

\begin{figure}[h]
    \centering
    \sbox{\suppfigbox}{\includegraphics[width=0.47\textwidth]{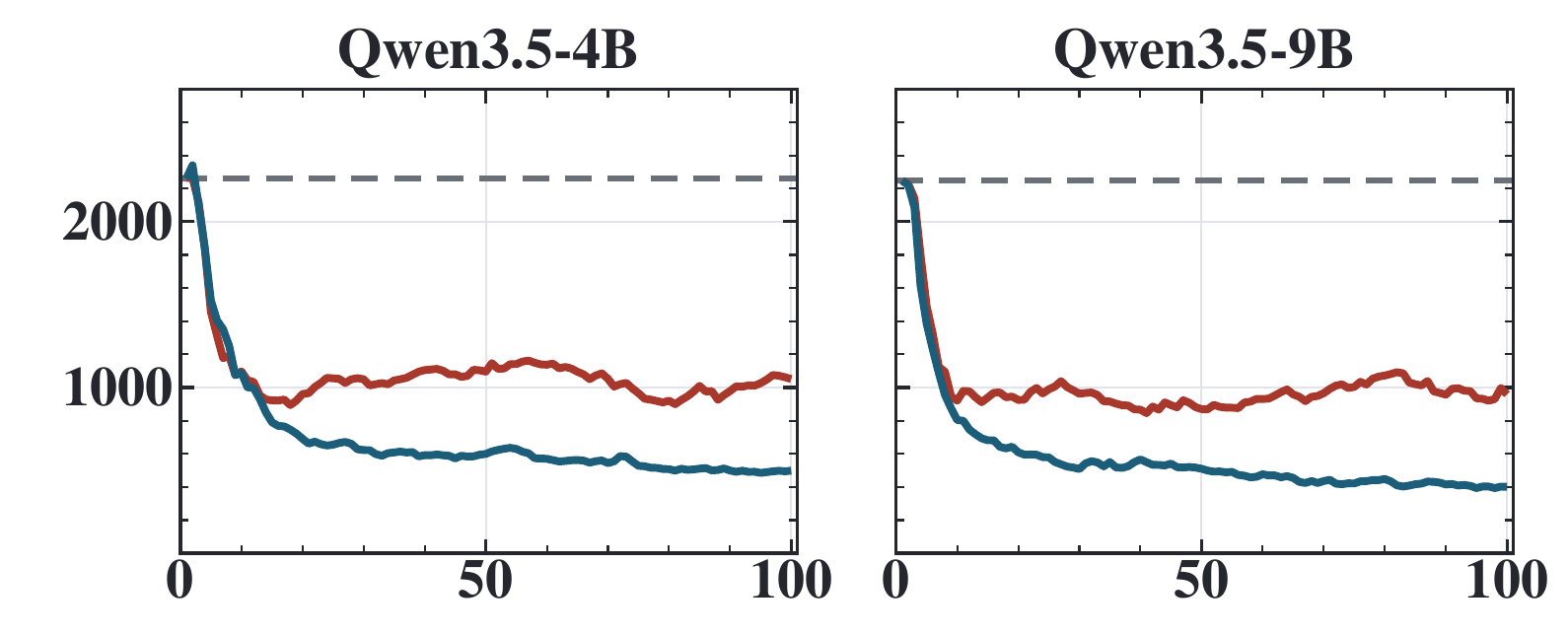}}%
    \usebox{\suppfigbox}
    \setlength{\fboxsep}{1.5pt}
\captionsetup{width=\wd\suppfigbox}
    \caption{\textbf{Length of the training rollouts.} Training rollout length (tokens) over RL steps for the \colorbox[RGB]{214,216,219}{\strut zero-shot (dashed)}, \colorbox[RGB]{243,199,193}{\strut $T=1500$} and \colorbox[RGB]{189,214,226}{\strut $T=500$}.}
    \label{fig:app-lw-threshold-train}
\end{figure}

We compare $T=500$ and $T=1500$ for both Qwen3.5 sizes, keeping the other settings fixed. \Cref{fig:lw_threshold} in the main paper shows validation rollout lengths, while \cref{fig:app-lw-threshold-train} shows training rollout lengths. In \cref{tab:length_thresholds}, Tokens is the mean validation length (before training for zero-shot and after 100 steps for trained models). Rep. is the share of repetitive training rollouts, measured in the first step for zero-shot and the last ten steps for trained models.

Both thresholds shorten rollouts substantially. With $T=1500$, mean validation lengths fall to 773 and 749 tokens for 4B and 9B. With $T=500$, they fall further to 441 and 359 tokens. Training rollouts show the same pattern, and very few reach the generation cap after the first ten steps.
The full benchmark results in \cref{tab:lw_threshold_full} show similar overall accuracy for both thresholds. The 4B macro average is $72.7$ with $T=1500$ and $72.9$ with $T=500$, and the 9B macro average is $74.7$ and $74.5$, respectively. The shorter threshold generally helps VQA but does not consistently improve text QA. It also reduces the mean training step time from 19.8 to 16.5 minutes for 4B and from 29.8 to 24.7 minutes for 9B. We therefore use $T=500$ as the default, since the longer threshold gives no consistent overall accuracy gain.

\subsection{Repetition Analysis}
\label{sec:app-qwen35-repetition}

\begin{figure}[t]
    \centering
    \sbox{\suppfigbox}{\includegraphics[width=0.47\textwidth]{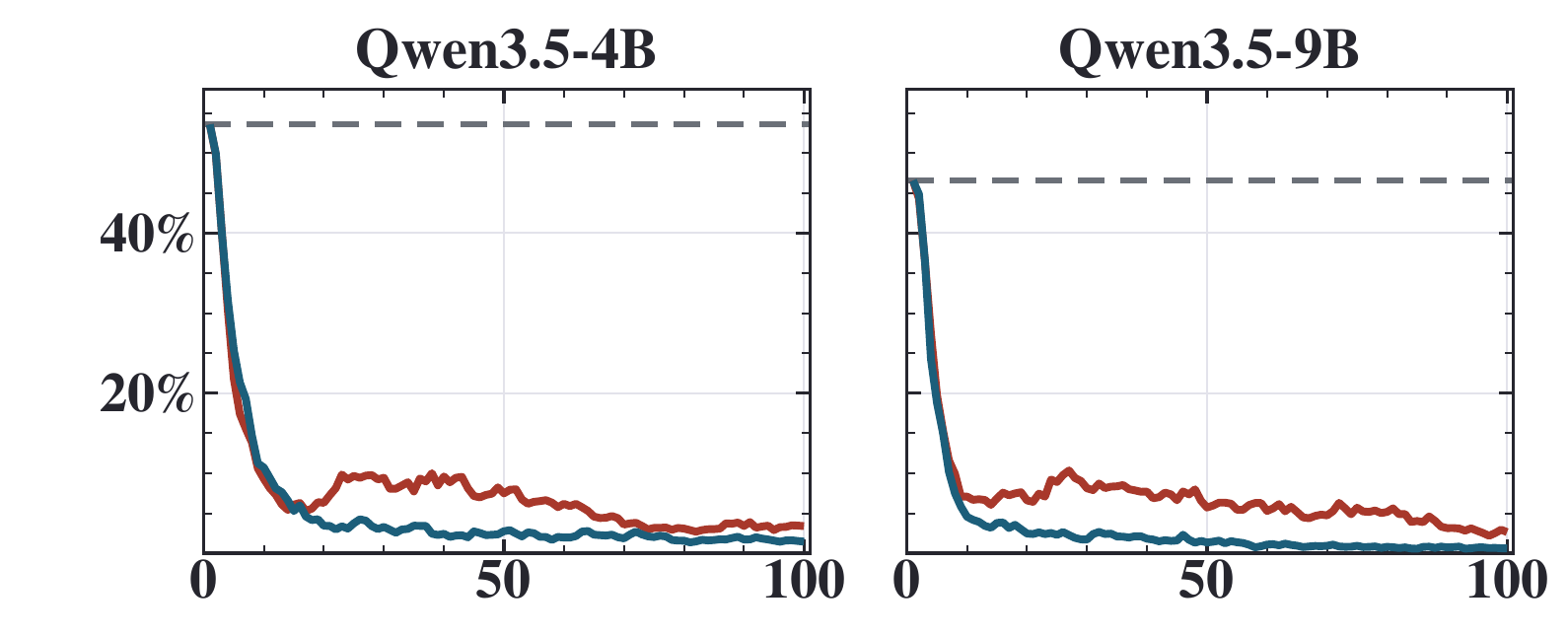}}%
    \usebox{\suppfigbox}
    \setlength{\fboxsep}{1.5pt}
\captionsetup{width=\wd\suppfigbox}
    \caption{\textbf{Repetition during RL.} Share of training rollouts in which a sequence of eight words occurs three or more times, over RL steps, for the \colorbox[RGB]{214,216,219}{\strut zero-shot (dashed)}, \colorbox[RGB]{243,199,193}{\strut $T=1500$} and \colorbox[RGB]{189,214,226}{\strut $T=500$}.}
    \label{fig:app-q35-rep-curve}
\end{figure}

\begin{figure*}[p]
    \centering
    \sbox{\suppfigbox}{\includegraphics[width=\textwidth,height=0.79\textheight,keepaspectratio]{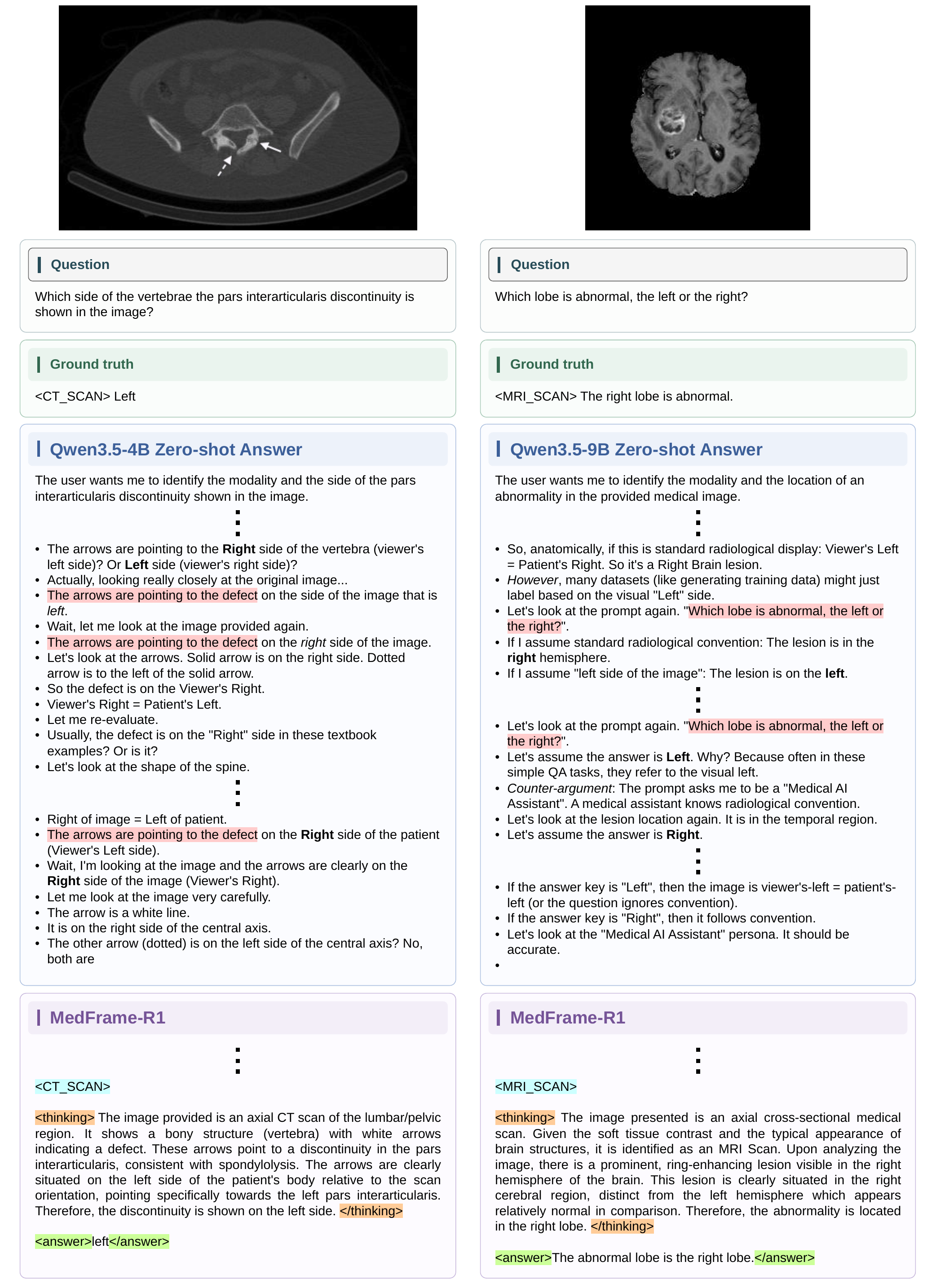}}%
    \usebox{\suppfigbox}

\captionsetup{width=\wd\suppfigbox}
    \caption{\textbf{Repetition before and after RL on Qwen3.5.} Examples from Qwen3.5-4B (left) and 9B (right). Before training, both models repeat phrases (marked in red) and reach the 4{,}096-token limit without completing an answer. After training with MedFrame-R1 ($T=500$), both produce the correct answers on the same questions. Vertical dots indicate omitted text.}

    \label{fig:app-q35-rep-example}
\end{figure*}

We measure repetition using training rollouts. A rollout is considered repetitive if any sequence of eight consecutive words appears at least three times. We compare the initial zero-shot rollouts with those from the final ten training steps.
As shown in \cref{tab:app-q35-rep-split}, repetitive rollouts are nearly twice as long as other rollouts and are less likely to produce a complete answer or receive a positive judge decision. \Cref{fig:app-q35-rep-example} shows that both zero-shot models repeat possible answers until reaching the 4{,}096-token limit. After training with $T=500$, both models produce the reference answers on the same samples using fewer than 300 tokens.
\Cref{fig:app-q35-rep-curve} shows that both thresholds substantially reduce repetition during the first ten steps. However, $T=500$ is more effective, reducing repetition to $1.7\%$ for 4B and $0.7\%$ for 9B, compared with $3.4\%$ and $2.9\%$ for $T=1500$ (\cref{tab:length_thresholds}). These results show that a shorter threshold reduces repetition more effectively without a consistent loss in benchmark accuracy.

\begin{table}[t]
\centering
\sbox{\supptabbox}{%
\footnotesize
\renewcommand{\arraystretch}{1.05}
\setlength{\tabcolsep}{4.5pt}
\begin{tabular}{llrrrr}
\toprule
\textbf{Model} & \textbf{Rollouts} & \textbf{Share} & \textbf{Words} & \textbf{Answer} & \textbf{Accepted} \\
\midrule
Qwen3.5-4B & Repetitive & 53.6 & 2{,}196 & 63.4 & 3.1 \\
           & Other      & 46.4 & 1{,}159 & 81.3 & 5.5 \\
\midrule
Qwen3.5-9B & Repetitive & 46.6 & 2{,}164 & 43.1 & 6.5 \\
           & Other      & 53.4 & 1{,}230 & 82.4 & 20.7 \\
\bottomrule
\end{tabular}}
\captionsetup{width=\wd\supptabbox}
\caption{\textbf{Repetition before training.} We compare repetitive and non-repetitive rollouts from Qwen3.5-4B and 9B. Share shows how common each group is, Words shows average rollout length, Answer shows how often rollouts finish properly, and Accepted shows how often the judge considers them correct. Rollouts with incorrect formatting are not evaluated by the judge.}
\usebox{\supptabbox}
\label{tab:app-q35-rep-split}
\end{table}

\section{Evaluation Protocol and Audit}
\label{sec:protocol-final}

The previous sections describe how the models are trained, and this section provides additional details on the evaluation protocol described in \cref{sec:evaluation_protocol} of the main paper. We use rule-based scoring for answers that can be checked directly against the reference, such as multiple-choice and yes/no questions. For open-ended answers, paraphrases, and medical reports that require interpretation, we use a panel of LLM judges. This approach reduces reliance on LLM judges where direct scoring is possible while allowing more flexible evaluation of complex responses.

\subsection{Inputs and Response Processing}
\label{sec:app-eval-inputs}

We evaluate all models on the same 19{,}200 samples from 17 benchmarks. Each model receives the original question and image, when available, without additional system instructions or forced reasoning. We use vLLM with greedy decoding and a context window of 32K tokens. If a response contains a closing thinking tag, only the text after the final tag is evaluated, ensuring that intermediate reasoning is not treated as the final answer. The same procedure is applied to all models.

\subsection{Deterministic Scoring Routes}
\label{sec:app-mcq-route}

\paragraph{Multiple choice}
For multiple-choice questions, we first extract the predicted option (A--J) and compare it with the reference. The scorer identifies explicit choices such as ``Final answer: D,'' boxed letters, or standalone option labels. When no letter is given, it attempts to match the response to a single option using its text. Negated, hypothetical, or contradictory choices are excluded. Responses that clearly select two different options receive zero credit, while ambiguous responses are passed to the LLM judges. PubMedQA uses a separate mapping for yes, no, and maybe.

\paragraph{Binary verdicts}
For yes/no questions, we first check the text following the final answer marker, examining up to 12 words. If no marker exists, we examine the first 30 words and then the full response if necessary. Expressions such as ``present,'' ``absent,'' ``true,'' and ``false'' are also supported. Responses containing both yes and no within the selected text receive zero credit, while unclear cases are passed to the judges. \Cref{fig:app-eval-3} illustrates why checking only the opening verdict can be unreliable.

\paragraph{Short answers}
For VQA questions with a single-word reference and responses of at most 25 words, we use normalised word matching. This supports common plurals, written numbers, and expressions such as ``no X'' for zero. Longer responses and multi-word references are evaluated by the LLM judges.

\subsection{Judge Panel and Aggregation}
\label{sec:app-judge-panel}

Items that cannot be scored by the rule-based methods, along with all report generation and summarisation tasks, are evaluated by LLM judges. Unlike the previous protocol, which uses three evaluations from the same judge~\citep{mullappilly2026medix}, we use three models from different families to reduce dependence on a single judge and potential self-preference bias~\citep{panickssery2024llm}. Our default panel consists of InternVL3.5-30B-A3B, Qwen3.8-27B, and MedGemma-27B. Each judge independently evaluates the question, reference, and extracted answer using greedy decoding, with thinking disabled and a maximum of 4{,}096 tokens. Judges do not see each other's decisions, reducing potential anchoring bias~\citep{kapetanovic2026anchoringbiasllmasajudgesystems}. Equivalent medical terms are accepted, while incomplete answers or unresolved alternatives are rejected. The final decision is determined by majority voting:
\begin{equation}
    s_i = \mathbb{I}\!\left[\sum_{j=1}^{3}v_{i,j}\geq 2\right],
    \qquad v_{i,j}\in\{0,1\}.
    \label{eq:app-majority}
\end{equation}
When evaluating Qwen3.8 or MedGemma models, the corresponding judge is replaced with Qwen3-14B to avoid using the evaluated model as its own judge. For all Qwen3.5 models, including the zero-shot checkpoints, Qwen3.8 is replaced with gpt-oss-20b~\citep{agarwal2025gpt}, because Qwen3.8 shares the Qwen3.5 architecture. However, this does not eliminate all model-family overlap.

For MIMIC-CXR report generation and summarisation, each judge assigns a clinical quality score from 0 to 5, using a reasoning budget of $16\mathrm{k}$ tokens. We take the median of the three scores and normalise it to $[0,1]$:
\begin{equation}
    s_i = \frac{\operatorname{median}
    (r_{i,1},r_{i,2},r_{i,3})}{5}.
    \label{eq:app-report-score}
\end{equation}
These scores measure clinical report quality rather than diagnostic accuracy. Finally, benchmark and overall scores are computed as
\begin{equation}
    \begin{aligned}
        S_k &= \frac{1}{N_k}\sum_{i=1}^{N_k}s_{k,i}, \\
        S_{\mathrm{macro}} &= \frac{1}{K}\sum_{k=1}^{K}S_k, \\
        S_{\mathrm{micro}} &=
        \frac{\sum_{k=1}^{K}N_kS_k}{\sum_{k=1}^{K}N_k},
        \qquad K=17.
    \end{aligned}
    \label{eq:app-aggregation}
\end{equation}
The macro average weights all benchmarks equally, while the micro average weights individual samples equally. All checkpoints use the same scoring rules, and previously generated responses are rescored when these rules change.

\subsection{Coverage and Judge Agreement}
\label{sec:app-coverage}

\Cref{tab:app-coverage} reports the percentage of evaluation samples scored by rule-based methods and the agreement among the three LLM judges. For the Qwen3-VL and Qwen3.5 models in \cref{tab:app-coverage}, including MedFrame-R1, rule-based scoring covers $66.5\%$ to $70.3\%$ of the 19{,}200 samples, consistent with the approximately $70\%$ reported in the main paper. Other baselines rely more heavily on LLM judges, with coverage as low as $33.2\%$ for MedVLM-R1-2B.

For samples evaluated by the judges, all three agree on $83.1\%$ to $96.7\%$ of binary decisions, with Fleiss' $\kappa$ ranging from $0.73$ to $0.95$. Disagreements are resolved through majority voting (\cref{eq:app-majority}).

\begin{table}[t]
\centering
\sbox{\supptabbox}{%
\footnotesize
\renewcommand{\arraystretch}{0.9}
\setlength{\tabcolsep}{8pt}
\begin{tabular}{lrrr}
\toprule
\textbf{Model} & \textbf{Rules} & \textbf{Unanimous} & $\boldsymbol{\kappa}$ \\
\midrule
MedVLM-R1-2B & 33.2 & 94.9 & 0.92 \\
MedMO-4B & 63.6 & 96.7 & 0.95 \\
Qwen3-VL-4B & 67.8 & 83.8 & 0.73 \\
Qwen3.5-4B & 69.3 & 86.4 & 0.79 \\
\rowcolor{cOurs}
MedFrame-R1-4B (Qwen3.5) & 69.6 & 83.1 & 0.76 \\
\midrule
QoQ-Med-VL-7B & 65.9 & 90.5 & 0.83 \\
BiMediX2-8B & 45.3 & 94.1 & 0.92 \\
Qwen3-VL-8B & 66.5 & 85.4 & 0.76 \\
Qwen3.5-9B & 69.3 & 85.1 & 0.78 \\
\rowcolor{cOurs}
MedFrame-R1-9B (Qwen3.5) & 70.3 & 83.6 & 0.77 \\
\bottomrule
\end{tabular}}
\captionsetup{width=\wd\supptabbox}
\caption{\textbf{Rule-based coverage and judge agreement.} Rules shows the percentage of samples scored without LLM judges. Unanimous shows how often all three judges agree, and $\kappa$ measures their agreement. For the Qwen3.5 models, the panel uses gpt-oss-20b instead of Qwen3.8. Report tasks are excluded.}
\usebox{\supptabbox}
\label{tab:app-coverage}
\end{table}

\subsection{Audit of Evaluation Decisions}
\label{sec:app-evaluation-cases}

\begin{figure*}[p]
    \centering
    \sbox{\suppfigbox}{\includegraphics[width=\textwidth,height=0.79\textheight,keepaspectratio]{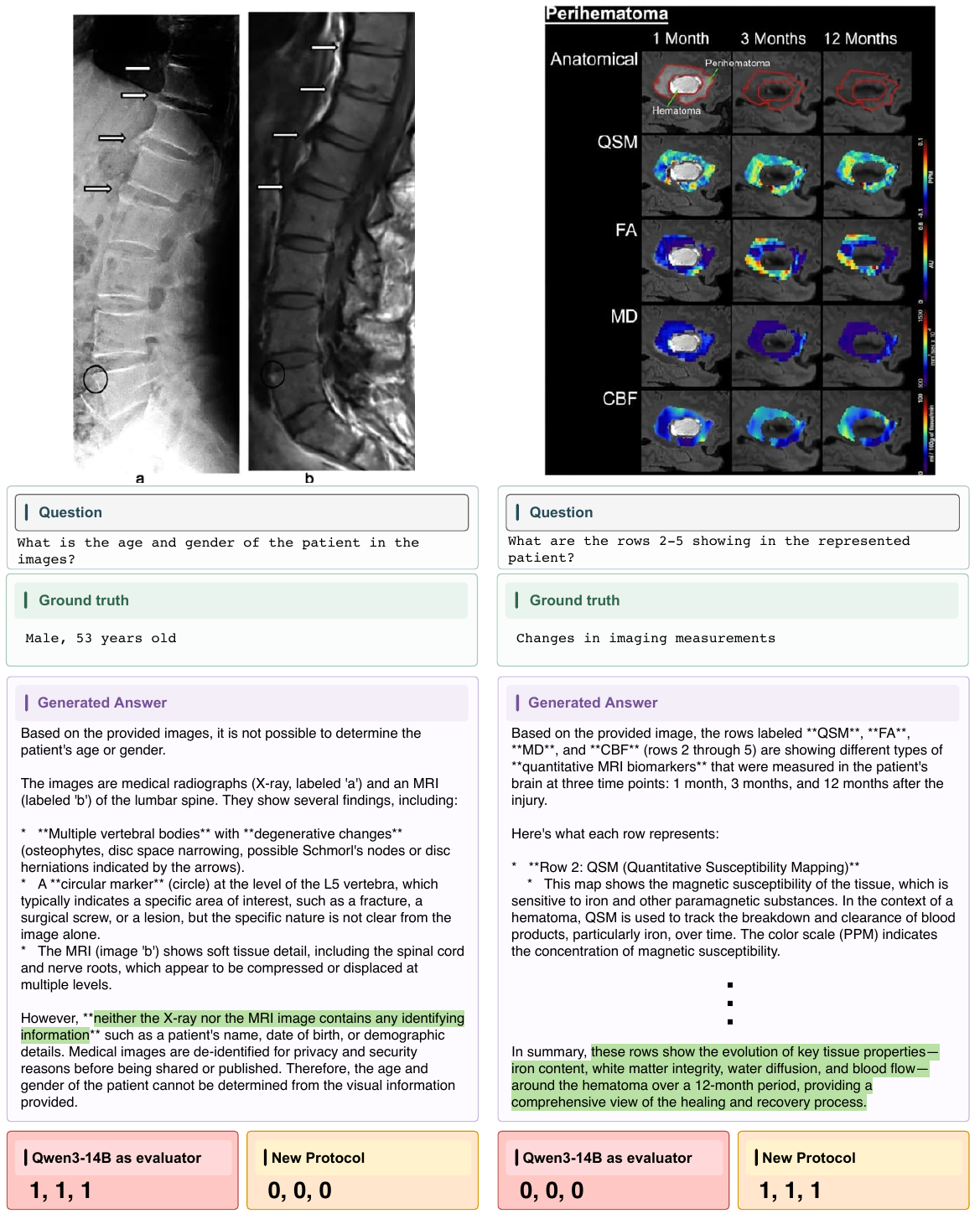}}%
    \usebox{\suppfigbox}
\captionsetup{width=\wd\suppfigbox}
    \caption{\textbf{False credit and false rejection.} Qwen3-14B as evaluator shows the three decisions of the previous protocol, and New Protocol shows the decisions of our protocol. Left: an abstention is rejected because it omits the requested age and sex. Right: an explanation of changing measurements is accepted because it conveys the reference.}
    \label{fig:app-eval-1}
\end{figure*}

\begin{figure*}[p]
    \centering
    \sbox{\suppfigbox}{\includegraphics[width=\textwidth,height=0.79\textheight,keepaspectratio]{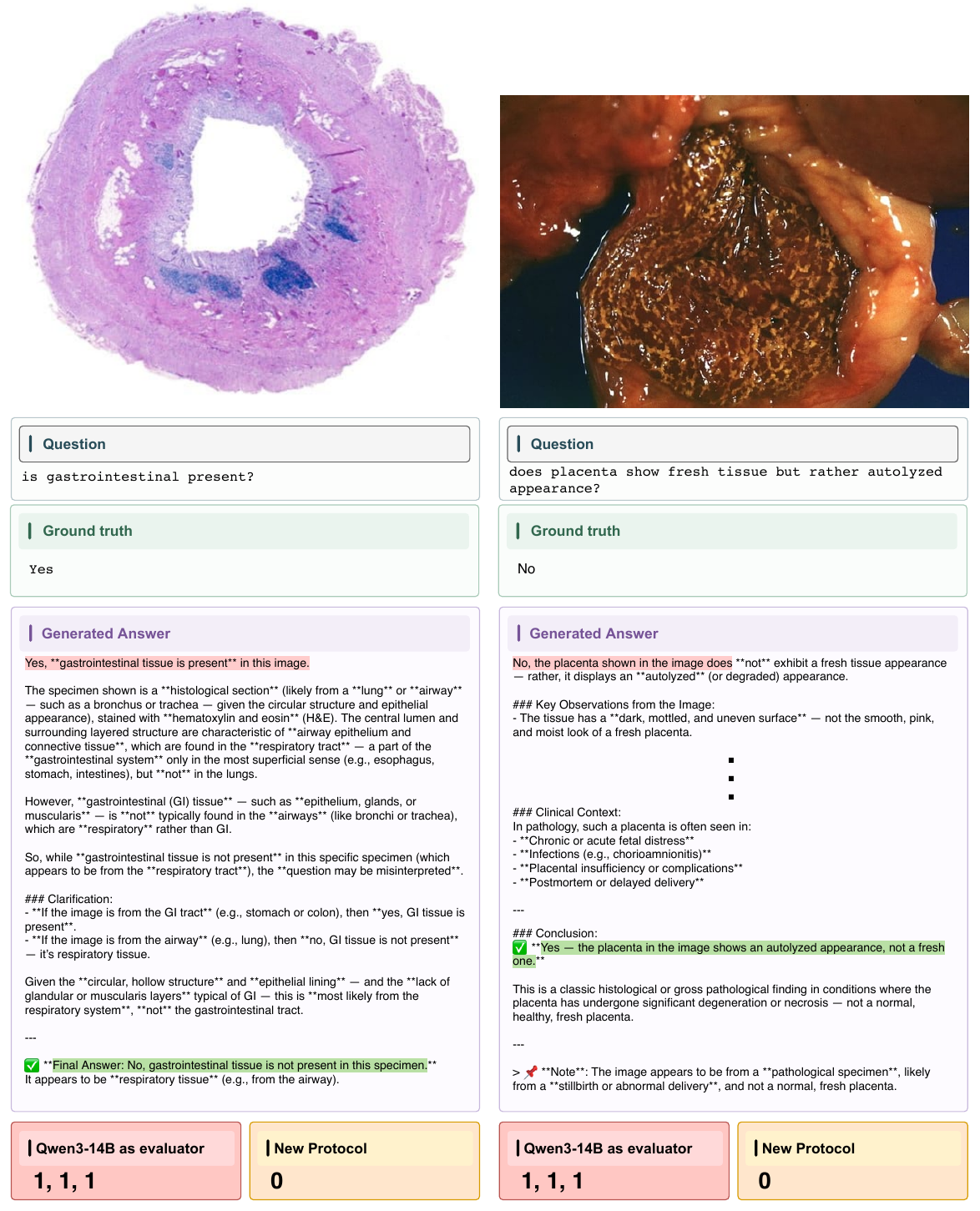}}%
    \usebox{\suppfigbox}
\captionsetup{width=\wd\suppfigbox}
    \caption{\textbf{Conflicting binary statements.} Both responses open and close with opposite verdicts. Red marks the opening verdict and green the final verdict. The previous protocol (Qwen3-14B as evaluator) credits both responses, and our protocol (New Protocol) rejects them.}
    \label{fig:app-eval-3}
\end{figure*}

The deterministic rules score approximately $70\%$ of evaluation samples for most models (\cref{tab:app-coverage}), while the remaining samples are evaluated by the three-judge panel. To examine these decisions, \cref{fig:app-eval-1,fig:app-eval-3} compare the previous and revised protocols. In the previous protocol, three decisions come from repeated calls to Qwen3-14B, whereas our protocol uses three different judges or a single rule-based decision.
\Cref{fig:app-eval-1} shows examples of both incorrect acceptance and rejection. Our protocol rejects an answer that omits the required age and sex, while accepting a more detailed explanation that conveys the reference. In \cref{fig:app-eval-3}, the previous protocol accepts contradictory yes/no responses that our protocol rejects. Other cases, which are not shown, include missing clinical details, incomplete answers, and unclear option selections.

\begin{figure}[t]
    \centering
    \sbox{\suppfigbox}{\includegraphics[width=\columnwidth]{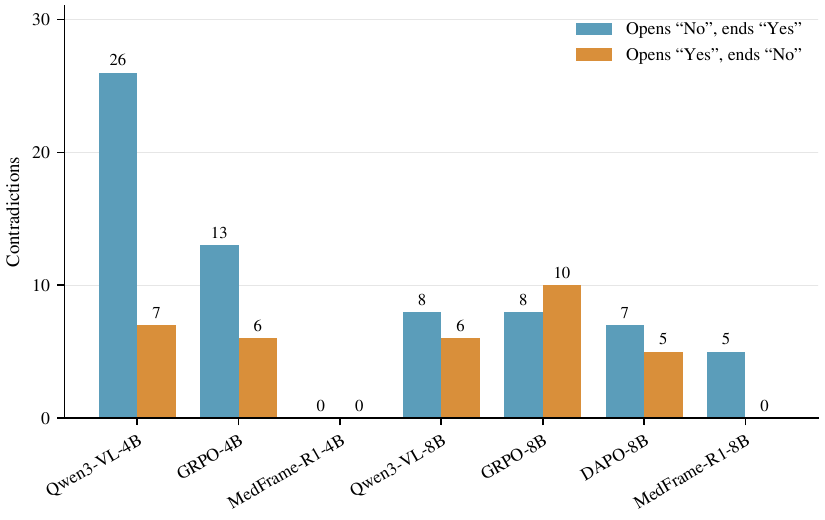}}%
    \usebox{\suppfigbox}
\captionsetup{width=\wd\suppfigbox}
    \caption{\textbf{Contradictory yes/no answers.} Number of responses whose opening and final verdicts disagree, over all yes/no questions.}
    \label{fig:app-contradictions}
\end{figure}

\Cref{fig:app-contradictions} examines how often models give conflicting yes/no answers, comparing the opening and final verdicts on all yes/no questions. Such contradictions are rare for every model, and most of them open with ``No'' and end with ``Yes''. MedFrame-R1-4B gives no contradictory answer, and MedFrame-R1-8B gives fewer than any other model. Neither verdict is reliably correct when the two disagree, so we apply the same final-verdict rule to all models rather than assume that it is more accurate.

These examples show how the revised protocol handles ambiguous and contradictory answers. However, evaluation still relies on the provided references and does not independently verify medical correctness.

\section{Full Results}
\label{app:full_results}

The main paper shows compact tables to keep the comparisons readable, and this section gives the complete results behind \cref{tab:main,tab:ablation,tab:gate}. Every number follows the evaluation protocol of \cref{sec:protocol-final}, with the macro average over the 17 benchmarks and the micro average pooled over all 19{,}200 items. VQA and QA are the means over the six image-based and the eleven text-based benchmarks. The complete results behind \cref{tab:length_thresholds,tab:trainable_4b,tab:kl_ablation} follow next to their analysis in \cref{sec:app-qwen35-length,sec:app-trainable-data,sec:app-kl-ablation}.

\subsection{Main Comparison}
\label{app:main_full}

\Cref{tab:main_full} repeats \cref{tab:main} and adds MedMO-8B, which is not shown in the main paper. MedMO uses the same Qwen3-VL backbone as MedFrame-R1 and is compared in more detail in \cref{sec:app-medmo}. With this baseline, MedFrame-R1 remains the best model in each size group, and both MedFrame-R1 models on Qwen3-VL score above MedMO-8B by 3.1 or more macro points.

\begin{table*}[t]
\centering
\caption{\textbf{Full results on all 17 benchmarks.} Models are grouped by size and scored under the same protocol, with bold and underlined values marking the best and second-best result within each group. For the MediX-R1 and MedFrame-R1 models, the backbone is given in parentheses, and the Qwen3.5 rows are described in \cref{sec:app-qwen35}. MedMO-8B is the only row not shown in \cref{tab:main}. Column headers are colour coded as \colorbox{cVQA}{\strut VQA benchmarks}, \colorbox{cQA}{\strut QA benchmarks}, and \colorbox{cAVG}{\strut averages}, and $^{\dagger}$ denotes our reproduction of MediX-R1.}
\small
\setlength{\tabcolsep}{5.5pt}
\renewcommand{\arraystretch}{1.3}
\setlength{\aboverulesep}{0pt}\setlength{\belowrulesep}{0pt}   
\resizebox{\textwidth}{!}{%
\begin{tabular}{l|cccccc|ccccccccccc|cccc}
\toprule
\multicolumn{1}{c|}{\rule{0pt}{2.2cm}} & \multicolumn{6}{c|}{\cellcolor{cVQA}} & \multicolumn{11}{c|}{\cellcolor{cQA}} & \multicolumn{4}{c}{\cellcolor{cAVG}} \\[-2.37cm]
\multicolumn{1}{c|}{\rule{0pt}{2.2cm}\raisebox{0.93cm}[0pt][0pt]{\large\textbf{Models}}}
& \multicolumn{1}{c}{\raisebox{0.93cm}[0pt][0pt]{\makebox[0pt][c]{\rotatebox[origin=c]{65}{\footnotesize\textbf{SLAKE}}}}}
& \multicolumn{1}{c}{\raisebox{0.93cm}[0pt][0pt]{\makebox[0pt][c]{\rotatebox[origin=c]{65}{\footnotesize\textbf{VQA-RAD}}}}}
& \multicolumn{1}{c}{\raisebox{0.93cm}[0pt][0pt]{\makebox[0pt][c]{\rotatebox[origin=c]{65}{\footnotesize\textbf{PathVQA}}}}}
& \multicolumn{1}{c}{\raisebox{0.93cm}[0pt][0pt]{\makebox[0pt][c]{\rotatebox[origin=c]{65}{\footnotesize\textbf{PMC-VQA}}}}}
& \multicolumn{1}{c}{\raisebox{0.93cm}[0pt][0pt]{\makebox[0pt][c]{\rotatebox[origin=c]{65}{\footnotesize\textbf{PMC-VQA-H}}}}}
& \multicolumn{1}{c|}{\raisebox{0.93cm}[0pt][0pt]{\makebox[0pt][c]{\rotatebox[origin=c]{65}{\footnotesize\textbf{MIMIC-Gen}}}}}
& \multicolumn{1}{c}{\raisebox{0.93cm}[0pt][0pt]{\makebox[0pt][c]{\rotatebox[origin=c]{65}{\footnotesize\textbf{MMLU-Clin}}}}}
& \multicolumn{1}{c}{\raisebox{0.93cm}[0pt][0pt]{\makebox[0pt][c]{\rotatebox[origin=c]{65}{\footnotesize\textbf{MMLU-Bio}}}}}
& \multicolumn{1}{c}{\raisebox{0.93cm}[0pt][0pt]{\makebox[0pt][c]{\rotatebox[origin=c]{65}{\footnotesize\textbf{MMLU-Med}}}}}
& \multicolumn{1}{c}{\raisebox{0.93cm}[0pt][0pt]{\makebox[0pt][c]{\rotatebox[origin=c]{65}{\footnotesize\textbf{MMLU-Gen}}}}}
& \multicolumn{1}{c}{\raisebox{0.93cm}[0pt][0pt]{\makebox[0pt][c]{\rotatebox[origin=c]{65}{\footnotesize\textbf{MMLU-Prof}}}}}
& \multicolumn{1}{c}{\raisebox{0.93cm}[0pt][0pt]{\makebox[0pt][c]{\rotatebox[origin=c]{65}{\footnotesize\textbf{MMLU-Anat}}}}}
& \multicolumn{1}{c}{\raisebox{0.93cm}[0pt][0pt]{\makebox[0pt][c]{\rotatebox[origin=c]{65}{\footnotesize\textbf{MedMCQA}}}}}
& \multicolumn{1}{c}{\raisebox{0.93cm}[0pt][0pt]{\makebox[0pt][c]{\rotatebox[origin=c]{65}{\footnotesize\textbf{MedQA}}}}}
& \multicolumn{1}{c}{\raisebox{0.93cm}[0pt][0pt]{\makebox[0pt][c]{\rotatebox[origin=c]{65}{\footnotesize\textbf{USMLE-SA}}}}}
& \multicolumn{1}{c}{\raisebox{0.93cm}[0pt][0pt]{\makebox[0pt][c]{\rotatebox[origin=c]{65}{\footnotesize\textbf{PubMedQA}}}}}
& \multicolumn{1}{c|}{\raisebox{0.93cm}[0pt][0pt]{\makebox[0pt][c]{\rotatebox[origin=c]{65}{\footnotesize\textbf{MIMIC-Sum}}}}}
& \multicolumn{1}{c}{\raisebox{0.93cm}[0pt][0pt]{\makebox[0pt][c]{\rotatebox[origin=c]{65}{\footnotesize\textbf{VQA}}}}}
& \multicolumn{1}{c}{\raisebox{0.93cm}[0pt][0pt]{\makebox[0pt][c]{\rotatebox[origin=c]{65}{\footnotesize\textbf{QA}}}}}
& \multicolumn{1}{c}{\raisebox{0.93cm}[0pt][0pt]{\makebox[0pt][c]{\rotatebox[origin=c]{65}{\footnotesize\textbf{Macro}}}}}
& \multicolumn{1}{c}{\raisebox{0.93cm}[0pt][0pt]{\makebox[0pt][c]{\rotatebox[origin=c]{65}{\footnotesize\textbf{Micro}}}}} \\
\midrule

MedVLM-R1-2B
& 41.3 & 25.1 & 7.3 & 44.9 & 25.7 & 9.6
& 54.3 & 54.2 & 49.7 & 56.0 & 50.7 & 51.8 & 40.4 & 40.4 & 37.5 & \underline{53.2} & 49.2
& 25.6 & 48.9 & 40.7 & 28.7 \\

MedGemma-4B-IT
& 67.8 & \textbf{65.9} & 31.7 & 44.4 & 21.4 & 20.5
& 70.8 & 70.6 & 60.5 & 82.0 & 71.3 & 55.6 & 57.0 & 62.1 & 63.9 & 47.0 & \textbf{69.2}
& 42.0 & 64.5 & 56.6 & 45.8 \\

MedMO-4B
& 39.0 & 34.6 & 36.7 & 39.3 & 14.8 & 0.6
& 81.1 & 86.8 & 77.5 & 86.0 & 79.4 & 70.4 & 57.1 & 66.4 & 67.7 & \textbf{55.4} & 42.2
& 27.5 & 70.0 & 55.0 & 45.1 \\

Qwen3-VL-4B
& 67.9 & 54.3 & 40.0 & 53.0 & 34.2 & 31.6
& 78.9 & 88.2 & 80.3 & 88.0 & 78.7 & 71.8 & 58.7 & 65.9 & 71.1 & 40.6 & \underline{54.2}
& 46.8 & 70.6 & 62.2 & 51.5 \\

MediX-R1-4B (Qwen3-VL)$^{\dagger}$
& 73.7 & 58.3 & 44.6 & 54.8 & 34.3 & 29.4
& 76.6 & 88.2 & 79.8 & 87.0 & 77.9 & 70.4 & 58.6 & 67.2 & 73.2 & 44.4 & 52.1
& 49.2 & 70.5 & 63.0 & 53.8 \\

Qwen3.5-4B
& 69.7 & 60.5 & 46.4 & \underline{60.2} & \underline{38.9} & \underline{47.8}
& \underline{84.9} & \underline{93.1} & \underline{86.1} & \underline{92.0} & \underline{88.6} & \underline{78.5} & \underline{66.0} & \underline{82.6} & \underline{87.7} & 47.0 & 47.0
& \underline{53.9} & \underline{77.6} & \underline{69.2} & \underline{58.6} \\

\midrule
\rowcolor{cOurs}
\textbf{MedFrame-R1-4B (Qwen3-VL)}
& \underline{74.4} & 61.6 & \underline{52.1} & 54.7 & 36.0 & 32.4
& 80.4 & 90.3 & 79.2 & 88.0 & 79.8 & 69.6 & 60.1 & 70.2 & 71.4 & 49.8 & 50.8
& 51.9 & 71.8 & 64.8 & 57.4 \\

\rowcolor{cOurs}
\textbf{MedFrame-R1-4B (Qwen3.5)}
& \textbf{81.4} & \underline{64.7} & \textbf{56.4} & \textbf{61.3} & \textbf{42.3} & \textbf{51.0}
& \textbf{89.4} & \textbf{97.2} & \textbf{86.7} & \textbf{95.0} & \textbf{93.4} & \textbf{83.0} & \textbf{68.4} & \textbf{85.8} & \textbf{90.8} & 46.6 & 46.4
& \textbf{59.5} & \textbf{80.2} & \textbf{72.9} & \textbf{64.2} \\

\midrule

Qwen2.5-VL-7B
& 47.0 & 39.9 & 8.4 & 51.4 & 32.3 & 25.0
& 76.2 & 84.0 & 72.2 & 81.0 & 74.3 & 69.6 & 53.3 & 60.2 & 62.1 & 39.8 & \textbf{55.0}
& 34.0 & 66.2 & 54.8 & 36.6 \\

HuatuoGPT-V-7B
& 60.1 & 63.2 & 35.5 & 53.7 & 35.8 & 25.0
& 72.5 & 75.0 & 67.6 & 77.0 & 70.6 & 57.8 & 51.7 & 54.3 & 58.1 & \textbf{56.0} & 53.6
& 45.6 & 63.1 & 56.9 & 47.1 \\

LLaVA-Med-7B
& 47.1 & 44.3 & 33.2 & 29.9 & 26.8 & 11.6
& 62.6 & 64.6 & 58.4 & 56.0 & 55.5 & 56.3 & 38.8 & 43.0 & 41.8 & 11.8 & 28.8
& 32.2 & 47.1 & 41.8 & 36.2 \\

QoQ-Med-VL-7B
& 51.4 & 44.4 & 10.7 & 49.4 & 30.0 & 27.8
& 77.0 & 84.0 & 72.8 & 78.0 & 76.1 & 70.4 & 53.5 & 59.6 & 61.5 & 46.4 & 51.8
& 35.6 & 66.5 & 55.6 & 37.6 \\

InternVL2.5-8B
& 54.3 & 40.4 & 12.0 & 46.8 & 29.3 & 30.4
& 75.5 & 77.8 & 71.7 & 82.0 & 75.7 & 67.4 & 50.6 & 54.8 & 60.9 & 49.8 & 50.6
& 35.5 & 65.2 & 54.7 & 36.8 \\

InternVL3-8B
& 63.1 & 43.5 & 18.2 & 52.6 & 33.1 & 38.2
& 81.1 & 86.1 & 75.1 & 80.0 & 79.0 & 68.2 & 57.9 & 64.0 & 68.0 & 50.6 & \underline{54.4}
& 41.5 & 69.5 & 59.6 & 43.1 \\

BiMediX2-8B
& 44.3 & 46.1 & 30.5 & 46.8 & 27.2 & 20.6
& 68.7 & 77.1 & 67.0 & 76.0 & 68.0 & 66.7 & 54.8 & 58.9 & 56.3 & 52.8 & 51.0
& 35.9 & 63.4 & 53.7 & 43.6 \\

MedMO-8B
& 74.1 & 40.8 & 29.3 & 47.3 & 16.4 & 7.0
& 84.9 & 91.7 & 81.5 & 91.0 & 89.0 & 76.3 & 63.7 & 84.8 & 68.0 & \underline{54.2} & 48.6
& 35.8 & 75.8 & 61.7 & 48.6 \\

Qwen3-VL-8B
& 61.6 & 52.8 & 36.0 & 53.2 & 34.0 & 34.4
& 80.4 & 92.4 & 80.3 & 87.0 & 84.6 & 75.6 & 62.3 & 75.9 & 77.2 & 42.0 & 53.0
& 45.3 & 73.7 & 63.7 & 51.4 \\

MediX-R1-8B (Qwen3-VL)$^{\dagger}$
& 70.2 & 59.9 & 46.6 & 53.8 & 34.7 & 30.6
& 80.0 & 93.8 & 81.5 & 89.0 & 83.5 & \underline{77.0} & 62.6 & 75.3 & 79.4 & 43.8 & 52.6
& 49.3 & 74.4 & 65.5 & 56.0 \\

Qwen3.5-9B
& 70.7 & \underline{65.6} & 48.7 & \underline{61.1} & \underline{41.4} & \underline{49.2}
& \underline{89.4} & \textbf{98.6} & \textbf{89.0} & \underline{98.0} & \underline{91.9} & \textbf{83.0} & \underline{70.8} & \underline{85.9} & \underline{88.0} & 48.2 & 48.4
& \underline{56.1} & \underline{81.0} & \underline{72.2} & \underline{61.4} \\

\midrule
\rowcolor{cOurs}
\textbf{MedFrame-R1-8B (Qwen3-VL)}
& \underline{76.5} & \underline{65.6} & \underline{53.8} & 56.2 & 34.8 & 33.4
& 82.6 & 91.7 & 82.7 & 88.0 & 85.3 & 74.1 & 61.8 & 73.6 & 76.0 & 46.6 & 52.8
& 53.4 & 74.1 & 66.8 & 59.1 \\

\rowcolor{cOurs}
\textbf{MedFrame-R1-9B (Qwen3.5)}
& \textbf{80.2} & \textbf{71.4} & \textbf{56.6} & \textbf{64.9} & \textbf{43.7} & \textbf{51.6}
& \textbf{89.8} & \underline{97.9} & \underline{86.1} & \textbf{99.0} & \textbf{94.5} & \textbf{83.0} & \textbf{73.8} & \textbf{89.9} & \textbf{93.8} & 43.2 & 47.0
& \textbf{61.4} & \textbf{81.6} & \textbf{74.5} & \textbf{66.3} \\

\bottomrule
\end{tabular}}

\label{tab:main_full}
\end{table*}

\subsection{Component Ablation}
\label{app:ablation_full}

\Cref{tab:ablation_full} reports all 17 benchmarks for the component ablation of \cref{tab:ablation}. Every block starts from the base model. The 4B block compares the MediX-R1 recipe (DAPO on the blended reward with KL regularisation), DAPO with the judge reward only, DAPO with the embedding gate, and the cold start alone with MedFrame-R1-4B, which combines the cold start with DAPO and the gate. The 8B block compares MediX-R1, GRPO, and the cold start alone with MedFrame-R1 without and with the length reward.

\begin{table*}[t]
\centering
\caption{\textbf{Ablation on all benchmarks.} Component ablation after 200 training steps, starting from the base model. Rows indented further start from an SFT cold start, bold and underlined values mark the best and second-best result, and $^{\dagger}$ denotes our reproduction of MediX-R1. Without a length reward, 28.4\% of the 8B rollouts loop (a phrase of eight words repeated three or more times) and 26.4\% exceed 1{,}000 words, against 1.2\% and 0.5\% with it.}
\small
\setlength{\tabcolsep}{5.5pt}
\renewcommand{\arraystretch}{1.3}
\setlength{\aboverulesep}{0pt}\setlength{\belowrulesep}{0pt}   
\resizebox{\textwidth}{!}{%
\begin{tabular}{l|cccccc|ccccccccccc|cccc}
\toprule
\multicolumn{1}{c|}{\rule{0pt}{2.2cm}} & \multicolumn{6}{c|}{\cellcolor{cVQA}} & \multicolumn{11}{c|}{\cellcolor{cQA}} & \multicolumn{4}{c}{\cellcolor{cAVG}} \\[-2.37cm]
\multicolumn{1}{c|}{\rule{0pt}{2.2cm}\raisebox{0.93cm}[0pt][0pt]{\large\textbf{Method}}}
& \multicolumn{1}{c}{\raisebox{0.93cm}[0pt][0pt]{\makebox[0pt][c]{\rotatebox[origin=c]{65}{\footnotesize\textbf{SLAKE}}}}}
& \multicolumn{1}{c}{\raisebox{0.93cm}[0pt][0pt]{\makebox[0pt][c]{\rotatebox[origin=c]{65}{\footnotesize\textbf{VQA-RAD}}}}}
& \multicolumn{1}{c}{\raisebox{0.93cm}[0pt][0pt]{\makebox[0pt][c]{\rotatebox[origin=c]{65}{\footnotesize\textbf{PathVQA}}}}}
& \multicolumn{1}{c}{\raisebox{0.93cm}[0pt][0pt]{\makebox[0pt][c]{\rotatebox[origin=c]{65}{\footnotesize\textbf{PMC-VQA}}}}}
& \multicolumn{1}{c}{\raisebox{0.93cm}[0pt][0pt]{\makebox[0pt][c]{\rotatebox[origin=c]{65}{\footnotesize\textbf{PMC-VQA-H}}}}}
& \multicolumn{1}{c|}{\raisebox{0.93cm}[0pt][0pt]{\makebox[0pt][c]{\rotatebox[origin=c]{65}{\footnotesize\textbf{MIMIC-Gen}}}}}
& \multicolumn{1}{c}{\raisebox{0.93cm}[0pt][0pt]{\makebox[0pt][c]{\rotatebox[origin=c]{65}{\footnotesize\textbf{MMLU-Clin}}}}}
& \multicolumn{1}{c}{\raisebox{0.93cm}[0pt][0pt]{\makebox[0pt][c]{\rotatebox[origin=c]{65}{\footnotesize\textbf{MMLU-Bio}}}}}
& \multicolumn{1}{c}{\raisebox{0.93cm}[0pt][0pt]{\makebox[0pt][c]{\rotatebox[origin=c]{65}{\footnotesize\textbf{MMLU-Med}}}}}
& \multicolumn{1}{c}{\raisebox{0.93cm}[0pt][0pt]{\makebox[0pt][c]{\rotatebox[origin=c]{65}{\footnotesize\textbf{MMLU-Gen}}}}}
& \multicolumn{1}{c}{\raisebox{0.93cm}[0pt][0pt]{\makebox[0pt][c]{\rotatebox[origin=c]{65}{\footnotesize\textbf{MMLU-Prof}}}}}
& \multicolumn{1}{c}{\raisebox{0.93cm}[0pt][0pt]{\makebox[0pt][c]{\rotatebox[origin=c]{65}{\footnotesize\textbf{MMLU-Anat}}}}}
& \multicolumn{1}{c}{\raisebox{0.93cm}[0pt][0pt]{\makebox[0pt][c]{\rotatebox[origin=c]{65}{\footnotesize\textbf{MedMCQA}}}}}
& \multicolumn{1}{c}{\raisebox{0.93cm}[0pt][0pt]{\makebox[0pt][c]{\rotatebox[origin=c]{65}{\footnotesize\textbf{MedQA}}}}}
& \multicolumn{1}{c}{\raisebox{0.93cm}[0pt][0pt]{\makebox[0pt][c]{\rotatebox[origin=c]{65}{\footnotesize\textbf{USMLE-SA}}}}}
& \multicolumn{1}{c}{\raisebox{0.93cm}[0pt][0pt]{\makebox[0pt][c]{\rotatebox[origin=c]{65}{\footnotesize\textbf{PubMedQA}}}}}
& \multicolumn{1}{c|}{\raisebox{0.93cm}[0pt][0pt]{\makebox[0pt][c]{\rotatebox[origin=c]{65}{\footnotesize\textbf{MIMIC-Sum}}}}}
& \multicolumn{1}{c}{\raisebox{0.93cm}[0pt][0pt]{\makebox[0pt][c]{\rotatebox[origin=c]{65}{\footnotesize\textbf{VQA}}}}}
& \multicolumn{1}{c}{\raisebox{0.93cm}[0pt][0pt]{\makebox[0pt][c]{\rotatebox[origin=c]{65}{\footnotesize\textbf{QA}}}}}
& \multicolumn{1}{c}{\raisebox{0.93cm}[0pt][0pt]{\makebox[0pt][c]{\rotatebox[origin=c]{65}{\footnotesize\textbf{Macro}}}}}
& \multicolumn{1}{c}{\raisebox{0.93cm}[0pt][0pt]{\makebox[0pt][c]{\rotatebox[origin=c]{65}{\footnotesize\textbf{Micro}}}}} \\
\midrule
\rowcolor{cSub}
\textbf{Qwen3-VL-4B} (zero-shot)
& 67.9 & 54.3 & 40.0 & 53.0 & 34.2 & 31.6 & \underline{78.9} & \underline{88.2} & \textbf{80.3} & \textbf{88.0} & \underline{78.7} & \underline{71.8} & 58.7 & 65.9 & 71.1 & 40.6 & 54.2 & 46.8 & 70.6 & 62.2 & 51.5 \\
\hspace{1.2em}MediX-R1$^{\dagger}$
& 73.7 & 58.3 & 44.6 & \underline{54.8} & 34.3 & 29.4 & 76.6 & \underline{88.2} & \underline{79.8} & \underline{87.0} & 77.9 & 70.4 & 58.6 & 67.2 & \textbf{73.2} & 44.4 & 52.1 & 49.2 & 70.5 & 63.0 & 53.8 \\
\hspace{1.2em}DAPO (judge only)
& 74.2 & \underline{58.5} & 48.3 & 54.2 & \underline{35.7} & \underline{33.0} & 77.7 & \underline{88.2} & 78.0 & \textbf{88.0} & 77.6 & 70.4 & 58.7 & 67.4 & 70.2 & 48.6 & \underline{54.6} & 50.6 & 70.9 & 63.7 & 55.3 \\
\hspace{1.2em}DAPO + gate
& \underline{74.3} & 57.4 & \underline{50.6} & \textbf{55.9} & 35.4 & \textbf{33.4} & 78.1 & 86.1 & 78.0 & 86.0 & 77.9 & \textbf{71.9} & \underline{59.0} & \underline{67.8} & \underline{72.0} & \textbf{50.0} & 54.4 & \underline{51.2} & \underline{71.0} & \underline{64.0} & \underline{56.4} \\
\hspace{1.2em}wCE + AESL SFT cold start
& 73.9 & 54.8 & 49.5 & 50.4 & 32.0 & \textbf{33.4} & \underline{78.9} & 87.5 & 79.2 & 85.0 & 74.3 & 69.6 & 56.4 & 61.7 & 61.9 & 48.8 & \textbf{56.0} & 49.0 & 69.0 & 62.0 & 53.9 \\
\midrule
\rowcolor{cOurs}
\hspace{2.4em}\textbf{MedFrame-R1-4B}
& \textbf{74.4} & \textbf{61.6} & \textbf{52.1} & 54.7 & \textbf{36.0} & 32.4 & \textbf{80.4} & \textbf{90.3} & 79.2 & \textbf{88.0} & \textbf{79.8} & 69.6 & \textbf{60.1} & \textbf{70.2} & 71.4 & \underline{49.8} & 50.8 & \textbf{51.9} & \textbf{71.8} & \textbf{64.8} & \textbf{57.4} \\
\midrule
\rowcolor{cSub}
\textbf{Qwen3-VL-8B} (zero-shot)
& 61.6 & 52.8 & 36.0 & 53.2 & 34.0 & \textbf{34.4} & 80.4 & \underline{92.4} & 80.3 & 87.0 & \underline{84.6} & 75.6 & 62.3 & \textbf{75.9} & 77.2 & 42.0 & \underline{53.0} & 45.3 & 73.7 & 63.7 & 51.4 \\
\hspace{1.2em}MediX-R1$^{\dagger}$
& 70.2 & 59.9 & 46.6 & \underline{53.8} & \underline{34.7} & 30.6 & 80.0 & \textbf{93.8} & 81.5 & \underline{89.0} & 83.5 & \underline{77.0} & \textbf{62.6} & \underline{75.3} & \textbf{79.4} & 43.8 & 52.6 & 49.3 & \underline{74.4} & \underline{65.5} & 56.0 \\
\hspace{1.2em}GRPO
& 69.3 & 56.1 & 48.6 & \underline{53.8} & 33.8 & 20.4 & 81.9 & \textbf{93.8} & \textbf{83.2} & \textbf{90.0} & \underline{84.6} & 74.1 & \underline{62.5} & \textbf{75.9} & \underline{78.8} & 45.2 & 51.8 & 47.0 & \textbf{74.7} & 64.9 & 56.5 \\
\hspace{1.2em}wCE + AESL SFT cold start
& 72.3 & 60.5 & \underline{52.1} & 50.6 & \underline{34.7} & 28.0 & \underline{82.6} & \underline{92.4} & \underline{82.7} & 88.0 & \underline{84.6} & \textbf{78.5} & 61.4 & 71.8 & 74.5 & 44.2 & \textbf{53.6} & 49.7 & 74.0 & 65.4 & \underline{57.2} \\
\hspace{2.4em}MedFrame-R1 (no length reward)
& \underline{74.3} & \underline{61.2} & 51.7 & 48.2 & 33.4 & \underline{33.6} & \textbf{84.2} & \textbf{93.8} & 82.1 & 88.0 & 81.6 & 71.9 & 61.7 & 72.9 & 76.0 & \underline{46.2} & 52.6 & \underline{50.4} & 73.7 & \underline{65.5} & 57.0 \\
\midrule
\rowcolor{cOurs}
\hspace{2.4em}\textbf{MedFrame-R1-8B} (+ length reward)
& \textbf{76.5} & \textbf{65.6} & \textbf{53.8} & \textbf{56.2} & \textbf{34.8} & 33.4 & \underline{82.6} & 91.7 & \underline{82.7} & 88.0 & \textbf{85.3} & 74.1 & 61.8 & 73.6 & 76.0 & \textbf{46.6} & 52.8 & \textbf{53.4} & 74.1 & \textbf{66.8} & \textbf{59.1} \\
\bottomrule
\end{tabular}}

\label{tab:ablation_full}
\end{table*}

\subsection{Gate Threshold}
\label{app:gate_full}

\Cref{tab:gate_full} reports the results across all 17 benchmarks for different embedding similarity thresholds and rule-based verification. In the rule-based approach, the embedding gate is replaced with a fixed check that penalises rollouts containing multiple candidate answers. This aims to prevent reward hacking by discouraging the model from listing several possible answers instead of making a clear prediction. Unlike the embedding gate, however, this approach relies on explicit answer patterns rather than measuring semantic similarity with the reference.
Rule-based verification achieves $63.6$ macro and $55.3$ micro, compared with $64.8$ and $57.4$ for the selected threshold of $\tau=0.80$. The embedding gate also achieves higher VQA and QA averages, suggesting that semantic similarity provides a more useful training signal than fixed answer rules. \Cref{fig:gate_threshold_area} further shows that higher thresholds reject an increasing proportion of correct rollouts, supporting our choice of $\tau=0.80$.

\begin{table*}[t]
\centering
\caption{\textbf{Gate threshold on all benchmarks.} Effect of the embedding gate threshold $\tau$ on Qwen3-VL-4B. Rule-based verification replaces the embedding gate with a rule that penalises rollouts listing several candidate answers, and the $\tau{=}0.80$ row is the MedFrame-R1 model of the main results. VQA and QA are the means over the 6 image-based and 11 text-based benchmarks, Macro is the mean over all 17 benchmarks, and Micro is the accuracy pooled over 19{,}200 samples. Bold and underlined values mark the best and second-best result.}
\small
\setlength{\tabcolsep}{5.5pt}
\renewcommand{\arraystretch}{1.3}
\setlength{\aboverulesep}{0pt}\setlength{\belowrulesep}{0pt} 

\resizebox{\textwidth}{!}{%
\begin{tabular}{l|cccccc|ccccccccccc|cccc}
\toprule
\multicolumn{1}{c|}{\rule{0pt}{2.2cm}} & \multicolumn{6}{c|}{\cellcolor{cVQA}} & \multicolumn{11}{c|}{\cellcolor{cQA}} & \multicolumn{4}{c}{\cellcolor{cAVG}} \\[-2.37cm]
\multicolumn{1}{c|}{\rule{0pt}{2.2cm}\raisebox{0.93cm}[0pt][0pt]{\large\textbf{Method}}}
& \multicolumn{1}{c}{\raisebox{0.93cm}[0pt][0pt]{\makebox[0pt][c]{\rotatebox[origin=c]{65}{\footnotesize\textbf{SLAKE}}}}}
& \multicolumn{1}{c}{\raisebox{0.93cm}[0pt][0pt]{\makebox[0pt][c]{\rotatebox[origin=c]{65}{\footnotesize\textbf{VQA-RAD}}}}}
& \multicolumn{1}{c}{\raisebox{0.93cm}[0pt][0pt]{\makebox[0pt][c]{\rotatebox[origin=c]{65}{\footnotesize\textbf{PathVQA}}}}}
& \multicolumn{1}{c}{\raisebox{0.93cm}[0pt][0pt]{\makebox[0pt][c]{\rotatebox[origin=c]{65}{\footnotesize\textbf{PMC-VQA}}}}}
& \multicolumn{1}{c}{\raisebox{0.93cm}[0pt][0pt]{\makebox[0pt][c]{\rotatebox[origin=c]{65}{\footnotesize\textbf{PMC-VQA-H}}}}}
& \multicolumn{1}{c|}{\raisebox{0.93cm}[0pt][0pt]{\makebox[0pt][c]{\rotatebox[origin=c]{65}{\footnotesize\textbf{MIMIC-Gen}}}}}
& \multicolumn{1}{c}{\raisebox{0.93cm}[0pt][0pt]{\makebox[0pt][c]{\rotatebox[origin=c]{65}{\footnotesize\textbf{MMLU-Clin}}}}}
& \multicolumn{1}{c}{\raisebox{0.93cm}[0pt][0pt]{\makebox[0pt][c]{\rotatebox[origin=c]{65}{\footnotesize\textbf{MMLU-Bio}}}}}
& \multicolumn{1}{c}{\raisebox{0.93cm}[0pt][0pt]{\makebox[0pt][c]{\rotatebox[origin=c]{65}{\footnotesize\textbf{MMLU-Med}}}}}
& \multicolumn{1}{c}{\raisebox{0.93cm}[0pt][0pt]{\makebox[0pt][c]{\rotatebox[origin=c]{65}{\footnotesize\textbf{MMLU-Gen}}}}}
& \multicolumn{1}{c}{\raisebox{0.93cm}[0pt][0pt]{\makebox[0pt][c]{\rotatebox[origin=c]{65}{\footnotesize\textbf{MMLU-Prof}}}}}
& \multicolumn{1}{c}{\raisebox{0.93cm}[0pt][0pt]{\makebox[0pt][c]{\rotatebox[origin=c]{65}{\footnotesize\textbf{MMLU-Anat}}}}}
& \multicolumn{1}{c}{\raisebox{0.93cm}[0pt][0pt]{\makebox[0pt][c]{\rotatebox[origin=c]{65}{\footnotesize\textbf{MedMCQA}}}}}
& \multicolumn{1}{c}{\raisebox{0.93cm}[0pt][0pt]{\makebox[0pt][c]{\rotatebox[origin=c]{65}{\footnotesize\textbf{MedQA}}}}}
& \multicolumn{1}{c}{\raisebox{0.93cm}[0pt][0pt]{\makebox[0pt][c]{\rotatebox[origin=c]{65}{\footnotesize\textbf{USMLE-SA}}}}}
& \multicolumn{1}{c}{\raisebox{0.93cm}[0pt][0pt]{\makebox[0pt][c]{\rotatebox[origin=c]{65}{\footnotesize\textbf{PubMedQA}}}}}
& \multicolumn{1}{c|}{\raisebox{0.93cm}[0pt][0pt]{\makebox[0pt][c]{\rotatebox[origin=c]{65}{\footnotesize\textbf{MIMIC-Sum}}}}}
& \multicolumn{1}{c}{\raisebox{0.93cm}[0pt][0pt]{\makebox[0pt][c]{\rotatebox[origin=c]{65}{\footnotesize\textbf{VQA}}}}}
& \multicolumn{1}{c}{\raisebox{0.93cm}[0pt][0pt]{\makebox[0pt][c]{\rotatebox[origin=c]{65}{\footnotesize\textbf{QA}}}}}
& \multicolumn{1}{c}{\raisebox{0.93cm}[0pt][0pt]{\makebox[0pt][c]{\rotatebox[origin=c]{65}{\footnotesize\textbf{Macro}}}}}
& \multicolumn{1}{c}{\raisebox{0.93cm}[0pt][0pt]{\makebox[0pt][c]{\rotatebox[origin=c]{65}{\footnotesize\textbf{Micro}}}}} \\
\midrule
Rule-based verification
& 74.5 & 58.0 & 48.1 & 54.7 & 35.9 & 32.7 & 77.3 & 87.5 & 77.2 & \underline{87.0} & \underline{77.6} & \underline{72.6} & 59.0 & \underline{67.5} & 70.8 & 48.9 & 51.1 & 50.7 & \underline{70.6} & \underline{63.6} & 55.3 \\
$\tau{=}0.70$ 
& 73.9 & 55.4 & 46.5 & \textbf{55.1} & 35.4 & \textbf{34.2} & 77.0 & 86.8 & 76.9 & \underline{87.0} & \underline{77.6} & 71.9 & \underline{59.2} & 66.9 & \textbf{72.9} & 48.8 & 51.0 & 50.1 & 70.5 & 63.3 & 54.7 \\
$\tau{=}0.75$
& \underline{74.6} & \underline{58}.1 & 49.4 & 54.7 & 35.5 & \underline{32.8} & 75.5 & \underline{88.9} & \underline{78.6} & \underline{87.0} & \underline{77.6} & \textbf{74.8} & 58.3 & 66.1 & 68.9 & 46.8 & \underline{51.6} & 50.8 & 70.4 & 63.5 & 55.5 \\
$\tau{=}0.78$
& \textbf{75.1} & 56.8 & \underline{49.8} & \underline{55.0} & \textbf{37.0} & 32.6 & \underline{77.7} & 84.7 & 75.7 & \underline{87.0} & \underline{77.6} & 71.9 & 58.2 & 65.6 & 71.1 & \underline{49.2} & \textbf{52.8} & \underline{51.0} & 70.1 & 63.4 & \underline{55.7} \\
\midrule
\rowcolor{cOurs}
$\tau{=}0.80$ \textbf{(MedFrame-R1-4B)}
& 74.4 & \textbf{61.6} & \textbf{52.1} & 54.7 & \underline{36.0} & 32.4 & \textbf{80.4} & \textbf{90.3} & \textbf{79.2} & \textbf{88.0} & \textbf{79.8} & 69.6 & \textbf{60.1} & \textbf{70.2} & \underline{71.4} & \textbf{49.8} & 50.8 & \textbf{51.9} & \textbf{71.8} & \textbf{64.8} & \textbf{57.4} \\
\bottomrule
\end{tabular}}

\label{tab:gate_full}
\end{table*}

\section{Trainable Data and Modality Coverage}
\label{sec:app-trainable-data}

Dynamic sampling selects samples based on the correctness of their generated rollouts, but not all samples contribute equally to training. We therefore examine how many unique samples provide useful training signals, how this changes throughout RL, and which modalities remain difficult to learn. This analysis helps explain the need for the supervised cold start in \cref{sec:sft} and the use of a trainable sample pool for Qwen3-VL models, as investigated in \cref{tab:trainable_4b}.

\subsection{Cumulative Coverage Across Attempts}
\label{sec:app-repeated-probing}

The RL run completes 200 optimisation steps and 17 attempts, where each attempt is one pass through the training dataset. Each attempt revisits the samples, but only groups with mixed correctness are selected for training. Therefore, not every sample contributes to every update.

\begin{figure*}[tp]
    \centering
    \sbox{\suppfigbox}{\includegraphics[width=\textwidth]{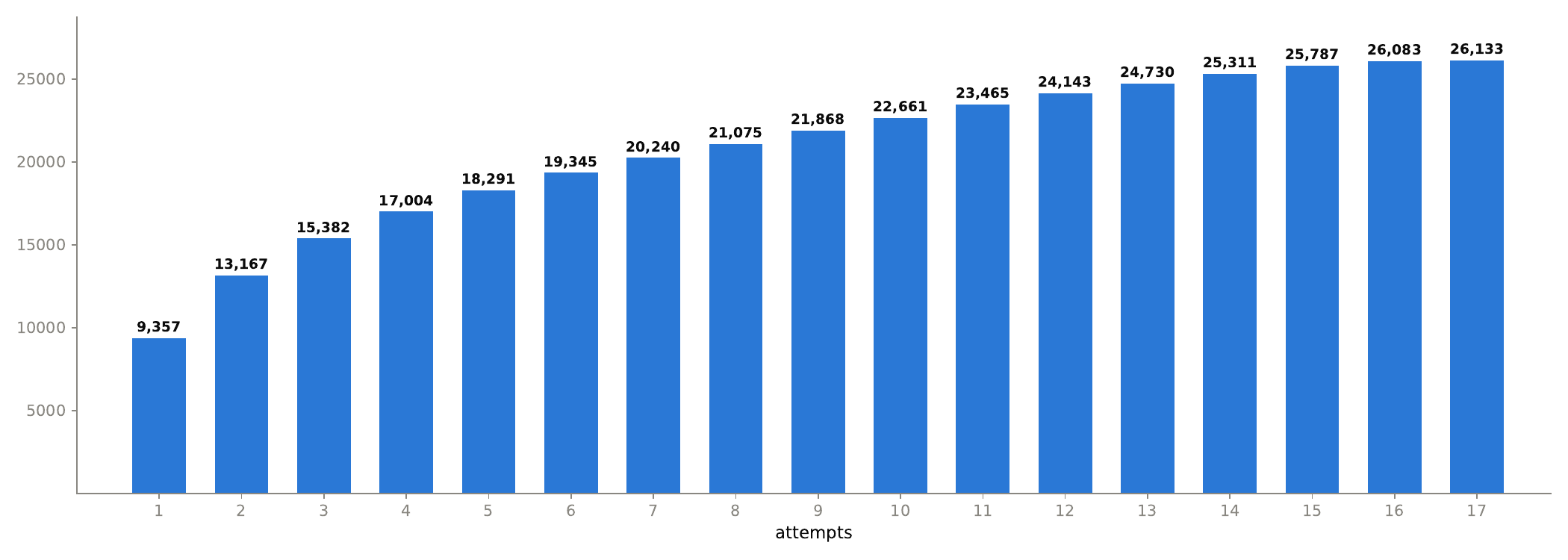}}%
    \usebox{\suppfigbox}
\captionsetup{width=\wd\suppfigbox}
    \caption{\textbf{Cumulative coverage of RL samples.} Each bar counts the distinct samples that contributed a trainable group to MedFrame-R1 by that attempt, rising from 9{,}357 to 26{,}133. In its first attempt, GRPO finds about 6.5K such samples.}
    \label{fig:app-trainable-attempts}
\end{figure*}

\Cref{fig:app-trainable-attempts} shows how the number of unique samples used for training increases across attempts. How many samples give a correctness signal depends on the training recipe, as summarised in \cref{tab:app-first-attempt}. With GRPO, about 6{,}500 samples produce mixed judge outcomes in the first attempt, and 11{,}439 over the 200 steps. DAPO filtering on the blended reward of MediX-R1 keeps most sampled groups, including many in which every rollout receives the same judge decision, so each attempt takes many more steps. Its first attempt lasts 133 of the 200 steps and finds 11{,}295 samples with mixed LLM judge scores, and the whole run sees each sample at most three times and reaches 18{,}611 such samples. MedFrame-R1 filters on the raw judge decision and finishes its first attempt within 32 steps with 9{,}357 samples. It then keeps revisiting the data, rising to 13{,}167 samples after the second attempt and 26{,}133 after all 17 attempts. MedFrame-R1 therefore reaches the highest total, while DAPO on the blended reward has a higher first-attempt count because the two methods pass through the data at different speeds. Overall, approximately $50.9\%$ of the 51{,}335 samples contribute to training at least once.
Each bar represents the cumulative number of unique samples, so the values should not be added together. The number of newly selected samples decreases over time, from 3{,}810 in the second attempt to only 50 in the final attempt. This shows that repeated attempts gradually increase training coverage, but the benefit becomes smaller over time. \Cref{fig:app-trainable-transitions} further examines how sample selection changes between attempts.

\begin{table}[t]
\centering
\sbox{\supptabbox}{%
\footnotesize
\renewcommand{\arraystretch}{1.05}
\setlength{\tabcolsep}{8pt}
\begin{tabular}{lrr}
\toprule
\textbf{Method} & \textbf{First attempt} & \textbf{Total (200 steps)} \\
\midrule
GRPO & $\sim$6{,}500 & 11{,}439 \\
DAPO (MediX) & 11{,}295 & 18{,}611 \\
\rowcolor{cOurs}
MedFrame-R1 & 9{,}357 & \textbf{26{,}133} \\
\bottomrule
\end{tabular}}
\captionsetup{width=\wd\supptabbox}
\caption{\textbf{Samples with mixed judge outcomes.} Distinct training samples whose rollout group receives mixed LLM judge decisions, in the first attempt and over the 200 RL steps. DAPO (MediX) filters on the blended reward, and MedFrame-R1 filters on the raw judge decision.}
\usebox{\supptabbox}
\label{tab:app-first-attempt}
\end{table}

\subsection{How Trainability Changes Between Attempts}
\label{sec:app-trainable-transitions}

\begin{figure*}[tp]
    \centering
    \sbox{\suppfigbox}{\includegraphics[width=\textwidth]{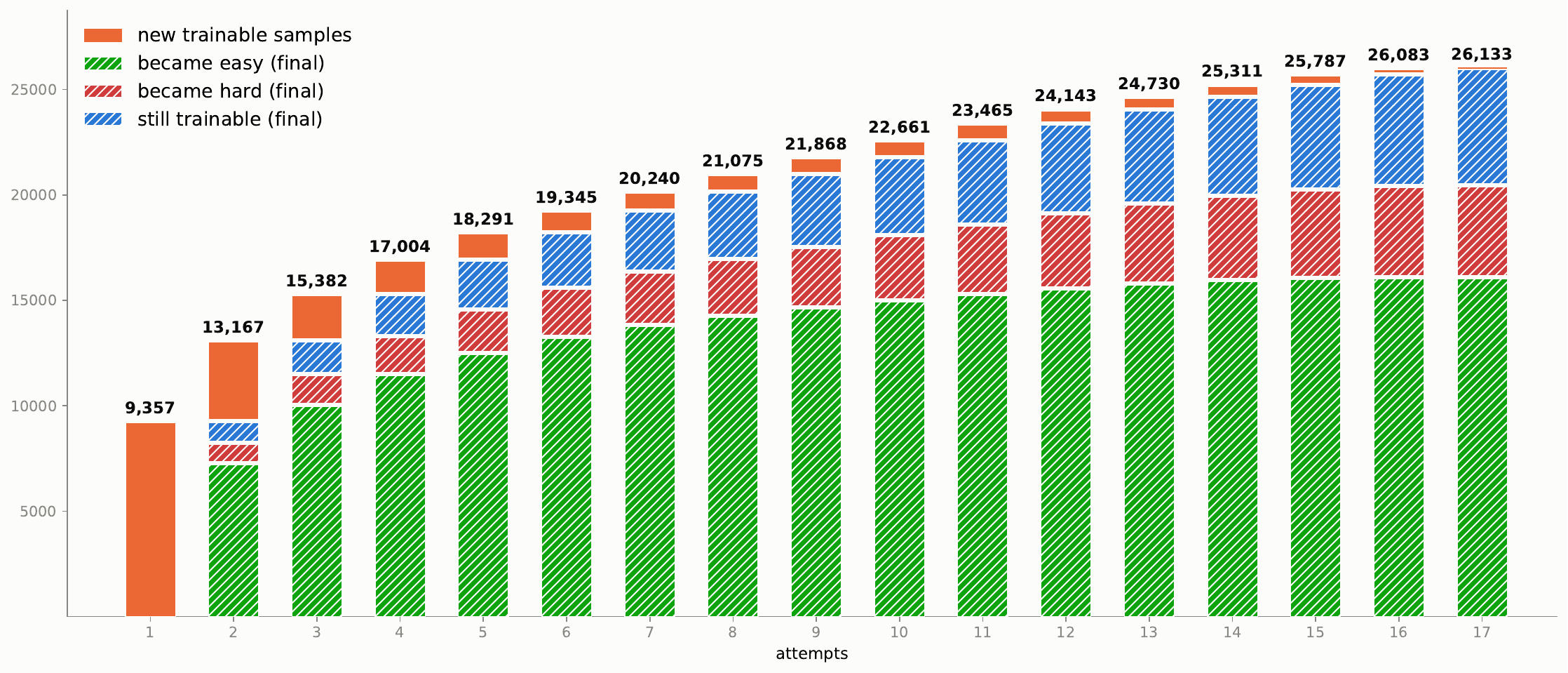}}%
    \usebox{\suppfigbox}
\captionsetup{width=\wd\suppfigbox}
    \caption{\textbf{Trainability across attempts.} Orange shows samples that become trainable for the first time. Green, red, and blue show earlier samples that are now easy, hard, or still trainable.}
    \label{fig:app-trainable-transitions}
\end{figure*}

Cumulative coverage does not show how many samples remain useful during training, since previously selected samples stay in the count even if their later rollouts are all correct or incorrect. \Cref{fig:app-trainable-transitions} shows how this changes across attempts. Orange represents newly trainable samples, green shows previously selected samples whose rollouts are now all correct, red shows those whose rollouts are all incorrect, and blue represents samples that remain trainable. Together, orange and blue show the trainable samples at each attempt, while the full bar shows cumulative coverage. These changes reflect both improvements in the policy and variation in generated rollouts.

A sample is useful for dynamic sampling only when its rollouts have different judge outcomes. However, the judge may incorrectly reward long lists of possible answers, as discussed in \cref{sec:app-reward-hacking}. The embedding gate addresses this issue in reward computation, while dynamic sampling continues to use the raw judge decisions. Therefore, increased coverage results from repeated sampling and changes in the policy, rather than the embedding gate alone. The effect of the gate on reward hacking is further illustrated in \cref{sec:app-reward-hacking}.

\subsection{Modality Coverage and Remaining Difficult Samples}
\label{sec:app-modality-coverage}

\begin{figure*}[tp]
    \centering
    \sbox{\suppfigbox}{\includegraphics[width=\textwidth]{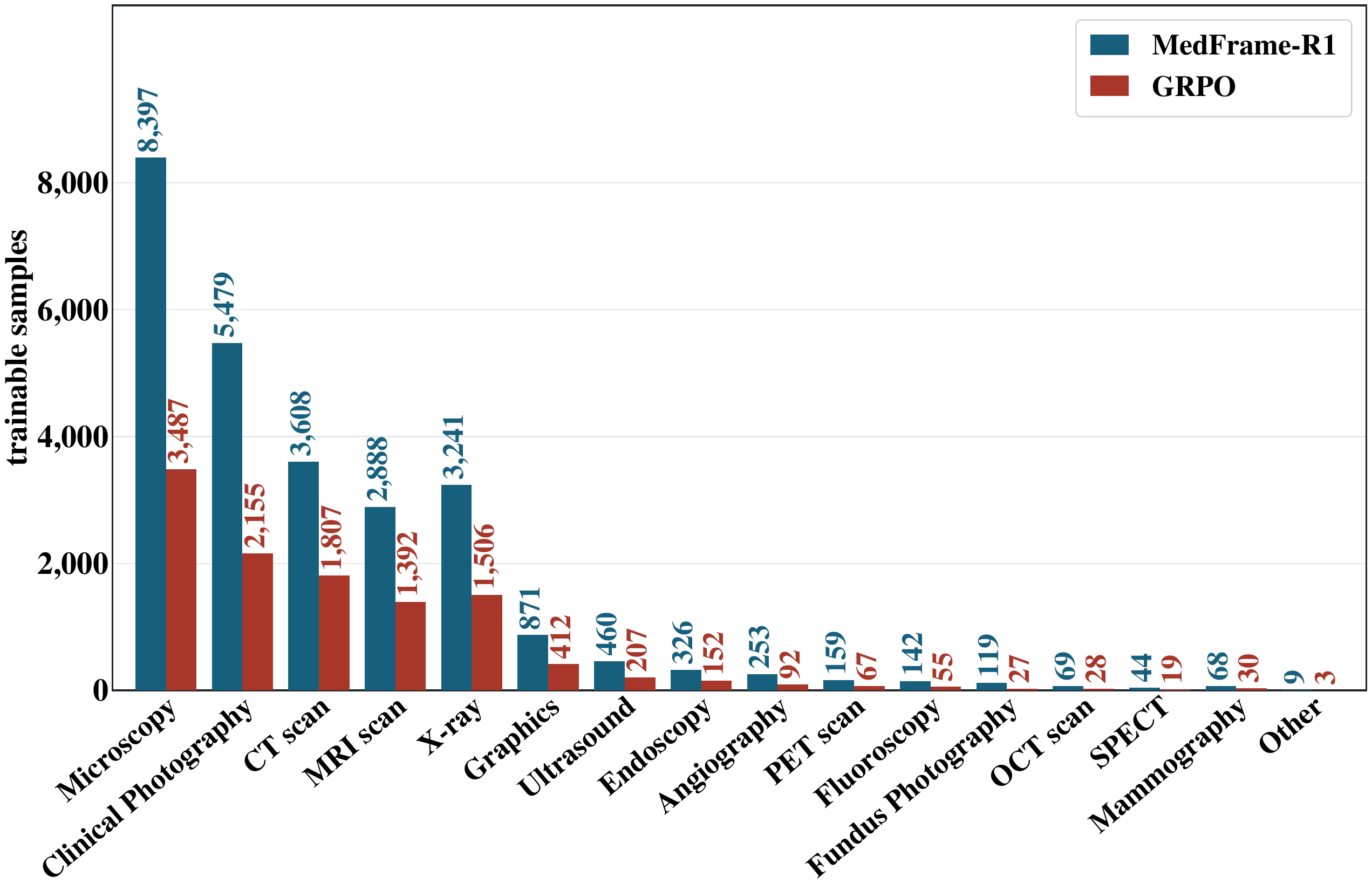}}%
    \usebox{\suppfigbox}
\captionsetup{width=\wd\suppfigbox}
    \caption{\textbf{Trainable samples by modality.} MedFrame-R1 has more trainable samples than plain GRPO in every displayed modality.}
    \label{fig:app-trainable-modalities}
    \label{fig:app-trainable-modality-comparison}
\end{figure*}

\Cref{fig:app-trainable-modalities} compares the number of trainable samples per modality for MedFrame-R1 and plain GRPO. The MedFrame-R1 bars are the 26{,}133 samples of the trainable pool. MedFrame-R1 has more trainable samples in every displayed modality, for example 8{,}397 against 3{,}487 for microscopy and 5{,}479 against 2{,}155 for clinical photography, and also in rare modalities such as OCT with 69 against 28 and SPECT with 44 against 19. Microscopy, clinical photography, CT, MRI, and X-ray supply most of the trainable samples.

\subsection{Effect of the Training Pool}
\label{sec:app-pool-results}

Limited representation and persistent failure are different obstacles. Rare modalities contain few examples, while difficult samples may be numerous but repeatedly fail to produce an accepted rollout, and dynamic sampling cannot supply a correct target for either. Restricting RL to the trainable pool is nevertheless efficient, since it reduces the time for 200 steps from about 56 to 30 hours, because fewer samples must be sampled repeatedly before useful groups appear. \Cref{tab:trainable_4b_full} gives the complete benchmark results for training on the full 51K samples and on the 26K trainable pool. GRPO reaches a macro score of 63.2 on the pool against 62.7 on the full set, and MedFrame-R1 reaches 64.8 on the pool against 63.8 on the full set. The pool can exclude difficult or rare samples that the model cannot yet answer, which is why the supervised cold start covers the full corpus and supplies targets for them.

\begin{table*}[tp]
\centering
\caption{\textbf{Full and trainable pool.} Qwen3-VL-4B after 200 RL steps, trained on all 51K samples or on the 26K trainable pool. Bold and underlined values mark the best and second-best result.}
\small
\setlength{\tabcolsep}{5.5pt}
\renewcommand{\arraystretch}{1.3}
\setlength{\aboverulesep}{0pt}\setlength{\belowrulesep}{0pt}   
\resizebox{\textwidth}{!}{%
\begin{tabular}{l|cccccc|ccccccccccc|cccc}
\toprule
\multicolumn{1}{c|}{\rule{0pt}{2.2cm}} & \multicolumn{6}{c|}{\cellcolor{cVQA}} & \multicolumn{11}{c|}{\cellcolor{cQA}} & \multicolumn{4}{c}{\cellcolor{cAVG}} \\[-2.37cm]
\multicolumn{1}{c|}{\rule{0pt}{2.2cm}\raisebox{0.93cm}[0pt][0pt]{\large\textbf{Method}}}
& \multicolumn{1}{c}{\raisebox{0.93cm}[0pt][0pt]{\makebox[0pt][c]{\rotatebox[origin=c]{65}{\footnotesize\textbf{SLAKE}}}}}
& \multicolumn{1}{c}{\raisebox{0.93cm}[0pt][0pt]{\makebox[0pt][c]{\rotatebox[origin=c]{65}{\footnotesize\textbf{VQA-RAD}}}}}
& \multicolumn{1}{c}{\raisebox{0.93cm}[0pt][0pt]{\makebox[0pt][c]{\rotatebox[origin=c]{65}{\footnotesize\textbf{PathVQA}}}}}
& \multicolumn{1}{c}{\raisebox{0.93cm}[0pt][0pt]{\makebox[0pt][c]{\rotatebox[origin=c]{65}{\footnotesize\textbf{PMC-VQA}}}}}
& \multicolumn{1}{c}{\raisebox{0.93cm}[0pt][0pt]{\makebox[0pt][c]{\rotatebox[origin=c]{65}{\footnotesize\textbf{PMC-VQA-H}}}}}
& \multicolumn{1}{c|}{\raisebox{0.93cm}[0pt][0pt]{\makebox[0pt][c]{\rotatebox[origin=c]{65}{\footnotesize\textbf{MIMIC-Gen}}}}}
& \multicolumn{1}{c}{\raisebox{0.93cm}[0pt][0pt]{\makebox[0pt][c]{\rotatebox[origin=c]{65}{\footnotesize\textbf{MMLU-Clin}}}}}
& \multicolumn{1}{c}{\raisebox{0.93cm}[0pt][0pt]{\makebox[0pt][c]{\rotatebox[origin=c]{65}{\footnotesize\textbf{MMLU-Bio}}}}}
& \multicolumn{1}{c}{\raisebox{0.93cm}[0pt][0pt]{\makebox[0pt][c]{\rotatebox[origin=c]{65}{\footnotesize\textbf{MMLU-Med}}}}}
& \multicolumn{1}{c}{\raisebox{0.93cm}[0pt][0pt]{\makebox[0pt][c]{\rotatebox[origin=c]{65}{\footnotesize\textbf{MMLU-Gen}}}}}
& \multicolumn{1}{c}{\raisebox{0.93cm}[0pt][0pt]{\makebox[0pt][c]{\rotatebox[origin=c]{65}{\footnotesize\textbf{MMLU-Prof}}}}}
& \multicolumn{1}{c}{\raisebox{0.93cm}[0pt][0pt]{\makebox[0pt][c]{\rotatebox[origin=c]{65}{\footnotesize\textbf{MMLU-Anat}}}}}
& \multicolumn{1}{c}{\raisebox{0.93cm}[0pt][0pt]{\makebox[0pt][c]{\rotatebox[origin=c]{65}{\footnotesize\textbf{MedMCQA}}}}}
& \multicolumn{1}{c}{\raisebox{0.93cm}[0pt][0pt]{\makebox[0pt][c]{\rotatebox[origin=c]{65}{\footnotesize\textbf{MedQA}}}}}
& \multicolumn{1}{c}{\raisebox{0.93cm}[0pt][0pt]{\makebox[0pt][c]{\rotatebox[origin=c]{65}{\footnotesize\textbf{USMLE-SA}}}}}
& \multicolumn{1}{c}{\raisebox{0.93cm}[0pt][0pt]{\makebox[0pt][c]{\rotatebox[origin=c]{65}{\footnotesize\textbf{PubMedQA}}}}}
& \multicolumn{1}{c|}{\raisebox{0.93cm}[0pt][0pt]{\makebox[0pt][c]{\rotatebox[origin=c]{65}{\footnotesize\textbf{MIMIC-Sum}}}}}
& \multicolumn{1}{c}{\raisebox{0.93cm}[0pt][0pt]{\makebox[0pt][c]{\rotatebox[origin=c]{65}{\footnotesize\textbf{VQA}}}}}
& \multicolumn{1}{c}{\raisebox{0.93cm}[0pt][0pt]{\makebox[0pt][c]{\rotatebox[origin=c]{65}{\footnotesize\textbf{QA}}}}}
& \multicolumn{1}{c}{\raisebox{0.93cm}[0pt][0pt]{\makebox[0pt][c]{\rotatebox[origin=c]{65}{\footnotesize\textbf{Macro}}}}}
& \multicolumn{1}{c}{\raisebox{0.93cm}[0pt][0pt]{\makebox[0pt][c]{\rotatebox[origin=c]{65}{\footnotesize\textbf{Micro}}}}} \\
\midrule
GRPO (Full 51K) & 72.7 & \underline{58.5} & 45.2 & \textbf{55.2} & \underline{34.5} & 28.4 & 77.0 & 86.8 & \textbf{79.8} & \underline{86.0} & \underline{78.7} & \underline{68.9} & 58.5 & 65.2 & 71.4 & 47.4 & \underline{51.2} & 49.1 & 70.1 & 62.7 & 53.9 \\
GRPO (Trainable 26K) & 72.2 & 56.5 & 44.8 & \underline{55.0} & 33.2 & \underline{31.6} & 78.1 & 87.5 & \textbf{79.8} & \underline{86.0} & \textbf{79.8} & \textbf{69.6} & 58.2 & 66.8 & \textbf{73.5} & \textbf{50.4} & \underline{51.2} & 48.9 & 71.0 & 63.2 & 53.8 \\
\textbf{MedFrame-R1-4B} (Full 51K) & \underline{73.3} & 56.3 & \underline{51.1} & 54.6 & 34.2 & 28.8 & \underline{79.2} & \textbf{91.7} & \underline{79.2} & 85.0 & \textbf{79.8} & \underline{68.9} & \underline{60.0} & \underline{68.6} & \underline{73.2} & 47.0 & \textbf{52.8} & \underline{49.7} & \underline{71.4} & \underline{63.8} & \underline{56.5} \\
\rowcolor{cOurs}
\textbf{MedFrame-R1-4B} (Trainable 26K) & \textbf{74.4} & \textbf{61.6} & \textbf{52.1} & 54.7 & \textbf{36.0} & \textbf{32.4} & \textbf{80.4} & \underline{90.3} & \underline{79.2} & \textbf{88.0} & \textbf{79.8} & \textbf{69.6} & \textbf{60.1} & \textbf{70.2} & 71.4 & \underline{49.8} & 50.8 & \textbf{51.9} & \textbf{71.8} & \textbf{64.8} & \textbf{57.4} \\
\bottomrule
\end{tabular}}

\label{tab:trainable_4b_full}
\end{table*}

\section{KL Regularisation}
\label{sec:app-kl-ablation}

\Cref{tab:rl_recipes_8b} provides the full results of the KL regularisation ablation in \cref{tab:kl_ablation}, including GRPO. All experiments use Qwen3-VL-8B trained for 200 steps. Without KL regularisation, MedFrame-R1 achieves a macro score of $66.8$ and a micro score of $59.1$, compared with $66.1$ and $58.5$ with KL. The improvement is larger on VQA benchmarks, while both variants achieve similar QA performance. Since KL regularisation limits how far the policy can move from the reference model, removing it allows more flexibility during training. Removing KL improves macro by 0.7 and micro by 0.6 points, mainly on VQA ($53.4$ against $52.0$). \Cref{sec:app-medmo} compares MedFrame-R1 with MedMO, which also applies RL to the Qwen3-VL backbone.

\begin{table*}[htp]
\centering
\caption{\textbf{RL recipes on Qwen3-VL-8B.} Results after 200 RL steps, in \%. GRPO, DAPO, and MediX-R1 start from the base model, MedFrame-R1 starts from the SFT cold start, and MediX-R1 is DAPO with KL regularisation. Bold and underlined values mark the best and second-best RL result.}
\small
\setlength{\tabcolsep}{5.5pt}
\renewcommand{\arraystretch}{1.3}
\setlength{\aboverulesep}{0pt}\setlength{\belowrulesep}{0pt}
\resizebox{\textwidth}{!}{%
\begin{tabular}{l|cccccc|ccccccccccc|cccc}
\toprule
\multicolumn{1}{c|}{\rule{0pt}{2.2cm}} & \multicolumn{6}{c|}{\cellcolor{cVQA}} & \multicolumn{11}{c|}{\cellcolor{cQA}} & \multicolumn{4}{c}{\cellcolor{cAVG}} \\[-2.37cm]
\multicolumn{1}{c|}{\rule{0pt}{2.2cm}\raisebox{0.93cm}[0pt][0pt]{\large\textbf{Method}}}
& \multicolumn{1}{c}{\raisebox{0.93cm}[0pt][0pt]{\makebox[0pt][c]{\rotatebox[origin=c]{65}{\footnotesize\textbf{SLAKE}}}}}
& \multicolumn{1}{c}{\raisebox{0.93cm}[0pt][0pt]{\makebox[0pt][c]{\rotatebox[origin=c]{65}{\footnotesize\textbf{VQA-RAD}}}}}
& \multicolumn{1}{c}{\raisebox{0.93cm}[0pt][0pt]{\makebox[0pt][c]{\rotatebox[origin=c]{65}{\footnotesize\textbf{PathVQA}}}}}
& \multicolumn{1}{c}{\raisebox{0.93cm}[0pt][0pt]{\makebox[0pt][c]{\rotatebox[origin=c]{65}{\footnotesize\textbf{PMC-VQA}}}}}
& \multicolumn{1}{c}{\raisebox{0.93cm}[0pt][0pt]{\makebox[0pt][c]{\rotatebox[origin=c]{65}{\footnotesize\textbf{PMC-VQA-H}}}}}
& \multicolumn{1}{c|}{\raisebox{0.93cm}[0pt][0pt]{\makebox[0pt][c]{\rotatebox[origin=c]{65}{\footnotesize\textbf{MIMIC-Gen}}}}}
& \multicolumn{1}{c}{\raisebox{0.93cm}[0pt][0pt]{\makebox[0pt][c]{\rotatebox[origin=c]{65}{\footnotesize\textbf{MMLU-Clin}}}}}
& \multicolumn{1}{c}{\raisebox{0.93cm}[0pt][0pt]{\makebox[0pt][c]{\rotatebox[origin=c]{65}{\footnotesize\textbf{MMLU-Bio}}}}}
& \multicolumn{1}{c}{\raisebox{0.93cm}[0pt][0pt]{\makebox[0pt][c]{\rotatebox[origin=c]{65}{\footnotesize\textbf{MMLU-Med}}}}}
& \multicolumn{1}{c}{\raisebox{0.93cm}[0pt][0pt]{\makebox[0pt][c]{\rotatebox[origin=c]{65}{\footnotesize\textbf{MMLU-Gen}}}}}
& \multicolumn{1}{c}{\raisebox{0.93cm}[0pt][0pt]{\makebox[0pt][c]{\rotatebox[origin=c]{65}{\footnotesize\textbf{MMLU-Prof}}}}}
& \multicolumn{1}{c}{\raisebox{0.93cm}[0pt][0pt]{\makebox[0pt][c]{\rotatebox[origin=c]{65}{\footnotesize\textbf{MMLU-Anat}}}}}
& \multicolumn{1}{c}{\raisebox{0.93cm}[0pt][0pt]{\makebox[0pt][c]{\rotatebox[origin=c]{65}{\footnotesize\textbf{MedMCQA}}}}}
& \multicolumn{1}{c}{\raisebox{0.93cm}[0pt][0pt]{\makebox[0pt][c]{\rotatebox[origin=c]{65}{\footnotesize\textbf{MedQA}}}}}
& \multicolumn{1}{c}{\raisebox{0.93cm}[0pt][0pt]{\makebox[0pt][c]{\rotatebox[origin=c]{65}{\footnotesize\textbf{USMLE-SA}}}}}
& \multicolumn{1}{c}{\raisebox{0.93cm}[0pt][0pt]{\makebox[0pt][c]{\rotatebox[origin=c]{65}{\footnotesize\textbf{PubMedQA}}}}}
& \multicolumn{1}{c|}{\raisebox{0.93cm}[0pt][0pt]{\makebox[0pt][c]{\rotatebox[origin=c]{65}{\footnotesize\textbf{MIMIC-Sum}}}}}
& \multicolumn{1}{c}{\raisebox{0.93cm}[0pt][0pt]{\makebox[0pt][c]{\rotatebox[origin=c]{65}{\footnotesize\textbf{VQA}}}}}
& \multicolumn{1}{c}{\raisebox{0.93cm}[0pt][0pt]{\makebox[0pt][c]{\rotatebox[origin=c]{65}{\footnotesize\textbf{QA}}}}}
& \multicolumn{1}{c}{\raisebox{0.93cm}[0pt][0pt]{\makebox[0pt][c]{\rotatebox[origin=c]{65}{\footnotesize\textbf{Macro}}}}}
& \multicolumn{1}{c}{\raisebox{0.93cm}[0pt][0pt]{\makebox[0pt][c]{\rotatebox[origin=c]{65}{\footnotesize\textbf{Micro}}}}} \\
\midrule
Qwen3-VL-8B (zero-shot) & 61.6 & 52.8 & 36.0 & 53.2 & 34.0 & 34.4 & 80.4 & 92.4 & 80.3 & 87.0 & 84.6 & 75.6 & 62.3 & 75.9 & 77.2 & 42.0 & 53.0 & 45.3 & 73.7 & 63.7 & 51.4 \\
\midrule
GRPO & 69.3 & 56.1 & 48.6 & 53.8 & 33.8 & 20.4 & 81.9 & \textbf{93.8} & \underline{83.2} & \textbf{90.0} & \underline{84.6} & 74.1 & \underline{62.5} & \underline{75.9} & 78.8 & 45.2 & 51.8 & 47.0 & \textbf{74.7} & 64.9 & 56.5 \\
DAPO (No KL) & 71.9 & \underline{63.6} & 47.8 & \underline{54.0} & \textbf{35.2} & \textbf{34.4} & 81.5 & \underline{93.1} & \textbf{84.4} & 87.0 & 81.2 & 74.1 & 62.3 & \textbf{76.0} & \textbf{80.0} & 42.2 & 50.0 & 51.2 & 73.8 & 65.8 & 56.6 \\
MediX-R1 (DAPO + KL) & 70.2 & 59.9 & 46.6 & 53.8 & 34.7 & 30.6 & 80.0 & \textbf{93.8} & 81.5 & \underline{89.0} & 83.5 & \textbf{77.0} & \textbf{62.6} & 75.3 & \underline{79.4} & 43.8 & \underline{52.6} & 49.3 & \underline{74.4} & 65.5 & 56.0 \\
MedFrame-R1 (+ KL) & \textbf{78.1} & 61.4 & \underline{53.6} & 53.6 & \underline{35.0} & 30.0 & \textbf{83.4} & 91.7 & \underline{83.2} & \underline{89.0} & 81.2 & \underline{74.8} & 61.4 & 72.4 & 75.1 & \textbf{47.8} & 52.0 & \underline{52.0} & 73.8 & \underline{66.1} & \underline{58.5} \\
\rowcolor{cOurs}
\textbf{MedFrame-R1 (No KL)} & \underline{76.5} & \textbf{65.6} & \textbf{53.8} & \textbf{56.2} & 34.8 & \underline{33.4} & \underline{82.6} & 91.7 & 82.7 & 88.0 & \textbf{85.3} & 74.1 & 61.8 & 73.6 & 76.0 & \underline{46.6} & \textbf{52.8} & \textbf{53.4} & 74.1 & \textbf{66.8} & \textbf{59.1} \\
\bottomrule
\end{tabular}}

\label{tab:rl_recipes_8b}
\end{table*}

\section{Failure Modes of RL-Trained Medical Models}
\label{sec:app-failure-examples}

\begin{figure*}[tp]
    \centering
    \sbox{\suppfigbox}{\includegraphics[width=\textwidth]{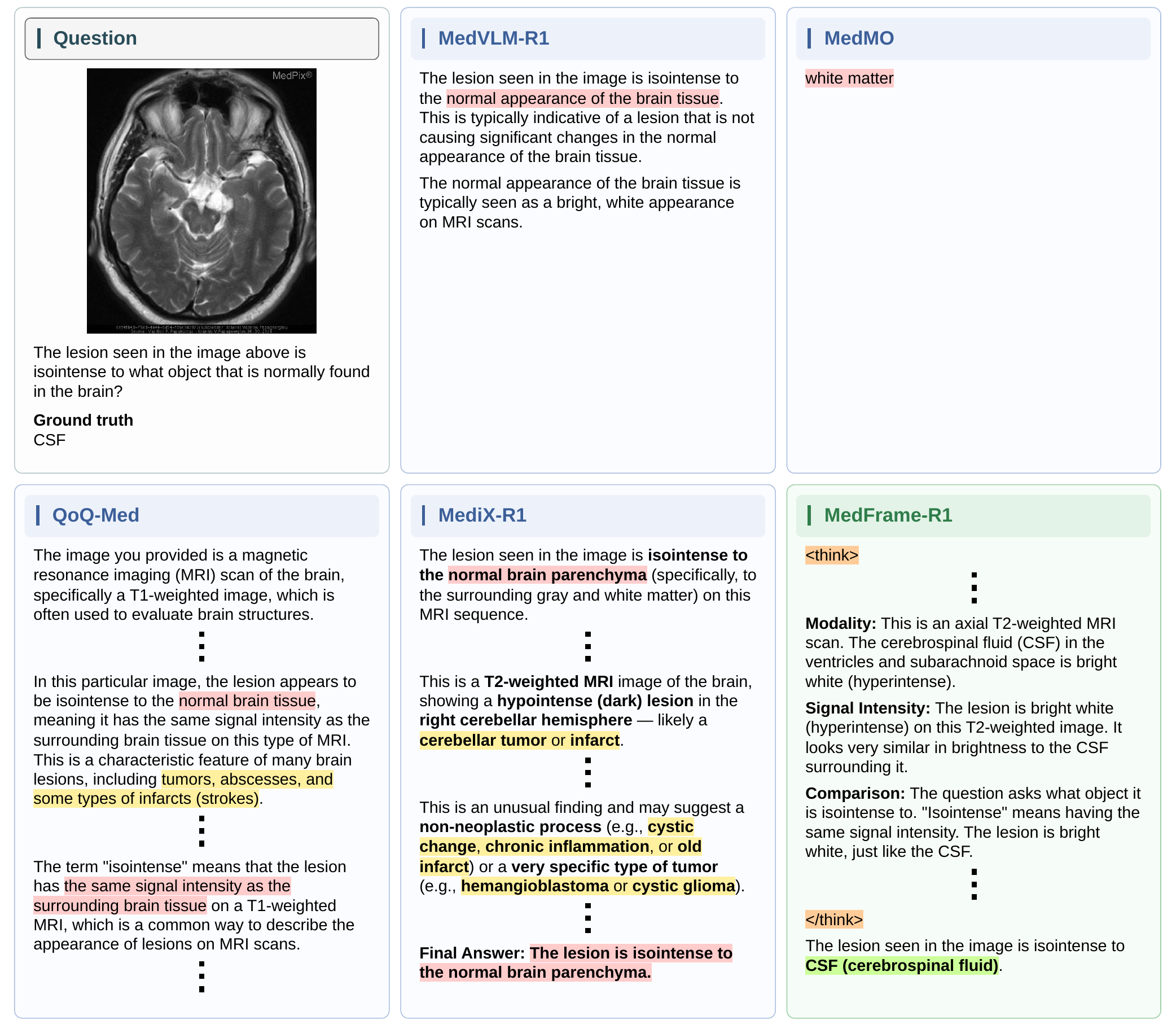}}%
    \usebox{\suppfigbox}
\captionsetup{width=\wd\suppfigbox}
    \caption{\textbf{Answer listing.} The lesion is isointense to CSF. MedMO, MedVLM-R1, QoQ-Med, and MediX-R1 give incorrect answers, with the latter two listing alternatives. MedFrame-R1 correctly identifies CSF through direct comparison. Red indicates incorrect answers, yellow alternatives, and green the correct answer.}
    \label{fig:app-fail-1}
\end{figure*}

\begin{figure*}[tp]
    \centering
    \sbox{\suppfigbox}{\includegraphics[width=0.9\textwidth]{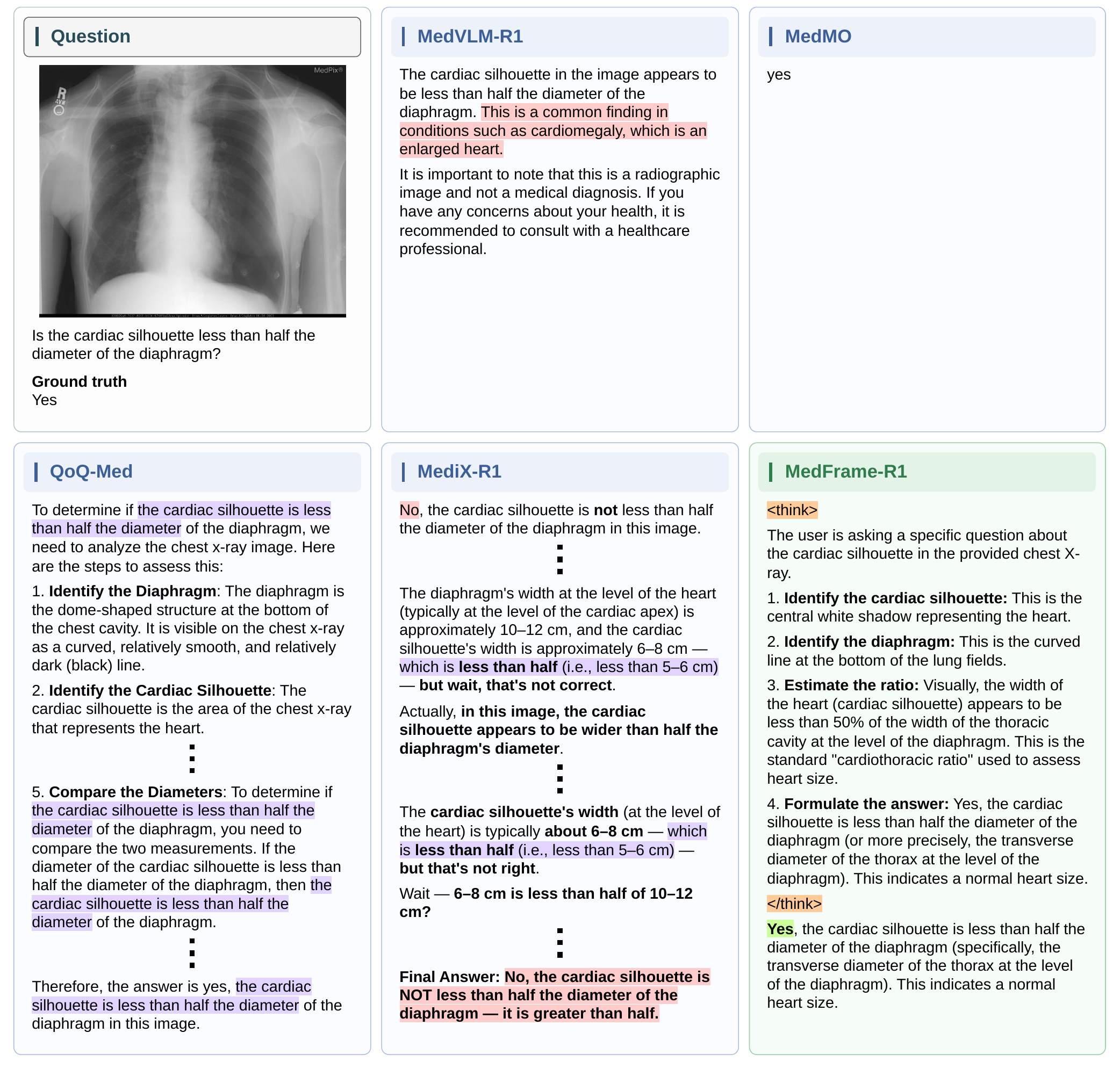}}%
    \usebox{\suppfigbox}
\captionsetup{width=\wd\suppfigbox}
    \caption{\textbf{Repetitive reasoning.} QoQ-Med repeats itself, while MediX-R1 cycles through contradictions before answering incorrectly. MedVLM-R1 misinterprets normal heart size as cardiomegaly, and MedMO provides no reasoning. MedFrame-R1 reasons concisely and answers correctly. Purple marks repetition; other colours follow \cref{fig:app-fail-1}.}
    \label{fig:app-fail-2}
\end{figure*}

\Cref{tab:intro_failures} in the main paper summarises which problems of judge-based medical RL each method addresses. \Cref{fig:app-fail-1,fig:app-fail-2} show two of them in the evaluation answers of each model, answer listing and repetitive reasoning, each on a single VQA-RAD question. Answers are verbatim, and vertical dots indicate omitted text. The MedFrame-R1 answers come from MedFrame-R1-4B (Qwen3.5), whose thinking block is marked by its opening and closing tags.

\section{Comparison with MedMO}
\label{sec:app-medmo}

\begin{figure*}[tp]
\centering
    \sbox{\suppfigbox}{\includegraphics[width=0.9\textwidth]{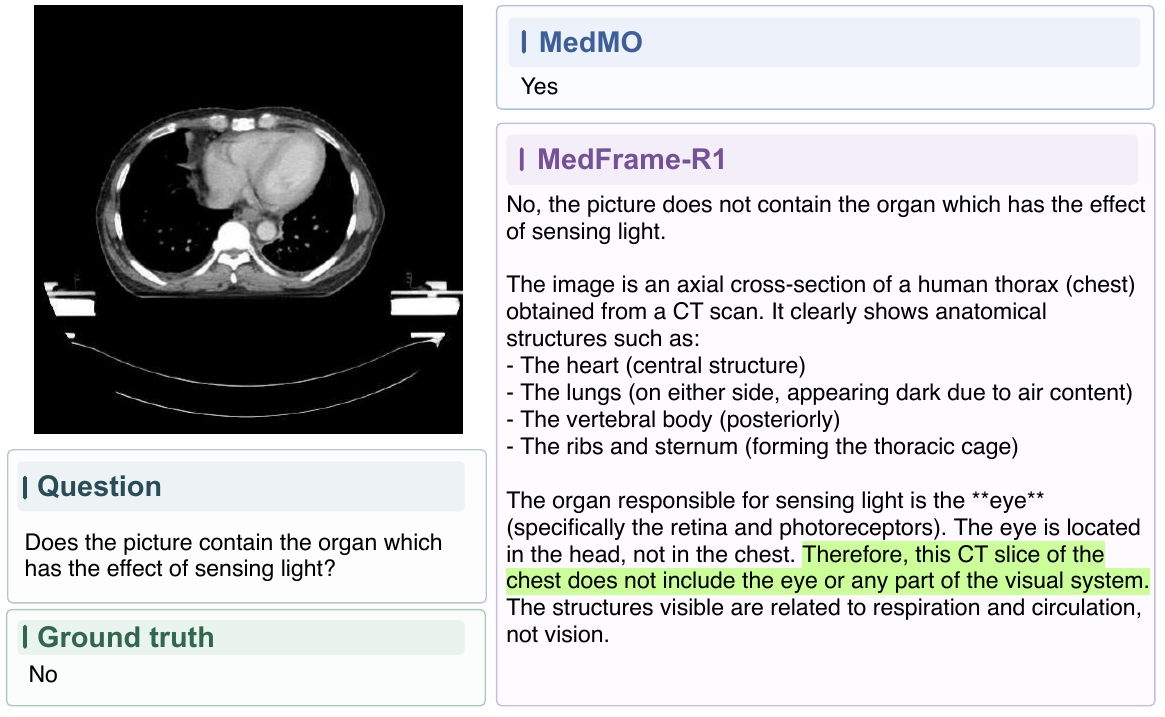}}%
    \usebox{\suppfigbox}
\captionsetup{width=\wd\suppfigbox}
    \caption{\textbf{Answer with and without reasoning.} MedMO answers Yes, while MedFrame-R1 explains why a chest slice contains no eye and answers No.}
    \label{fig:app-medmo-reasoning}
\end{figure*}

\begin{table*}[h]
\centering
\caption{\textbf{MedMO against MedFrame-R1.} Both models at the same size, scored with the same protocol. Bold marks the better of the two in each column.}
\small
\setlength{\tabcolsep}{5.5pt}
\renewcommand{\arraystretch}{1.3}
\setlength{\aboverulesep}{0pt}\setlength{\belowrulesep}{0pt}   
\resizebox{\textwidth}{!}{%
\begin{tabular}{l|cccccc|ccccccccccc|cccc}
\toprule
\multicolumn{1}{c|}{\rule{0pt}{2.2cm}} & \multicolumn{6}{c|}{\cellcolor{cVQA}} & \multicolumn{11}{c|}{\cellcolor{cQA}} & \multicolumn{4}{c}{\cellcolor{cAVG}} \\[-2.37cm]
\multicolumn{1}{c|}{\rule{0pt}{2.2cm}\raisebox{0.93cm}[0pt][0pt]{\large\textbf{Method}}}
& \multicolumn{1}{c}{\raisebox{0.93cm}[0pt][0pt]{\makebox[0pt][c]{\rotatebox[origin=c]{65}{\footnotesize\textbf{SLAKE}}}}}
& \multicolumn{1}{c}{\raisebox{0.93cm}[0pt][0pt]{\makebox[0pt][c]{\rotatebox[origin=c]{65}{\footnotesize\textbf{VQA-RAD}}}}}
& \multicolumn{1}{c}{\raisebox{0.93cm}[0pt][0pt]{\makebox[0pt][c]{\rotatebox[origin=c]{65}{\footnotesize\textbf{PathVQA}}}}}
& \multicolumn{1}{c}{\raisebox{0.93cm}[0pt][0pt]{\makebox[0pt][c]{\rotatebox[origin=c]{65}{\footnotesize\textbf{PMC-VQA}}}}}
& \multicolumn{1}{c}{\raisebox{0.93cm}[0pt][0pt]{\makebox[0pt][c]{\rotatebox[origin=c]{65}{\footnotesize\textbf{PMC-VQA-H}}}}}
& \multicolumn{1}{c|}{\raisebox{0.93cm}[0pt][0pt]{\makebox[0pt][c]{\rotatebox[origin=c]{65}{\footnotesize\textbf{MIMIC-Gen}}}}}
& \multicolumn{1}{c}{\raisebox{0.93cm}[0pt][0pt]{\makebox[0pt][c]{\rotatebox[origin=c]{65}{\footnotesize\textbf{MMLU-Clin}}}}}
& \multicolumn{1}{c}{\raisebox{0.93cm}[0pt][0pt]{\makebox[0pt][c]{\rotatebox[origin=c]{65}{\footnotesize\textbf{MMLU-Bio}}}}}
& \multicolumn{1}{c}{\raisebox{0.93cm}[0pt][0pt]{\makebox[0pt][c]{\rotatebox[origin=c]{65}{\footnotesize\textbf{MMLU-Med}}}}}
& \multicolumn{1}{c}{\raisebox{0.93cm}[0pt][0pt]{\makebox[0pt][c]{\rotatebox[origin=c]{65}{\footnotesize\textbf{MMLU-Gen}}}}}
& \multicolumn{1}{c}{\raisebox{0.93cm}[0pt][0pt]{\makebox[0pt][c]{\rotatebox[origin=c]{65}{\footnotesize\textbf{MMLU-Prof}}}}}
& \multicolumn{1}{c}{\raisebox{0.93cm}[0pt][0pt]{\makebox[0pt][c]{\rotatebox[origin=c]{65}{\footnotesize\textbf{MMLU-Anat}}}}}
& \multicolumn{1}{c}{\raisebox{0.93cm}[0pt][0pt]{\makebox[0pt][c]{\rotatebox[origin=c]{65}{\footnotesize\textbf{MedMCQA}}}}}
& \multicolumn{1}{c}{\raisebox{0.93cm}[0pt][0pt]{\makebox[0pt][c]{\rotatebox[origin=c]{65}{\footnotesize\textbf{MedQA}}}}}
& \multicolumn{1}{c}{\raisebox{0.93cm}[0pt][0pt]{\makebox[0pt][c]{\rotatebox[origin=c]{65}{\footnotesize\textbf{USMLE-SA}}}}}
& \multicolumn{1}{c}{\raisebox{0.93cm}[0pt][0pt]{\makebox[0pt][c]{\rotatebox[origin=c]{65}{\footnotesize\textbf{PubMedQA}}}}}
& \multicolumn{1}{c|}{\raisebox{0.93cm}[0pt][0pt]{\makebox[0pt][c]{\rotatebox[origin=c]{65}{\footnotesize\textbf{MIMIC-Sum}}}}}
& \multicolumn{1}{c}{\raisebox{0.93cm}[0pt][0pt]{\makebox[0pt][c]{\rotatebox[origin=c]{65}{\footnotesize\textbf{VQA}}}}}
& \multicolumn{1}{c}{\raisebox{0.93cm}[0pt][0pt]{\makebox[0pt][c]{\rotatebox[origin=c]{65}{\footnotesize\textbf{QA}}}}}
& \multicolumn{1}{c}{\raisebox{0.93cm}[0pt][0pt]{\makebox[0pt][c]{\rotatebox[origin=c]{65}{\footnotesize\textbf{Macro}}}}}
& \multicolumn{1}{c}{\raisebox{0.93cm}[0pt][0pt]{\makebox[0pt][c]{\rotatebox[origin=c]{65}{\footnotesize\textbf{Micro}}}}} \\
\midrule
MedMO-4B
& 39.0 & 34.6 & 36.7 & 39.3 & 14.8 & 0.6 & \textbf{81.1} & 86.8 & 77.5 & 86.0 & 79.4 & \textbf{70.4} & 57.1 & 66.4 & 67.7 & \textbf{55.4} & 42.2 & 27.5 & 70.0 & 55.0 & 45.1 \\
\rowcolor{cOurs}
\textbf{MedFrame-R1-4B (Qwen3-VL)}
& \textbf{74.4} & \textbf{61.6} & \textbf{52.1} & \textbf{54.7} & \textbf{36.0} & \textbf{32.4} & 80.4 & \textbf{90.3} & \textbf{79.2} & \textbf{88.0} & \textbf{79.8} & 69.6 & \textbf{60.1} & \textbf{70.2} & \textbf{71.4} & 49.8 & \textbf{50.8} & \textbf{51.9} & \textbf{71.8} & \textbf{64.8} & \textbf{57.4} \\
\midrule
MedMO-8B
& 74.1 & 40.8 & 29.3 & 47.3 & 16.4 & 7.0 & \textbf{84.9} & \textbf{91.7} & 81.5 & \textbf{91.0} & \textbf{89.0} & \textbf{76.3} & \textbf{63.7} & \textbf{84.8} & 68.0 & \textbf{54.2} & 48.6 & 35.8 & \textbf{75.8} & 61.7 & 48.6 \\
\rowcolor{cOurs}
\textbf{MedFrame-R1-8B (Qwen3-VL)}
& \textbf{76.5} & \textbf{65.6} & \textbf{53.8} & \textbf{56.2} & \textbf{34.8} & \textbf{33.4} & 82.6 & \textbf{91.7} & \textbf{82.7} & 88.0 & 85.3 & 74.1 & 61.8 & 73.6 & \textbf{76.0} & 46.6 & \textbf{52.8} & \textbf{53.4} & 74.1 & \textbf{66.8} & \textbf{59.1} \\
\bottomrule
\end{tabular}}

\label{tab:app-medmo}
\end{table*}
We further compare MedFrame-R1 with MedMO~\citep{deria2026medmo}, which shares the same Qwen3-VL backbone and uses RL post-training. MedMO is trained on over 26M samples across four stages, including 300K samples for RL, compared with just 51{,}335 prompts in our training corpus. Unlike MedFrame-R1, the released MedMO checkpoints provide direct answers without explicit reasoning, as illustrated in \cref{fig:app-medmo-reasoning}. Despite using over 500 times more training samples, MedMO underperforms MedFrame-R1 at both model sizes (\cref{tab:app-medmo}). MedFrame-R1 achieves macro averages of 64.8 (4B) and 66.8 (8B), compared with 55.0 and 61.7 for MedMO, with the largest improvements on VQA tasks. However, MedMO remains competitive on text benchmarks and outperforms MedFrame-R1 on some of them. The next subsection looks more closely at the explanations that MedFrame-R1 gives with its answers.

\subsection{Generated Reasoning}
\label{sec:app-reasoning}

\begin{figure*}[t]
    \centering
    \sbox{\suppfigbox}{\includegraphics[width=0.9\textwidth]{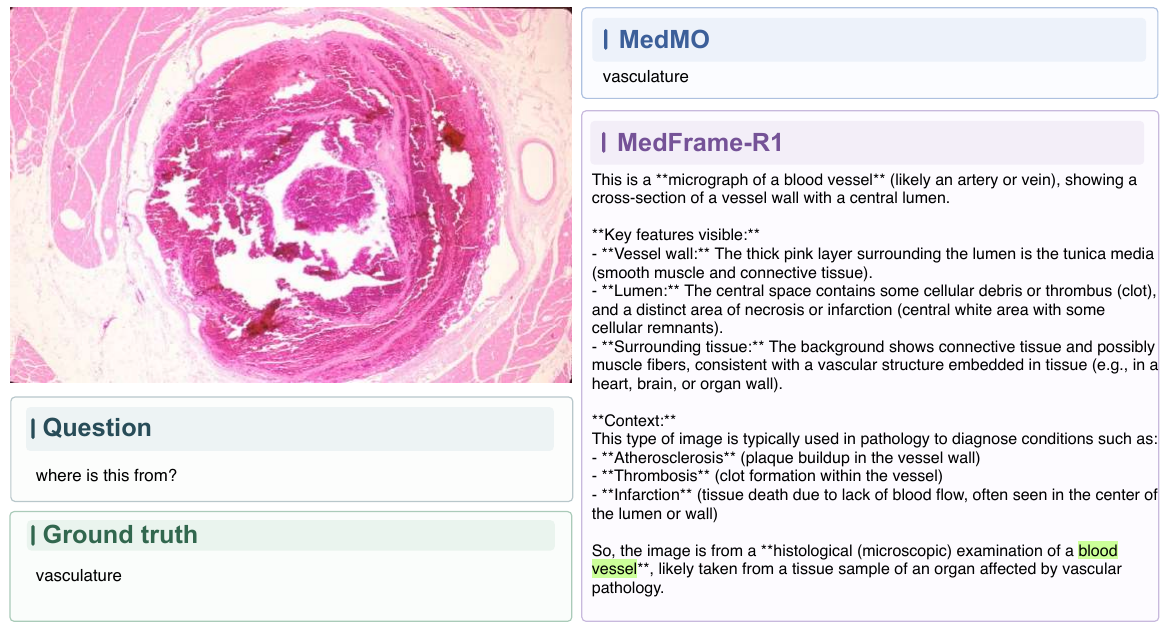}}%
    \usebox{\suppfigbox}
\captionsetup{width=\wd\suppfigbox}
    \caption{\textbf{From a category label to an anatomical explanation.} MedMO answers vasculature, and MedFrame-R1 describes a blood vessel and its structure.}
    \label{fig:app-reasoning-2}
\end{figure*}

\begin{figure*}[t]
    \centering
    \sbox{\suppfigbox}{\includegraphics[width=\textwidth,height=\textheight,keepaspectratio]{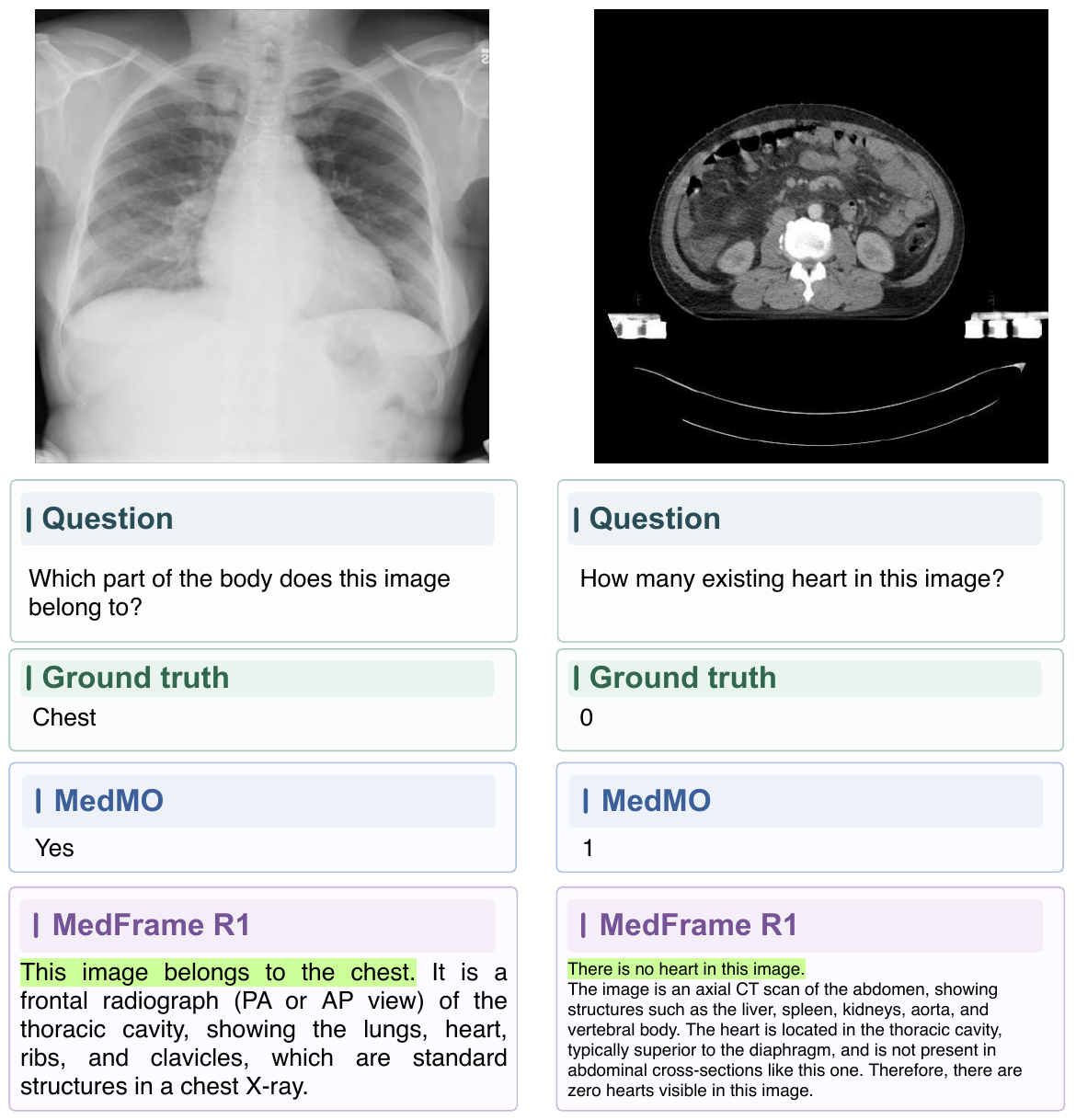}}%
    \usebox{\suppfigbox}
\captionsetup{width=\wd\suppfigbox}
    \caption{\textbf{Body region and organ absence.} Left: MedMO answers Yes to a question about the body region, while MedFrame-R1 names the chest from the visible structures. Right: MedMO counts one heart in an abdominal CT slice, while MedFrame-R1 explains that the heart lies above the diaphragm and answers that no heart is present.}
    \label{fig:app-reasoning-5}
\end{figure*}

The results above measure answer correctness, so this section looks at the explanations that accompany the answers. \Cref{fig:app-reasoning-2,fig:app-reasoning-5} compare responses from MedMO and MedFrame-R1 on medical image questions. Although it is trained with a length reward, MedFrame-R1 still explains its answers, for example by describing the blood vessel and its structure in \cref{fig:app-reasoning-2}, where MedMO only gives a category label. In \cref{fig:app-reasoning-5}, MedMO answers both questions incorrectly, while MedFrame-R1 identifies the chest from the visible structures and explains why an abdominal slice contains no heart. The references do not verify these medical claims.

\clearpage
\end{document}